\documentclass[11pt, a4paper, logo, twocolumn, copyright, nonumbering]{googledeepmind}

\pdftrailerid{redacted}

\makeatletter
\renewcommand\bibentry[1]{\nocite{#1}{\frenchspacing\@nameuse{BR@r@#1\@extra@b@citeb}}}
\makeatother

\usepackage{kantlipsum, lipsum}
\usepackage{dsfont}
\usepackage{gdm-colors}
\usepackage[draft]{minted}

\makeatletter
\def\PYGdefault@reset{\let\PYGdefault@it=\relax \let\PYGdefault@bf=\relax%
    \let\PYGdefault@ul=\relax \let\PYGdefault@tc=\relax%
    \let\PYGdefault@bc=\relax \let\PYGdefault@ff=\relax}
\def\PYGdefault@tok#1{\csname PYGdefault@tok@#1\endcsname}
\def\PYGdefault@toks#1+{\ifx\relax#1\empty\else%
    \PYGdefault@tok{#1}\expandafter\PYGdefault@toks\fi}
\def\PYGdefault@do#1{\PYGdefault@bc{\PYGdefault@tc{\PYGdefault@ul{%
    \PYGdefault@it{\PYGdefault@bf{\PYGdefault@ff{#1}}}}}}}
\def\PYGdefault#1#2{\PYGdefault@reset\PYGdefault@toks#1+\relax+\PYGdefault@do{#2}}

\@namedef{PYGdefault@tok@w}{\def\PYGdefault@tc##1{\textcolor[rgb]{0.73,0.73,0.73}{##1}}}
\@namedef{PYGdefault@tok@c}{\let\PYGdefault@it=\textit\def\PYGdefault@tc##1{\textcolor[rgb]{0.25,0.50,0.50}{##1}}}
\@namedef{PYGdefault@tok@cp}{\def\PYGdefault@tc##1{\textcolor[rgb]{0.74,0.48,0.00}{##1}}}
\@namedef{PYGdefault@tok@k}{\let\PYGdefault@bf=\textbf\def\PYGdefault@tc##1{\textcolor[rgb]{0.00,0.50,0.00}{##1}}}
\@namedef{PYGdefault@tok@kp}{\def\PYGdefault@tc##1{\textcolor[rgb]{0.00,0.50,0.00}{##1}}}
\@namedef{PYGdefault@tok@kt}{\def\PYGdefault@tc##1{\textcolor[rgb]{0.69,0.00,0.25}{##1}}}
\@namedef{PYGdefault@tok@o}{\def\PYGdefault@tc##1{\textcolor[rgb]{0.40,0.40,0.40}{##1}}}
\@namedef{PYGdefault@tok@ow}{\let\PYGdefault@bf=\textbf\def\PYGdefault@tc##1{\textcolor[rgb]{0.67,0.13,1.00}{##1}}}
\@namedef{PYGdefault@tok@nb}{\def\PYGdefault@tc##1{\textcolor[rgb]{0.00,0.50,0.00}{##1}}}
\@namedef{PYGdefault@tok@nf}{\def\PYGdefault@tc##1{\textcolor[rgb]{0.00,0.00,1.00}{##1}}}
\@namedef{PYGdefault@tok@nc}{\let\PYGdefault@bf=\textbf\def\PYGdefault@tc##1{\textcolor[rgb]{0.00,0.00,1.00}{##1}}}
\@namedef{PYGdefault@tok@nn}{\let\PYGdefault@bf=\textbf\def\PYGdefault@tc##1{\textcolor[rgb]{0.00,0.00,1.00}{##1}}}
\@namedef{PYGdefault@tok@ne}{\let\PYGdefault@bf=\textbf\def\PYGdefault@tc##1{\textcolor[rgb]{0.82,0.25,0.23}{##1}}}
\@namedef{PYGdefault@tok@nv}{\def\PYGdefault@tc##1{\textcolor[rgb]{0.10,0.09,0.49}{##1}}}
\@namedef{PYGdefault@tok@no}{\def\PYGdefault@tc##1{\textcolor[rgb]{0.53,0.00,0.00}{##1}}}
\@namedef{PYGdefault@tok@nl}{\def\PYGdefault@tc##1{\textcolor[rgb]{0.63,0.63,0.00}{##1}}}
\@namedef{PYGdefault@tok@ni}{\let\PYGdefault@bf=\textbf\def\PYGdefault@tc##1{\textcolor[rgb]{0.60,0.60,0.60}{##1}}}
\@namedef{PYGdefault@tok@na}{\def\PYGdefault@tc##1{\textcolor[rgb]{0.49,0.56,0.16}{##1}}}
\@namedef{PYGdefault@tok@nt}{\let\PYGdefault@bf=\textbf\def\PYGdefault@tc##1{\textcolor[rgb]{0.00,0.50,0.00}{##1}}}
\@namedef{PYGdefault@tok@nd}{\def\PYGdefault@tc##1{\textcolor[rgb]{0.67,0.13,1.00}{##1}}}
\@namedef{PYGdefault@tok@s}{\def\PYGdefault@tc##1{\textcolor[rgb]{0.73,0.13,0.13}{##1}}}
\@namedef{PYGdefault@tok@sd}{\let\PYGdefault@it=\textit\def\PYGdefault@tc##1{\textcolor[rgb]{0.73,0.13,0.13}{##1}}}
\@namedef{PYGdefault@tok@si}{\let\PYGdefault@bf=\textbf\def\PYGdefault@tc##1{\textcolor[rgb]{0.73,0.40,0.53}{##1}}}
\@namedef{PYGdefault@tok@se}{\let\PYGdefault@bf=\textbf\def\PYGdefault@tc##1{\textcolor[rgb]{0.73,0.40,0.13}{##1}}}
\@namedef{PYGdefault@tok@sr}{\def\PYGdefault@tc##1{\textcolor[rgb]{0.73,0.40,0.53}{##1}}}
\@namedef{PYGdefault@tok@ss}{\def\PYGdefault@tc##1{\textcolor[rgb]{0.10,0.09,0.49}{##1}}}
\@namedef{PYGdefault@tok@sx}{\def\PYGdefault@tc##1{\textcolor[rgb]{0.00,0.50,0.00}{##1}}}
\@namedef{PYGdefault@tok@m}{\def\PYGdefault@tc##1{\textcolor[rgb]{0.40,0.40,0.40}{##1}}}
\@namedef{PYGdefault@tok@gh}{\let\PYGdefault@bf=\textbf\def\PYGdefault@tc##1{\textcolor[rgb]{0.00,0.00,0.50}{##1}}}
\@namedef{PYGdefault@tok@gu}{\let\PYGdefault@bf=\textbf\def\PYGdefault@tc##1{\textcolor[rgb]{0.50,0.00,0.50}{##1}}}
\@namedef{PYGdefault@tok@gd}{\def\PYGdefault@tc##1{\textcolor[rgb]{0.63,0.00,0.00}{##1}}}
\@namedef{PYGdefault@tok@gi}{\def\PYGdefault@tc##1{\textcolor[rgb]{0.00,0.63,0.00}{##1}}}
\@namedef{PYGdefault@tok@gr}{\def\PYGdefault@tc##1{\textcolor[rgb]{1.00,0.00,0.00}{##1}}}
\@namedef{PYGdefault@tok@ge}{\let\PYGdefault@it=\textit}
\@namedef{PYGdefault@tok@gs}{\let\PYGdefault@bf=\textbf}
\@namedef{PYGdefault@tok@gp}{\let\PYGdefault@bf=\textbf\def\PYGdefault@tc##1{\textcolor[rgb]{0.00,0.00,0.50}{##1}}}
\@namedef{PYGdefault@tok@go}{\def\PYGdefault@tc##1{\textcolor[rgb]{0.53,0.53,0.53}{##1}}}
\@namedef{PYGdefault@tok@gt}{\def\PYGdefault@tc##1{\textcolor[rgb]{0.00,0.27,0.87}{##1}}}
\@namedef{PYGdefault@tok@err}{\def\PYGdefault@bc##1{{\setlength{\fboxsep}{\string -\fboxrule}\fcolorbox[rgb]{1.00,0.00,0.00}{1,1,1}{\strut ##1}}}}
\@namedef{PYGdefault@tok@kc}{\let\PYGdefault@bf=\textbf\def\PYGdefault@tc##1{\textcolor[rgb]{0.00,0.50,0.00}{##1}}}
\@namedef{PYGdefault@tok@kd}{\let\PYGdefault@bf=\textbf\def\PYGdefault@tc##1{\textcolor[rgb]{0.00,0.50,0.00}{##1}}}
\@namedef{PYGdefault@tok@kn}{\let\PYGdefault@bf=\textbf\def\PYGdefault@tc##1{\textcolor[rgb]{0.00,0.50,0.00}{##1}}}
\@namedef{PYGdefault@tok@kr}{\let\PYGdefault@bf=\textbf\def\PYGdefault@tc##1{\textcolor[rgb]{0.00,0.50,0.00}{##1}}}
\@namedef{PYGdefault@tok@bp}{\def\PYGdefault@tc##1{\textcolor[rgb]{0.00,0.50,0.00}{##1}}}
\@namedef{PYGdefault@tok@fm}{\def\PYGdefault@tc##1{\textcolor[rgb]{0.00,0.00,1.00}{##1}}}
\@namedef{PYGdefault@tok@vc}{\def\PYGdefault@tc##1{\textcolor[rgb]{0.10,0.09,0.49}{##1}}}
\@namedef{PYGdefault@tok@vg}{\def\PYGdefault@tc##1{\textcolor[rgb]{0.10,0.09,0.49}{##1}}}
\@namedef{PYGdefault@tok@vi}{\def\PYGdefault@tc##1{\textcolor[rgb]{0.10,0.09,0.49}{##1}}}
\@namedef{PYGdefault@tok@vm}{\def\PYGdefault@tc##1{\textcolor[rgb]{0.10,0.09,0.49}{##1}}}
\@namedef{PYGdefault@tok@sa}{\def\PYGdefault@tc##1{\textcolor[rgb]{0.73,0.13,0.13}{##1}}}
\@namedef{PYGdefault@tok@sb}{\def\PYGdefault@tc##1{\textcolor[rgb]{0.73,0.13,0.13}{##1}}}
\@namedef{PYGdefault@tok@sc}{\def\PYGdefault@tc##1{\textcolor[rgb]{0.73,0.13,0.13}{##1}}}
\@namedef{PYGdefault@tok@dl}{\def\PYGdefault@tc##1{\textcolor[rgb]{0.73,0.13,0.13}{##1}}}
\@namedef{PYGdefault@tok@s2}{\def\PYGdefault@tc##1{\textcolor[rgb]{0.73,0.13,0.13}{##1}}}
\@namedef{PYGdefault@tok@sh}{\def\PYGdefault@tc##1{\textcolor[rgb]{0.73,0.13,0.13}{##1}}}
\@namedef{PYGdefault@tok@s1}{\def\PYGdefault@tc##1{\textcolor[rgb]{0.73,0.13,0.13}{##1}}}
\@namedef{PYGdefault@tok@mb}{\def\PYGdefault@tc##1{\textcolor[rgb]{0.40,0.40,0.40}{##1}}}
\@namedef{PYGdefault@tok@mf}{\def\PYGdefault@tc##1{\textcolor[rgb]{0.40,0.40,0.40}{##1}}}
\@namedef{PYGdefault@tok@mh}{\def\PYGdefault@tc##1{\textcolor[rgb]{0.40,0.40,0.40}{##1}}}
\@namedef{PYGdefault@tok@mi}{\def\PYGdefault@tc##1{\textcolor[rgb]{0.40,0.40,0.40}{##1}}}
\@namedef{PYGdefault@tok@il}{\def\PYGdefault@tc##1{\textcolor[rgb]{0.40,0.40,0.40}{##1}}}
\@namedef{PYGdefault@tok@mo}{\def\PYGdefault@tc##1{\textcolor[rgb]{0.40,0.40,0.40}{##1}}}
\@namedef{PYGdefault@tok@ch}{\let\PYGdefault@it=\textit\def\PYGdefault@tc##1{\textcolor[rgb]{0.25,0.50,0.50}{##1}}}
\@namedef{PYGdefault@tok@cm}{\let\PYGdefault@it=\textit\def\PYGdefault@tc##1{\textcolor[rgb]{0.25,0.50,0.50}{##1}}}
\@namedef{PYGdefault@tok@cpf}{\let\PYGdefault@it=\textit\def\PYGdefault@tc##1{\textcolor[rgb]{0.25,0.50,0.50}{##1}}}
\@namedef{PYGdefault@tok@c1}{\let\PYGdefault@it=\textit\def\PYGdefault@tc##1{\textcolor[rgb]{0.25,0.50,0.50}{##1}}}
\@namedef{PYGdefault@tok@cs}{\let\PYGdefault@it=\textit\def\PYGdefault@tc##1{\textcolor[rgb]{0.25,0.50,0.50}{##1}}}

\makeatother

\makeatletter
\def\PYG@reset{\let\PYG@it=\relax \let\PYG@bf=\relax%
    \let\PYG@ul=\relax \let\PYG@tc=\relax%
    \let\PYG@bc=\relax \let\PYG@ff=\relax}
\def\PYG@tok#1{\csname PYG@tok@#1\endcsname}
\def\PYG@toks#1+{\ifx\relax#1\empty\else%
    \PYG@tok{#1}\expandafter\PYG@toks\fi}
\def\PYG@do#1{\PYG@bc{\PYG@tc{\PYG@ul{%
    \PYG@it{\PYG@bf{\PYG@ff{#1}}}}}}}
\def\PYG#1#2{\PYG@reset\PYG@toks#1+\relax+\PYG@do{#2}}

\@namedef{PYG@tok@w}{\def\PYG@tc##1{\textcolor[rgb]{0.73,0.73,0.73}{##1}}}
\@namedef{PYG@tok@c}{\let\PYG@it=\textit\def\PYG@tc##1{\textcolor[rgb]{0.25,0.50,0.50}{##1}}}
\@namedef{PYG@tok@cp}{\def\PYG@tc##1{\textcolor[rgb]{0.74,0.48,0.00}{##1}}}
\@namedef{PYG@tok@k}{\let\PYG@bf=\textbf\def\PYG@tc##1{\textcolor[rgb]{0.00,0.50,0.00}{##1}}}
\@namedef{PYG@tok@kp}{\def\PYG@tc##1{\textcolor[rgb]{0.00,0.50,0.00}{##1}}}
\@namedef{PYG@tok@kt}{\def\PYG@tc##1{\textcolor[rgb]{0.69,0.00,0.25}{##1}}}
\@namedef{PYG@tok@o}{\def\PYG@tc##1{\textcolor[rgb]{0.40,0.40,0.40}{##1}}}
\@namedef{PYG@tok@ow}{\let\PYG@bf=\textbf\def\PYG@tc##1{\textcolor[rgb]{0.67,0.13,1.00}{##1}}}
\@namedef{PYG@tok@nb}{\def\PYG@tc##1{\textcolor[rgb]{0.00,0.50,0.00}{##1}}}
\@namedef{PYG@tok@nf}{\def\PYG@tc##1{\textcolor[rgb]{0.00,0.00,1.00}{##1}}}
\@namedef{PYG@tok@nc}{\let\PYG@bf=\textbf\def\PYG@tc##1{\textcolor[rgb]{0.00,0.00,1.00}{##1}}}
\@namedef{PYG@tok@nn}{\let\PYG@bf=\textbf\def\PYG@tc##1{\textcolor[rgb]{0.00,0.00,1.00}{##1}}}
\@namedef{PYG@tok@ne}{\let\PYG@bf=\textbf\def\PYG@tc##1{\textcolor[rgb]{0.82,0.25,0.23}{##1}}}
\@namedef{PYG@tok@nv}{\def\PYG@tc##1{\textcolor[rgb]{0.10,0.09,0.49}{##1}}}
\@namedef{PYG@tok@no}{\def\PYG@tc##1{\textcolor[rgb]{0.53,0.00,0.00}{##1}}}
\@namedef{PYG@tok@nl}{\def\PYG@tc##1{\textcolor[rgb]{0.63,0.63,0.00}{##1}}}
\@namedef{PYG@tok@ni}{\let\PYG@bf=\textbf\def\PYG@tc##1{\textcolor[rgb]{0.60,0.60,0.60}{##1}}}
\@namedef{PYG@tok@na}{\def\PYG@tc##1{\textcolor[rgb]{0.49,0.56,0.16}{##1}}}
\@namedef{PYG@tok@nt}{\let\PYG@bf=\textbf\def\PYG@tc##1{\textcolor[rgb]{0.00,0.50,0.00}{##1}}}
\@namedef{PYG@tok@nd}{\def\PYG@tc##1{\textcolor[rgb]{0.67,0.13,1.00}{##1}}}
\@namedef{PYG@tok@s}{\def\PYG@tc##1{\textcolor[rgb]{0.73,0.13,0.13}{##1}}}
\@namedef{PYG@tok@sd}{\let\PYG@it=\textit\def\PYG@tc##1{\textcolor[rgb]{0.73,0.13,0.13}{##1}}}
\@namedef{PYG@tok@si}{\let\PYG@bf=\textbf\def\PYG@tc##1{\textcolor[rgb]{0.73,0.40,0.53}{##1}}}
\@namedef{PYG@tok@se}{\let\PYG@bf=\textbf\def\PYG@tc##1{\textcolor[rgb]{0.73,0.40,0.13}{##1}}}
\@namedef{PYG@tok@sr}{\def\PYG@tc##1{\textcolor[rgb]{0.73,0.40,0.53}{##1}}}
\@namedef{PYG@tok@ss}{\def\PYG@tc##1{\textcolor[rgb]{0.10,0.09,0.49}{##1}}}
\@namedef{PYG@tok@sx}{\def\PYG@tc##1{\textcolor[rgb]{0.00,0.50,0.00}{##1}}}
\@namedef{PYG@tok@m}{\def\PYG@tc##1{\textcolor[rgb]{0.40,0.40,0.40}{##1}}}
\@namedef{PYG@tok@gh}{\let\PYG@bf=\textbf\def\PYG@tc##1{\textcolor[rgb]{0.00,0.00,0.50}{##1}}}
\@namedef{PYG@tok@gu}{\let\PYG@bf=\textbf\def\PYG@tc##1{\textcolor[rgb]{0.50,0.00,0.50}{##1}}}
\@namedef{PYG@tok@gd}{\def\PYG@tc##1{\textcolor[rgb]{0.63,0.00,0.00}{##1}}}
\@namedef{PYG@tok@gi}{\def\PYG@tc##1{\textcolor[rgb]{0.00,0.63,0.00}{##1}}}
\@namedef{PYG@tok@gr}{\def\PYG@tc##1{\textcolor[rgb]{1.00,0.00,0.00}{##1}}}
\@namedef{PYG@tok@ge}{\let\PYG@it=\textit}
\@namedef{PYG@tok@gs}{\let\PYG@bf=\textbf}
\@namedef{PYG@tok@gp}{\let\PYG@bf=\textbf\def\PYG@tc##1{\textcolor[rgb]{0.00,0.00,0.50}{##1}}}
\@namedef{PYG@tok@go}{\def\PYG@tc##1{\textcolor[rgb]{0.53,0.53,0.53}{##1}}}
\@namedef{PYG@tok@gt}{\def\PYG@tc##1{\textcolor[rgb]{0.00,0.27,0.87}{##1}}}
\@namedef{PYG@tok@err}{\def\PYG@bc##1{{\setlength{\fboxsep}{\string -\fboxrule}\fcolorbox[rgb]{1.00,0.00,0.00}{1,1,1}{\strut ##1}}}}
\@namedef{PYG@tok@kc}{\let\PYG@bf=\textbf\def\PYG@tc##1{\textcolor[rgb]{0.00,0.50,0.00}{##1}}}
\@namedef{PYG@tok@kd}{\let\PYG@bf=\textbf\def\PYG@tc##1{\textcolor[rgb]{0.00,0.50,0.00}{##1}}}
\@namedef{PYG@tok@kn}{\let\PYG@bf=\textbf\def\PYG@tc##1{\textcolor[rgb]{0.00,0.50,0.00}{##1}}}
\@namedef{PYG@tok@kr}{\let\PYG@bf=\textbf\def\PYG@tc##1{\textcolor[rgb]{0.00,0.50,0.00}{##1}}}
\@namedef{PYG@tok@bp}{\def\PYG@tc##1{\textcolor[rgb]{0.00,0.50,0.00}{##1}}}
\@namedef{PYG@tok@fm}{\def\PYG@tc##1{\textcolor[rgb]{0.00,0.00,1.00}{##1}}}
\@namedef{PYG@tok@vc}{\def\PYG@tc##1{\textcolor[rgb]{0.10,0.09,0.49}{##1}}}
\@namedef{PYG@tok@vg}{\def\PYG@tc##1{\textcolor[rgb]{0.10,0.09,0.49}{##1}}}
\@namedef{PYG@tok@vi}{\def\PYG@tc##1{\textcolor[rgb]{0.10,0.09,0.49}{##1}}}
\@namedef{PYG@tok@vm}{\def\PYG@tc##1{\textcolor[rgb]{0.10,0.09,0.49}{##1}}}
\@namedef{PYG@tok@sa}{\def\PYG@tc##1{\textcolor[rgb]{0.73,0.13,0.13}{##1}}}
\@namedef{PYG@tok@sb}{\def\PYG@tc##1{\textcolor[rgb]{0.73,0.13,0.13}{##1}}}
\@namedef{PYG@tok@sc}{\def\PYG@tc##1{\textcolor[rgb]{0.73,0.13,0.13}{##1}}}
\@namedef{PYG@tok@dl}{\def\PYG@tc##1{\textcolor[rgb]{0.73,0.13,0.13}{##1}}}
\@namedef{PYG@tok@s2}{\def\PYG@tc##1{\textcolor[rgb]{0.73,0.13,0.13}{##1}}}
\@namedef{PYG@tok@sh}{\def\PYG@tc##1{\textcolor[rgb]{0.73,0.13,0.13}{##1}}}
\@namedef{PYG@tok@s1}{\def\PYG@tc##1{\textcolor[rgb]{0.73,0.13,0.13}{##1}}}
\@namedef{PYG@tok@mb}{\def\PYG@tc##1{\textcolor[rgb]{0.40,0.40,0.40}{##1}}}
\@namedef{PYG@tok@mf}{\def\PYG@tc##1{\textcolor[rgb]{0.40,0.40,0.40}{##1}}}
\@namedef{PYG@tok@mh}{\def\PYG@tc##1{\textcolor[rgb]{0.40,0.40,0.40}{##1}}}
\@namedef{PYG@tok@mi}{\def\PYG@tc##1{\textcolor[rgb]{0.40,0.40,0.40}{##1}}}
\@namedef{PYG@tok@il}{\def\PYG@tc##1{\textcolor[rgb]{0.40,0.40,0.40}{##1}}}
\@namedef{PYG@tok@mo}{\def\PYG@tc##1{\textcolor[rgb]{0.40,0.40,0.40}{##1}}}
\@namedef{PYG@tok@ch}{\let\PYG@it=\textit\def\PYG@tc##1{\textcolor[rgb]{0.25,0.50,0.50}{##1}}}
\@namedef{PYG@tok@cm}{\let\PYG@it=\textit\def\PYG@tc##1{\textcolor[rgb]{0.25,0.50,0.50}{##1}}}
\@namedef{PYG@tok@cpf}{\let\PYG@it=\textit\def\PYG@tc##1{\textcolor[rgb]{0.25,0.50,0.50}{##1}}}
\@namedef{PYG@tok@c1}{\let\PYG@it=\textit\def\PYG@tc##1{\textcolor[rgb]{0.25,0.50,0.50}{##1}}}
\@namedef{PYG@tok@cs}{\let\PYG@it=\textit\def\PYG@tc##1{\textcolor[rgb]{0.25,0.50,0.50}{##1}}}

\def\PYGZbs{\char`\\}
\def\PYGZus{\char`\_}
\def\PYGZob{\char`\{}
\def\PYGZcb{\char`\}}

\def\PYGZlt{\char`\<}
\def\PYGZgt{\char`\>}
\def\PYGZsh{\char`\#}
\def\PYGZpc{\char`\%}

\def\PYGZhy{\char`\-}
\def\PYGZsq{\char`\'}
\def\PYGZdq{\char`\"}

\makeatother

\usepackage[authoryear, sort&compress, round]{natbib}

\graphicspath{{figures/}}

\title{Amplifying human performance in combinatorial competitive programming}

\correspondingauthor{petarv@google.com, anovikov@google.com}

\reportnumber{} 

\author[*,1]{Petar Veli\v{c}kovi\'{c}}
\author[*,1]{Alex Vitvitskyi}
\author[*,1]{Larisa Markeeva}
\author[*,1]{Borja Ibarz}
\author[1]{Lars Buesing}
\author[1]{Matej Balog}
\author[*,1]{Alexander Novikov}

\affil[*]{Equal contributions}
\affil[1]{Google DeepMind}

\begin{abstract}
Recent years have seen a significant surge in complex AI systems for competitive programming, capable of performing at admirable levels against human competitors. While steady progress has been made, the highest percentiles still remain out of reach for these methods on standard competition platforms such as Codeforces. Here we instead focus on \emph{combinatorial} competitive programming, where the target is to find as-good-as-possible solutions to otherwise computationally intractable problems, over specific given inputs. We hypothesise that this scenario offers a unique testbed for human-AI synergy, as human programmers can write a backbone of a heuristic solution, after which AI can be used to optimise the scoring function used by the heuristic. We deploy our approach on previous iterations of Hash Code, a global team programming competition inspired by NP-hard software engineering problems at Google, and we leverage FunSearch to evolve our scoring functions. Our evolved solutions significantly improve the attained scores from their baseline, successfully breaking into the top percentile on all previous Hash Code online qualification rounds, and outperforming the top human teams on several. Our method is also performant on an optimisation problem that featured in a recent held-out AtCoder contest.
\end{abstract}

\begin{document}

\maketitle

\newcommand{\expect}[2]{\mathds{E}_{{#1}} \left[ {#2} \right]}
\newcommand{\myvec}[1]{\boldsymbol{#1}}
\newcommand{\myvecsym}[1]{\boldsymbol{#1}}
\newcommand{\vx}{\myvec{x}}
\newcommand{\vy}{\myvec{y}}
\newcommand{\vz}{\myvec{z}}
\newcommand{\vtheta}{\myvecsym{\theta}}

\section{Introduction}

Competitive programming---the art of writing highly specialised code for solving challenging problems under various constraints---represents a critical test of advanced reasoning skills. It has sparked interest as an evaluation of AI systems---with systems like AlphaCode \citep{li2022competition} demonstrating good performance given a language model equipped with a filtering procedure.

In recent years, capabilities of AI-powered competitive programming systems have been steadily increasing---breaking into and beyond the 85th percentile on platforms such as Codeforces \citep{leblond2023alphacode,openai2024o1}, largely relying on improved base model capabilities, data curation and test-time compute.

While such results are certainly impressive, they are still insufficient for consistently outperforming the highest echelons of human competitive programmers. The largest reported Codeforces ELO rating for an AI system at the time of writing is still below $1,900$. The most powerful human competitors are capable of significantly higher feats than that, with one -- Gennady Korotkevich (\texttt{tourist}) -- achieving ELO above $4,000$\footnote{\url{https://codeforces.com/profile/tourist}}. It is our opinion that further breakthroughs are needed before AI systems reach such levels at Codeforces-style competitions.

Meanwhile, another type of impressive AI system has recently emerged: leveraging evolutionary algorithms for searching in the function space \citep[FunSearch]{funsearch}, expressed using code implementations written by language models. We believe that FunSearch opens up an opportunity for a different kind of top-level competitive programming result: one on \emph{combinatorial optimisation} challenges, made \emph{in collaboration} with human competitors.

We present a successful result of this kind: by applying FunSearch on a human-designed solution backbone across several Hash Code competition tasks, we are able to significantly amplify the scores obtained by the backbone. In several contests, we recover solutions that would have outperformed the top human teams. We also validate our method on a variant of a recently-held AtCoder Heuristic Contest.

\section{Combinatorial competitive programming}

Before diving into the specifics of our approach, we briefly discuss why combinatorial tasks are a more immediate fit for present-day AI systems to achieve top-tier levels, compared to Codeforces.

Codeforces-style tasks typically require writing tractable, \emph{polynomial-time} algorithms---with no reward given for partially correct or inefficient solutions. As many of the testcases for such tasks rely on hidden edge cases, it is generally not sufficient for AI to produce code that works correctly in most cases---careful understanding is required of many constraints which might not be obvious at all from the task statement. Further, many such tasks follow specialised design patterns, which well-versed human competitors readily recognise and apply---making it harder for AI to catch up.

In contrast, combinatorial optimisation problems are typically \emph{intractable}, NP-hard problems, with testcases for which optimal solutions are usually unknown. As such, valid-but-suboptimal solutions are inevitable, and they are all rewarded depending on their measured performance on these testcases. Further, the testcases are often \emph{visible} to the competitors, avoiding any hidden edge-case situations. In summary, even if it might seem counter-intuitive, combinatorial optimisation competitions offer a more immediate opportunity for exceptional results: they do not punish suboptimality, all cases of importance are known upfront, and even the strongest human competitors do not know how to find optimal solutions.

Next, we describe the specific competition---Hash Code---we used to substantiate this claim.

\subsubsection{Hash Code competitions}

Hash Code is a former Google-organised global (2--4 person) team programming competition. It features NP-hard optimisation problems inspired by software engineering tasks at Google. Teams produce outputs for a given set of inputs to an intractable optimisation problem, and are evaluated on the quality of their outputs. At the peak of its popularity, it was globally renowned as one of the most popular combinatorial optimisation competitions, inviting over 125,000 competitors,\footnote{\url{www.youtube.com/watch?v=YPOVd-hQUjA}} and featuring some of the most successful competitive programmers---including \texttt{tourist}.\footnote{\url{https://x.com/que_tourist/status/1230760203903070208}} 

Hash Code is, at its core, a contest requiring design of a strong heuristic. Top teams tend to write scoring functions for a locally-optimal greedy search, often followed by randomised hill-climbing. We believe devising strong scoring functions for greedy search is highly non-trivial, and hence amenable to automated genetic programming tools such as FunSearch. Under the Hash Code rules, use of external tools (e.g. combinatorial solver packages) was allowed, which would have made our approach legal at the time as well.

Hash Code has had two phases in most years---first, an \emph{online qualification round}, followed by a \emph{final round} for the top $\sim$50 teams in the qualification. For our purposes, we use only the qualifications, as they have a much wider pool of competitors to compare against, with their tasks still being intractable in their nature. We access all problems and their inputs at \url{https://github.com/google/coding-competitions-archive}.

\section{Experimental setup}

\begin{figure}
    \centering
    \includegraphics[width=0.8\linewidth]{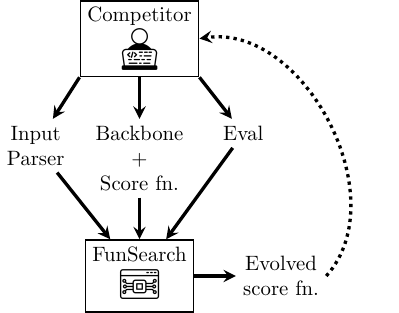}
    \caption{High-level overview of the collaborative competitor + AI approach explored in our work.}
    \label{fig:dataflow}
\end{figure}

Now we can dive deeper into how we leveraged FunSearch to amplify typical strategies that competent competitors may leverage in such contests. Please refer to Figure \ref{fig:dataflow} for a high-level overview.

\begin{figure*}
\centering
\begin{scriptsize}
\begin{Verbatim}[commandchars=\\\{\}]
\PYG{k}{def} \PYG{n+nf}{score\PYGZus{}greedy}\PYG{p}{(}\PYG{n}{project}\PYG{p}{:} \PYG{n}{Project}\PYG{p}{,}
                 \PYG{n}{role\PYGZus{}id}\PYG{p}{:} \PYG{n+nb}{int}\PYG{p}{,}
                 \PYG{n}{assignments}\PYG{p}{:} \PYG{n+nb}{list}\PYG{p}{[}\PYG{n+nb}{int}\PYG{p}{,} \PYG{n+nb}{list}\PYG{p}{[}\PYG{n+nb}{int}\PYG{p}{]],}
                 \PYG{n}{rate\PYGZus{}project}\PYG{p}{:} \PYG{n+nb}{bool}\PYG{p}{)} \PYG{o}{\PYGZhy{}\PYGZgt{}} \PYG{n+nb}{int}\PYG{p}{:}
  \PYG{l+s+sd}{\PYGZdq{}\PYGZdq{}\PYGZdq{}Scorer function for the greedy algorithm.}

\PYG{l+s+sd}{  Return either the value of choosing a particular project, or the value of a}
\PYG{l+s+sd}{  particular role within that project, conditioned on the already scheduled}
\PYG{l+s+sd}{  project assignments. The behaviour is controlled by a boolean, `rate\PYGZus{}project`:}
\PYG{l+s+sd}{  if it is True, we rate the provided project, if it is False, we rate the}
\PYG{l+s+sd}{  provided role ID.}

\PYG{l+s+sd}{  Args:}
\PYG{l+s+sd}{    project: the Project currently considered for choosing.}
\PYG{l+s+sd}{    role\PYGZus{}id: the currently considered role ID from the current project.}
\PYG{l+s+sd}{    assignments: the already\PYGZhy{}scheduled project assignments thus far.}
\PYG{l+s+sd}{    rate\PYGZus{}project: whether we\PYGZsq{}re scoring a project (True) or a role (False).}

\PYG{l+s+sd}{  Returns:}
\PYG{l+s+sd}{    score}
\PYG{l+s+sd}{  \PYGZdq{}\PYGZdq{}\PYGZdq{}}
  \PYG{k}{if} \PYG{n}{rate\PYGZus{}project}\PYG{p}{:}
    \PYG{k}{return} \PYG{l+m+mi}{1}
  \PYG{k}{else}\PYG{p}{:}
    \PYG{n}{skill}\PYG{p}{,} \PYG{n}{level} \PYG{o}{=} \PYG{n}{project}\PYG{o}{.}\PYG{n}{roles}\PYG{p}{[}\PYG{n}{role\PYGZus{}id}\PYG{p}{]}
    \PYG{k}{return} \PYG{l+m+mi}{1}
\end{Verbatim}
\end{scriptsize}
\caption{The base scoring function used within one of the backbones for the Hash Code 2022 Qualification Round (Mentorship and Teamwork). Note the split on the \texttt{rate\_project} variable in order to enable two different choice points to be evolved within the same scoring function.}
\label{fig:score_base}
\end{figure*}

\subsection{Overall workflow}

The workflow of our approach is as follows:
\begin{itemize}
    \item We implement a backbone of a \emph{greedy algorithm} that tackles the given problem, along with functions to parse the input file(s) and evaluate the fitness of candidate solutions;
    \item The greedy algorithm depends on a \emph{scoring function} that weighs in on each of its possible next steps. Initially, we can make this a simple function---see Figure \ref{fig:score_base} for an example;
    \item We use FunSearch to evolve this function, using the number of points achieved on a given testcase as the fitness function;
    \item When relevant, we may analyse the outputs of FunSearch to iterate on the backbone structure, strap on local search,\footnote{Local search was only used on one previous testcase of one previous problem---testcase \texttt{d\_tough\_choices.txt} of the 2020 Hash Code Qualification.} and similar.
    \item Note that we consider each of the provided inputs as a separate problem; where appropriate, different inputs to the same qualification round may feature different backbones.
\end{itemize}
This approach, in fact, closely mimics what a real competitor's workflow during Hash Code might look like; the main part being automated is the evolution of the scoring function, which is a hard task for humans. 

Note that the \emph{collaboration} between humans and AI was important to achieve the results with this workflow---the competitors can leverage their strengths and develop the backbone, whereas AI can automate the complex search in program space (likely to be necessary unless $\mathrm{P}=\mathrm{NP}$) that can extract higher returns out of the backbone.

\subsection{Evolving heuristics}

Next, we describe the methodology for evolving the scoring functions used for Hash Code.

\subsubsection{Introduction to FunSearch}

FunSearch \citep{funsearch} is an optimisation technique that pairs a pre-trained \emph{large language model} (LLM) with a \emph{systematic evaluator}. It is designed to address the limitations of LLMs, such as confabulations, through its usage of the evaluator to check the correctness of the LLM's output.

FunSearch operates by evolving a population of \emph{candidate programs} over multiple iterations.  The LLM proposes modifications or improvements to the programs, and the evaluator assesses the quality of these modifications. The process can in principle continue without termination, though in practice we would terminate it once either a desirable high-quality program is found, or no significant progress has been observed over a sufficient time frame.

The success of FunSearch is attributed to several key components, including \emph{best-shot prompting}---where the best-performing programs are fed back into prompts for the LLM to improve upon---and the use of \emph{program backbones} that focus the LLM's attention only on evolving critical program logic, leaving the backbone itself fixed.  Additionally, FunSearch employs an \emph{island-based} evolutionary method to maintain program diversity and avoid local optima. The asynchronous and parallel nature of FunSearch allows for efficient exploration of the program space, especially across multiple devices.

\subsubsection{Configuration specifics}

The setup and hyperparameters of our evolutionary program search largely match \citet{funsearch}. There are, however, three key differences, specifically tuned towards the competitive programming setting and latest developments in LLMs. We describe them here:
\begin{itemize}
    \item While \citet{funsearch} rely on a variant of PaLM 2 \citep{anil2023palm} specifically fine-tuned on code, we find that the code-writing capabilities of modern \emph{generalist fine-tuned LLMs} are sufficient to not require dedicated fine-tuning anymore. As such, we used Gemini 1.5 Flash 002 \citep{team2024gemini} as the LLM that proposes modifications to programs in the population.
    \item As combinatorial contests often feature very large inputs -- in order to make exhaustive search intractable -- we needed to increase the \emph{evaluation limits} for the generated programs from their defaults in \citet{funsearch}. Specifically, we limit memory usage of the entire program to $10$ GB, and wall-clock execution time to $1,800$ seconds.
    \item It is rather common that greedy solutions to combinatorial contests require \emph{multiple choice points}. This is in contention with the structure of \citet{funsearch}, which instead only optimises \emph{one} scoring function. We are able to practically work around this by introducing a \emph{switching variable} that the scoring function can branch on depending on which choice point needs to be scored at this point in time. This can be seen in Figure \ref{fig:score_base}, where the switching variable is \texttt{rate\_project} and it determines whether we're scoring a project or a project-role pair. The backbone takes care of invoking the scoring function with the correctly set switching.
\end{itemize}

\section{Results}

\begin{figure*}
    \centering
    \includegraphics[width=0.49\linewidth]{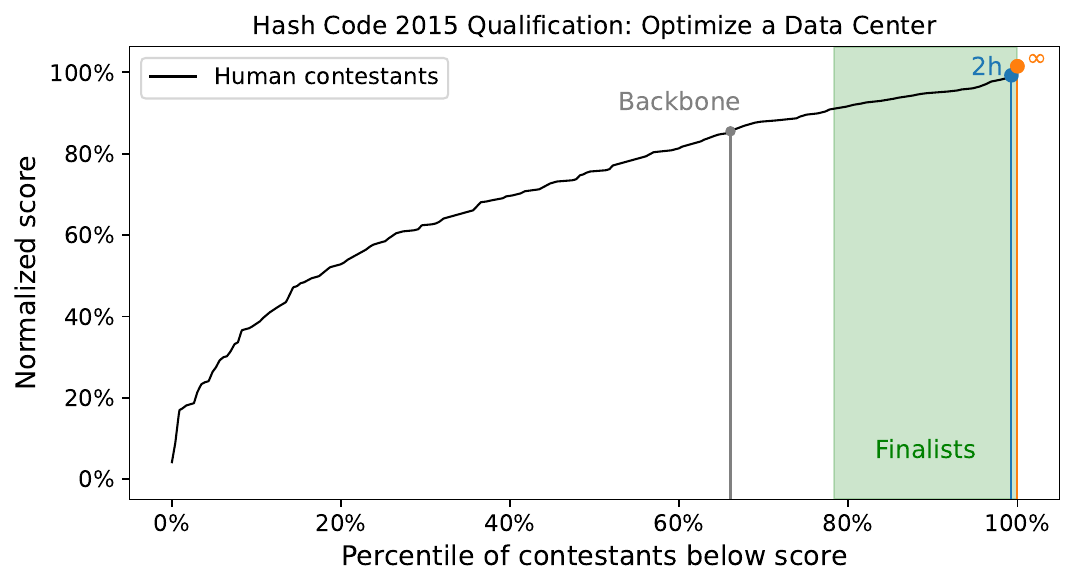} \hfill
    \includegraphics[width=0.49\linewidth]{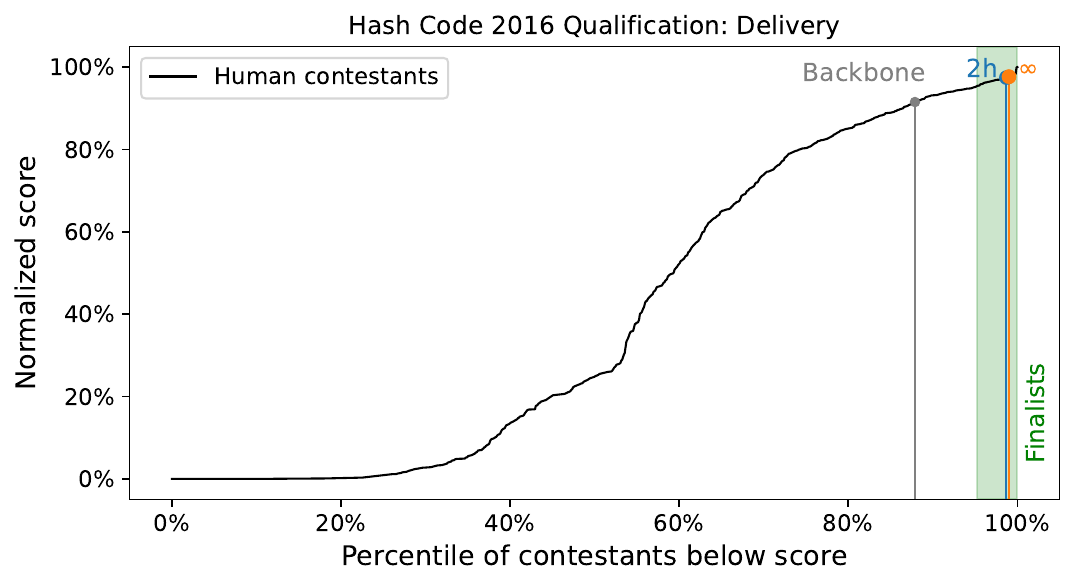}
    \includegraphics[width=0.49\linewidth]{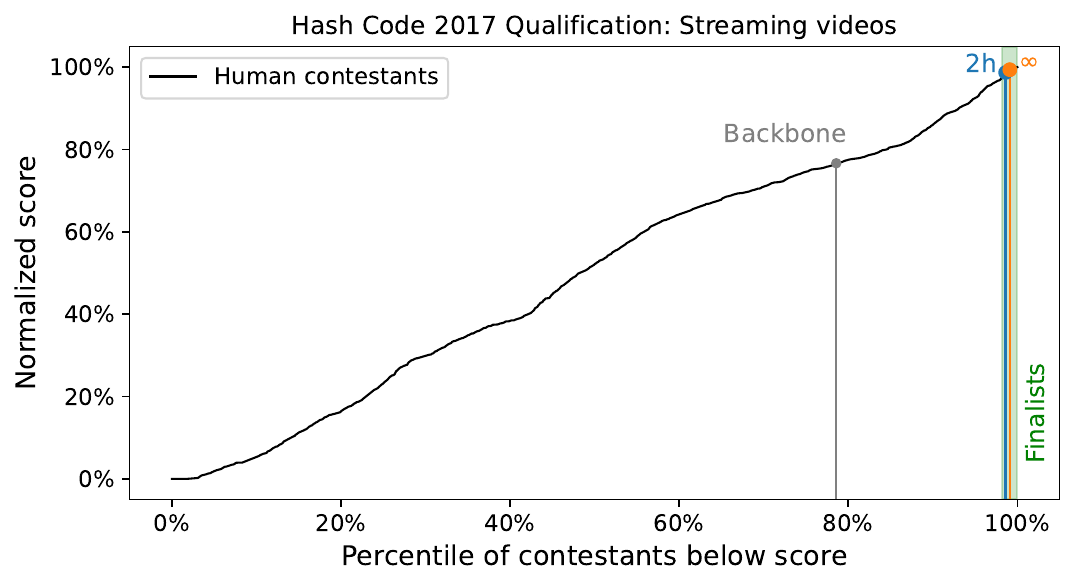} \hfill
    \includegraphics[width=0.49\linewidth]{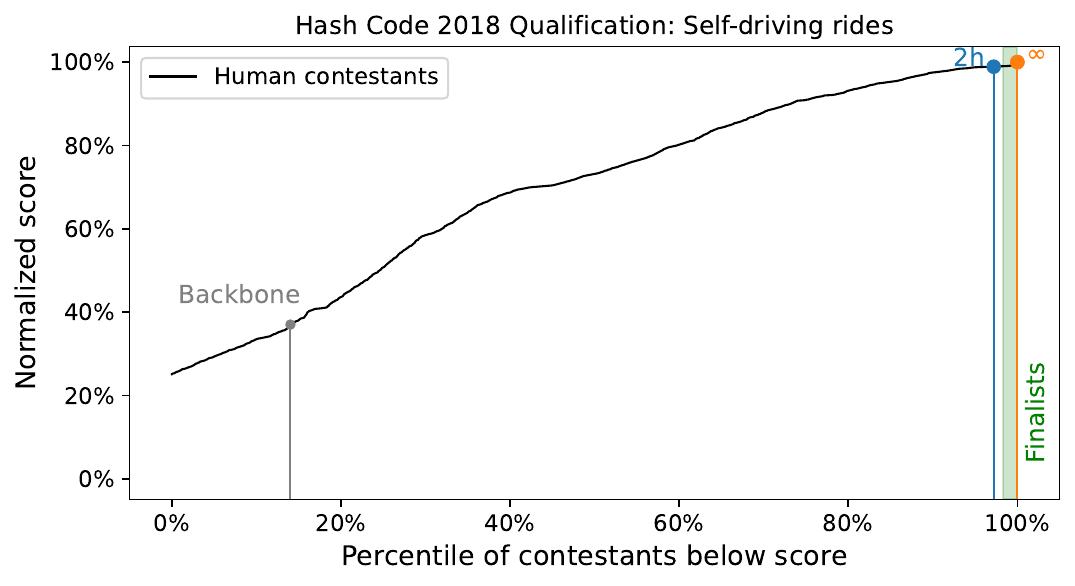} 
    \includegraphics[width=0.49\linewidth]{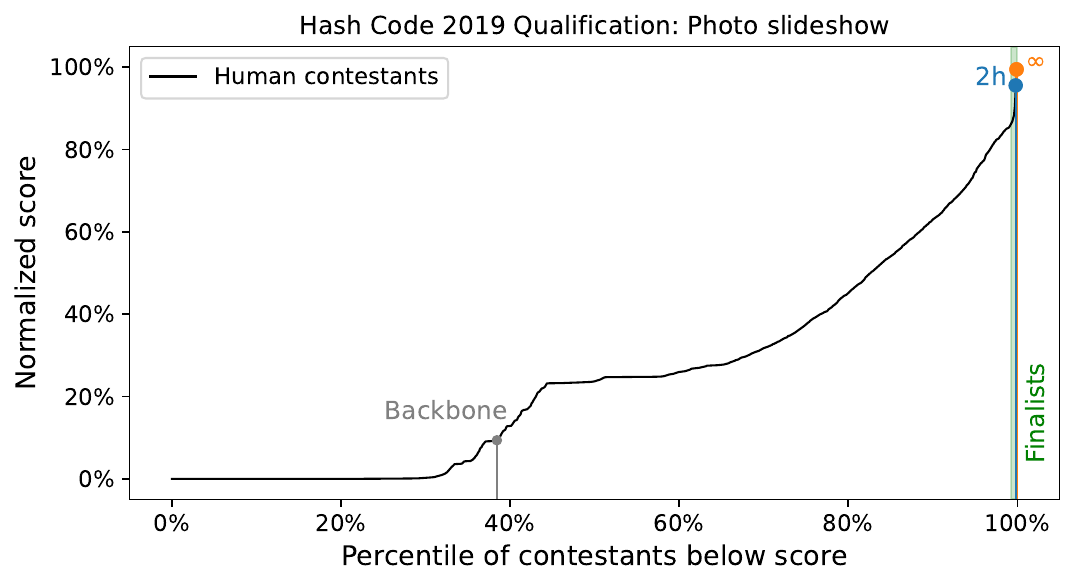}  \hfill
    \includegraphics[width=0.49\linewidth]{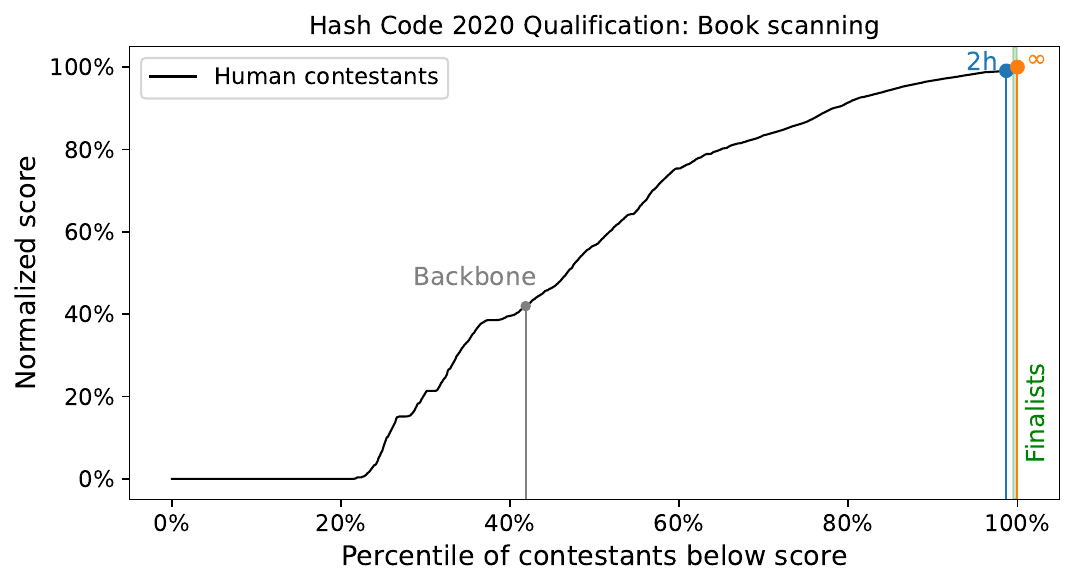}
    \includegraphics[width=0.49\linewidth]{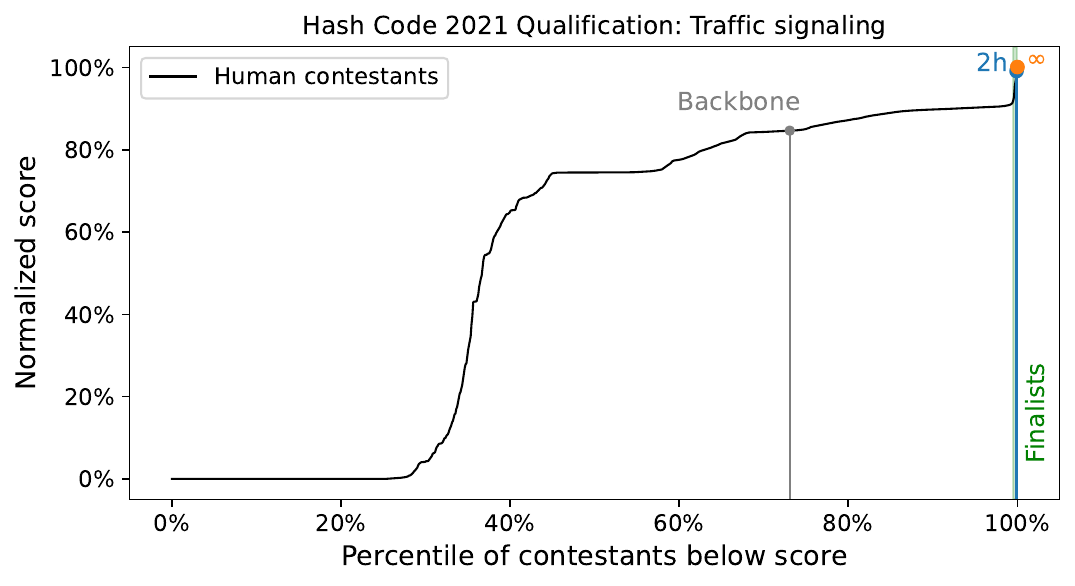}  \hfill
    \includegraphics[width=0.49\linewidth]{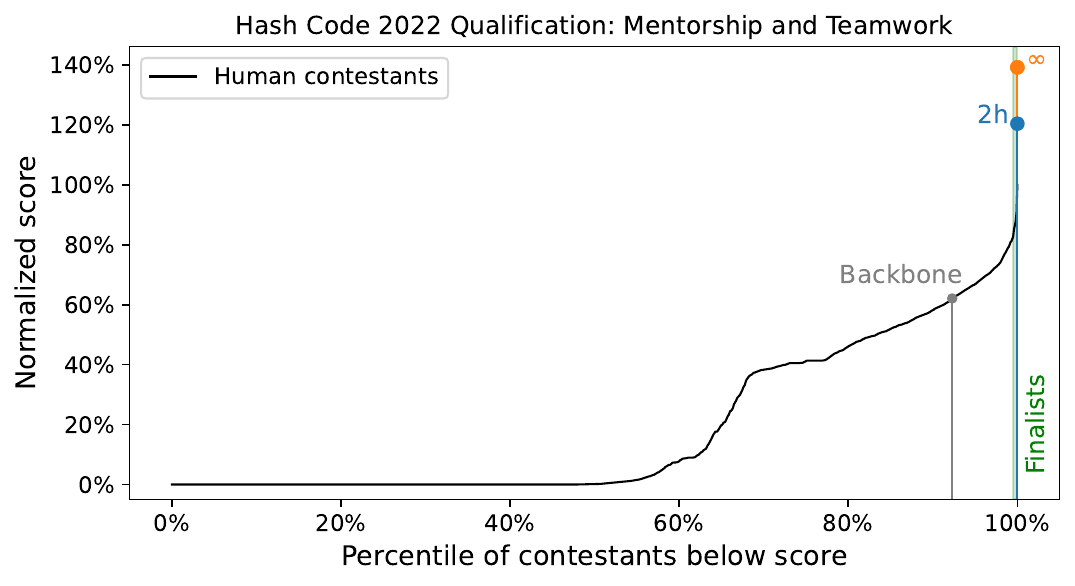}
    \caption{Rankings and scores of our backbone solutions with base scoring functions, and solutions evolved by FunSearch, across all eight Hash Code online qualification rounds. We plot the Hash Code fitness scores obtained by human competitor teams---normalised to the $[0, 1]$ range by dividing by the best team's score per contest---against the teams' rank in the contest. We then compute the fitness scores obtained by the backbone with base scoring function, as well as the best fitness we were able to achieve after evolving (as ``$\infty$'') and the fitness scores obtained after no more than two hours of evolving (as ``$2$h''). We report these scores on the ranking axis, and compare them against the ranks required to qualify into the finals. Our evolved solutions are consistently ranked in the top percentile, and outperform the top-scoring human team in five iterations (2015, 2018, 2020, 2021 and 2022).}
    \label{fig:my_label}
\end{figure*}

We deploy our approach on all eight Hash Code online qualification rounds (from 2015 until 2022). The main results, which we present in Figure \ref{fig:my_label}, will aim to quantify two key elements:

\noindent \textbf{Can FunSearch deliver meaningful improvements in the combinatorial competitive programming setting?} To measure this, we will compare the scores and ranks we obtain against the ones achieved by the backbone solution (with any iterative improvements applied) and our initial scoring function alone. We can assess the significance of this improvement by checking either the \emph{percentiles} achieved, or whether the recovered scores would be sufficient to qualify in the \emph{finals} (i.e. in the top 50 ranks\footnote{Note that this is an approximation---the exact number of finalist teams varied between 38 and 65 across the years.} that year). It is worth noting that, prior to 2019, Hash Code online qualification results comprised fewer than $5,000$ participating teams, meaning that achieving the top percentile was \emph{harder} than qualifying to the finals---afterwards it became \emph{easier}.

\noindent \textbf{Could these improvements be meaningfully obtained under contest conditions?} To measure this, we will also compare against the scores and ranks obtained after only running the evolutionary computation for \emph{two hours}. Given that Hash Code had a time limit of four hours, this leaves two hours for implementing the essential three parts of each backbone:
\begin{itemize}
    \item Parsing the input file into appropriate data class objects for further processing;
    \item A greedy algorithm with a base scoring function to produce candidate solutions;
    \item The systematic evaluator\footnote{While it may be argued that this part is not strictly necessary, as the contest environment already provides an evaluator, there are strict limits on how often solutions can be submitted to the evaluation server, and providing such an evaluator within the prompt allows FunSearch a significantly richer picture of how its solutions will be evaluated.} of candidate solutions, to be used as a fitness function;
\end{itemize}
Given that the combined backbone implementations we develop never exceeded $\sim400$ lines of Python code including docstrings (and were usually around $\sim200$ lines)---and Hash Code teams comprise up to four members---we find this to be a reasonable undertaking for a team of competent competitive programmers. As a reference, within two hours, our method evaluates around $10,500$ programs on average across all backbones.

\subsection{Result analysis and discussion}

The results outlined in Figure \ref{fig:my_label} provide conclusive evidence towards settling our two questions \emph{positively}. Specifically, we note that:
\begin{itemize}
    \item The backbones with base scoring functions are not capable of achieving the top percentile or a finals qualification in any Hash Code year we investigated;
    \item The evolved scoring functions make significant progress compared to the base scoring function in terms of both rank and fitness;
    \item The evolved functions reach the top percentile of competing teams, as well as surpassing the rank required to qualify in the finals, for every Hash Code iteration.
\end{itemize}
Beyond this, our evolved solutions were capable of outperforming the rank-1 team on \emph{five} online qualification rounds: 2015 (\emph{Optimize a Data Center}), 2018 (\emph{Self-driving rides}), 2020 (\emph{Book scanning}), 2021 (\emph{Traffic signaling}) and 2022 (\emph{Mentorship and Teamwork}). This is clearly a result that is beyond reach of the backbone developers (i.e. the authors of this paper) without AI assistance.

It is interesting to note that the starting point of our backbone solutions varies significantly across different Hash Code iterations. To give just two examples: in 2018, they start below the 20th percentile; while in 2022, they start above the 90th. This phenomenon nicely illustrates the diversity in challenge and contestant cohorts offered in various Hash Code iterations. The complexity of writing a working backbone may in and of itself pose quite a challenge---for example, writing an appropriate evaluator and greedy algorithm proved a significant undertaking in Hash Code 2016 (for reference, our own implementation has 415 lines of Python). Accordingly, having a working and complete backbone allowed for a relatively high rank compared to the competition cohort. 

Further, after Hash Code significantly rose in popularity---especially from 2019 onward---a longer tail of competitor teams with low relative scores started to emerge. This naturally uplifted the percentile of any backbone solution which achieves a modest relative score. For example, achieving a normalised score of $0.6$ would be insufficient even for the 30th perecentile in Hash Code 2018, while it is sufficient for the 90th percentile in Hash Code 2022.

We also believe that many of our method's benefits should be recoverable under contest conditions---considering the proximity of the ``two-hours'' solution to the maximal fitness we were able to obtain, and its relative merit against participating teams. In all but two iterations of Hash Code, this capped fitness would have been sufficient to qualify into the finals. Additionally, as we now know that sufficiently better solutions are within reach of the method given more computational resources, we consider that discovering them under the two-hour time constraint now amounts to an engineering challenge, rather than a research one. 

A possible limitation of our result is that, since the fitness on these contests was evaluated post-hoc rather than in real time, it is not unlikely that our base Gemini 1.5 Flash 002 model had been exposed to Hash Code subroutines within its training data. We believe this does not diminish the significance of our results, and elaborate further on why that is in the following paragraphs.

Firstly, unlike the context of code generation on Codeforces---where the model is asked to output the \emph{entirety} of the solution---in our collaborative setting the model is prompted to evolve \emph{only the scoring function}, in a way that is compatible to the rest of the backbone. This means that the solution needs to conform to the API prescribed by the backbone as well as utilising dataclasses defined within it, which is a prompt setting completely unseen in prior training, as our backbones are not available in LLM pre-training data.

Further, unlike the Codeforces setting, \emph{no information specific to the Hash Code tasks is given to the model in the prompt}---only the associated backbone code and previous best-performing scoring function, which makes it unlikely that a retrieval approach can even be successfully invoked in the first place. The fact that our models do not generate fully optimised solutions after one LLM mutation but rather tend to make steady, iterative progress towards improving solution fitness throughout training is further evidence to the fact that no final solutions are immediately recalled by the models. Indeed, most of our optimised solutions required a chain of at least 10--30 iterative improvement calls to Gemini 1.5 Flash 002.

In addition, towards the end of the paper we will provide a \emph{held-out contest case study}, on a variant of the recently held AtCoder Heuristic Contest 039 (which took place on 10 November 2024, \emph{after} the release of Gemini 1.5 Flash 002). This will serve to illustrate how our FunSearch-augmented method still yields tangible and significant benefits, even over a recent contest with a substantially different setup than Hash Code.

\section{Evolved Hash Code heuristics: qualitative case studies}

We will now present several case studies elaborating on the solutions evolved by FunSearch across several (sub)problem backbones in prior Hash Code qualification rounds. This analysis will both elucidate the insight discovered within some of the discovered solutions, and further emphasise the low likelihood that these solutions could have been derived through retrieval or recall.

\subsection{Qualification 2015: Optimizing a Data Center}

2015 marked the first Hash Code qualification round, where the task was to find the most fault-tolerant way to arrange servers within an abstractified data center.

Specifically, it is assumed that the data center has a certain number of rows, with each row having a certain number of slots (some of which may be blocked). Each server has a certain capacity, can be assigned to a particular pool, and occupies a certain contiguous number of slots. The task is to decide on locations where every server is installed, such that the \emph{guaranteed capacity}---the minimal remaining capacity across all of the pools, should any data center row stop working---is optimised.

There was exactly one input testcase, \texttt{hashcode\_2015\_qualification\_round.txt}, and we focus on it here. 
We provide the input parsing and greedy algorithm parts of the backbone we used in the Appendix (Figure \ref{fig:backbone_data}).

The greedy backbone operates in two phases: first, it determines where to place each server, then it assigns each server to a pool. The servers are iteratively assigned to each free space in a row, repeatedly cycling between rows to promote evenly distributed servers across rows.

Much like in Figure \ref{fig:score_base}, here we really need \emph{two} scoring functions: one to decide which server to place in the current position, and another to decide which pool to allocate a server to. We handle this using a boolean variable \texttt{rate\_server}. Our base scoring function rates servers based on their capacity relative to amount of spaces they occupy, and it rates pools based on how much would they additionally improve their pool without over-representing it in this particular row:

\begin{footnotesize}
\begin{Verbatim}[commandchars=\\\{\}]
\PYG{k}{if} \PYG{n}{rate\PYGZus{}server}\PYG{p}{:}
  \PYG{k}{return} \PYG{n}{server}\PYG{o}{.}\PYG{n}{capacity} \PYG{o}{/} \PYG{n}{server}\PYG{o}{.}\PYG{n}{size}
\PYG{k}{else}\PYG{p}{:}
  \PYG{n}{total\PYGZus{}sum} \PYG{o}{=} \PYG{l+m+mi}{0}
  \PYG{k}{for} \PYG{n}{c\PYGZus{}row} \PYG{o+ow}{in} \PYG{n}{pools\PYGZus{}per\PYGZus{}row}\PYG{p}{:}
    \PYG{n}{total\PYGZus{}sum} \PYG{o}{+=} \PYG{n}{pools\PYGZus{}per\PYGZus{}row}\PYG{p}{[}\PYG{n}{c\PYGZus{}row}\PYG{p}{][}\PYG{n}{pool}\PYG{p}{]}
  \PYG{k}{return} \PYG{o}{\PYGZhy{}}\PYG{n}{total\PYGZus{}sum} \PYG{o}{+} \PYG{n}{pools\PYGZus{}per\PYGZus{}row}\PYG{p}{[}\PYG{n}{row}\PYG{p}{][}\PYG{n}{pool}\PYG{p}{]}
\end{Verbatim}
\end{footnotesize}
This scoring function, while simple, achieves a commendable score of $348$ points. After evolving this function for two hours -- roughly remaining under contest conditions -- the optimal function becomes significantly more elaborate:

\begin{tiny}
\begin{Verbatim}[commandchars=\\\{\}]
\PYG{k}{assert} \PYG{n}{server}\PYG{o}{.}\PYG{n}{size} \PYG{o}{\PYGZgt{}} \PYG{l+m+mi}{0}
\PYG{k}{assert} \PYG{n}{server}\PYG{o}{.}\PYG{n}{capacity} \PYG{o}{\PYGZgt{}} \PYG{l+m+mi}{0}
\PYG{k}{if} \PYG{n}{pool} \PYG{o+ow}{is} \PYG{o+ow}{not} \PYG{k+kc}{None}\PYG{p}{:}
  \PYG{n}{total\PYGZus{}sum} \PYG{o}{=} \PYG{n}{pools\PYGZus{}per\PYGZus{}row}\PYG{p}{[}\PYG{n}{row}\PYG{p}{][}\PYG{n}{pool}\PYG{p}{]}
  \PYG{n}{max\PYGZus{}sum} \PYG{o}{=} \PYG{l+m+mi}{0}
  \PYG{n}{pool\PYGZus{}size} \PYG{o}{=} \PYG{l+m+mi}{0}
  \PYG{k}{for} \PYG{n}{p} \PYG{o+ow}{in} \PYG{n}{pools\PYGZus{}per\PYGZus{}row}\PYG{p}{:}
    \PYG{n}{total\PYGZus{}sum} \PYG{o}{+=} \PYG{n}{pools\PYGZus{}per\PYGZus{}row}\PYG{p}{[}\PYG{n}{p}\PYG{p}{][}\PYG{n}{pool}\PYG{p}{]}
    \PYG{n}{max\PYGZus{}sum} \PYG{o}{=} \PYG{n+nb}{max}\PYG{p}{(}\PYG{n}{pools\PYGZus{}per\PYGZus{}row}\PYG{p}{[}\PYG{n}{p}\PYG{p}{][}\PYG{n}{pool}\PYG{p}{],} \PYG{n}{max\PYGZus{}sum}\PYG{p}{)}
    \PYG{k}{if} \PYG{n}{p} \PYG{o}{==} \PYG{n}{pool}\PYG{p}{:}
      \PYG{n}{pool\PYGZus{}size} \PYG{o}{+=} \PYG{n}{pools\PYGZus{}per\PYGZus{}row}\PYG{p}{[}\PYG{n}{p}\PYG{p}{][}\PYG{n}{pool}\PYG{p}{]}
  \PYG{k}{if} \PYG{n}{total\PYGZus{}sum} \PYG{o}{==} \PYG{l+m+mi}{0}\PYG{p}{:}
    \PYG{k}{return} \PYG{o}{\PYGZhy{}}\PYG{l+m+mi}{100}
  \PYG{c+c1}{\PYGZsh{} Compute max gain after adding 1 server in row:}
  \PYG{n}{row\PYGZus{}score} \PYG{o}{=} \PYG{o}{\PYGZhy{}}\PYG{n}{total\PYGZus{}sum} \PYG{o}{/} \PYG{n}{server}\PYG{o}{.}\PYG{n}{size} \PYG{o}{+} \PYG{n}{max\PYGZus{}sum} \PYG{o}{/} \PYG{n}{server}\PYG{o}{.}\PYG{n}{size}
  \PYG{k}{if} \PYG{n}{row} \PYG{o+ow}{not} \PYG{o+ow}{in} \PYG{n}{pools\PYGZus{}per\PYGZus{}row}\PYG{p}{:}
    \PYG{k}{return} \PYG{n}{row\PYGZus{}score}
  \PYG{k}{else}\PYG{p}{:}
    \PYG{n}{pool\PYGZus{}score} \PYG{o}{=} \PYG{o}{\PYGZhy{}}\PYG{l+m+mf}{0.5} \PYG{o}{*} \PYG{n}{total\PYGZus{}sum} \PYG{o}{/} \PYG{n}{server}\PYG{o}{.}\PYG{n}{size} \PYG{o}{+} \PYG{p}{(}
        \PYG{l+m+mf}{0.5} \PYG{o}{*} \PYG{n}{max\PYGZus{}sum} \PYG{o}{/} \PYG{n}{server}\PYG{o}{.}\PYG{n}{size}\PYG{p}{)}
    \PYG{n}{pool\PYGZus{}bonus} \PYG{o}{=} \PYG{l+m+mf}{0.015} \PYG{o}{*} \PYG{p}{(}\PYG{n}{total\PYGZus{}sum} \PYG{o}{\PYGZhy{}} \PYG{n}{pool\PYGZus{}size}\PYG{p}{)}
    \PYG{k}{if} \PYG{n}{server}\PYG{o}{.}\PYG{n}{capacity} \PYG{o}{\PYGZgt{}=} \PYG{n}{total\PYGZus{}sum} \PYG{o}{/} \PYG{l+m+mf}{2.0}\PYG{p}{:}
      \PYG{n}{pool\PYGZus{}bonus} \PYG{o}{*=} \PYG{l+m+mf}{1.2}
    \PYG{k}{elif} \PYG{n}{server}\PYG{o}{.}\PYG{n}{capacity} \PYG{o}{\PYGZgt{}=} \PYG{p}{(}\PYG{n}{total\PYGZus{}sum} \PYG{o}{\PYGZhy{}} \PYG{n}{pool\PYGZus{}size}\PYG{p}{)} \PYG{o}{/} \PYG{l+m+mf}{4.0}\PYG{p}{:}
      \PYG{n}{pool\PYGZus{}bonus} \PYG{o}{*=} \PYG{l+m+mf}{1.5}
    \PYG{k}{return} \PYG{p}{(}\PYG{n}{row\PYGZus{}score} \PYG{o}{+} \PYG{n}{pool\PYGZus{}score} \PYG{o}{+} \PYG{n}{pool\PYGZus{}bonus} \PYG{o}{/} \PYG{l+m+mf}{1000.} \PYG{o}{+} \PYG{l+m+mf}{0.00005} \PYG{o}{*}
            \PYG{p}{(}\PYG{n}{server}\PYG{o}{.}\PYG{n}{capacity} \PYG{o}{/} \PYG{n}{server}\PYG{o}{.}\PYG{n}{size} \PYG{o}{+} \PYG{n}{total\PYGZus{}sum} \PYG{o}{/} \PYG{n+nb}{min}\PYG{p}{(}
                \PYG{n}{total\PYGZus{}sum}\PYG{p}{,} \PYG{n}{server}\PYG{o}{.}\PYG{n}{capacity} \PYG{o}{*} \PYG{l+m+mf}{1.1}\PYG{p}{))} \PYG{o}{\PYGZhy{}} \PYG{l+m+mf}{0.004} \PYG{o}{*} \PYG{n}{row} \PYG{o}{/}
            \PYG{n+nb}{len}\PYG{p}{(}\PYG{n}{pools\PYGZus{}per\PYGZus{}row}\PYG{p}{)} \PYG{o}{*} \PYG{p}{(}
                \PYG{n}{server}\PYG{o}{.}\PYG{n}{size} \PYG{o}{\PYGZgt{}=} \PYG{n}{total\PYGZus{}sum} \PYG{o}{/} \PYG{l+m+mf}{10.0}\PYG{p}{)} \PYG{o}{\PYGZhy{}} \PYG{l+m+mf}{0.0004} \PYG{o}{*} \PYG{n}{pool} \PYG{o}{/}
            \PYG{n+nb}{len}\PYG{p}{(}\PYG{n}{pools\PYGZus{}per\PYGZus{}row}\PYG{p}{)} \PYG{o}{\PYGZhy{}} \PYG{l+m+mf}{0.0007} \PYG{o}{*} \PYG{n}{server}\PYG{o}{.}\PYG{n}{size} \PYG{o}{/} \PYG{n+nb}{len}\PYG{p}{(}
                \PYG{n}{pools\PYGZus{}per\PYGZus{}row}\PYG{p}{)} \PYG{o}{*} \PYG{p}{(}\PYG{n}{server}\PYG{o}{.}\PYG{n}{capacity} \PYG{o}{\PYGZgt{}=} \PYG{p}{(}
                    \PYG{n}{total\PYGZus{}sum} \PYG{o}{/} \PYG{l+m+mf}{2.0}\PYG{p}{)} \PYG{o}{*} \PYG{l+m+mf}{1.005}\PYG{p}{))}
\PYG{k}{elif} \PYG{n}{rate\PYGZus{}server}\PYG{p}{:}
  \PYG{k}{return} \PYG{n}{server}\PYG{o}{.}\PYG{n}{capacity} \PYG{o}{/} \PYG{n}{server}\PYG{o}{.}\PYG{n}{size} \PYG{o}{*} \PYG{p}{(}
      \PYG{l+m+mf}{2.0} \PYG{o}{+} \PYG{l+m+mf}{2.0} \PYG{o}{*} \PYG{n}{server}\PYG{o}{.}\PYG{n}{size} \PYG{o}{/} \PYG{l+m+mf}{3.0}\PYG{p}{)}
\PYG{k}{else}\PYG{p}{:}
  \PYG{k}{raise} \PYG{n+ne}{ValueError}\PYG{p}{(}\PYG{l+s+s1}{\PYGZsq{}should not call this function with None pool\PYGZsq{}}\PYG{p}{)}
\end{Verbatim}
\end{tiny}
This scoring function significantly improves the score to $405$, sufficient to achieve \emph{second place}! This solution is also substantially mutated, requiring a chain of $15$ LLM calls starting from the original. This implementation supports an elaborate algebraic expression for a score function with evolved constants assigned to each component. It is also interesting to note that the evolved function inserts assertions and throws exceptions that would trigger under invalid inputs---though such inputs will never appear in practice.

Finally, it is interesting to make note of the \texttt{return -100} statement which, while not likely to make a big difference in this particular setting (it is only invoked when a pool has no servers assigned to it yet), it sets up the model for a more peculiar kind of mechanism to be discovered later.

If we allow for more time for evolving the scoring function, it is possible to achieve a \textbf{rank-1} score of $413$, with the following snippet:

\begin{tiny}
\begin{Verbatim}[commandchars=\\\{\}]
\PYG{k}{assert} \PYG{n}{row} \PYG{o}{\PYGZgt{}=} \PYG{l+m+mi}{0}
\PYG{n}{cap} \PYG{o}{=} \PYG{l+m+mi}{0}
\PYG{n}{prev\PYGZus{}row} \PYG{o}{=} \PYG{n}{row} \PYG{o}{\PYGZhy{}} \PYG{l+m+mi}{1}
\PYG{k}{if} \PYG{n}{rate\PYGZus{}server}\PYG{p}{:}
  \PYG{k}{if} \PYG{n}{server}\PYG{o}{.}\PYG{n}{size} \PYG{o}{\PYGZgt{}} \PYG{l+m+mi}{125}\PYG{p}{:}
    \PYG{k}{return} \PYG{o}{\PYGZhy{}}\PYG{l+m+mi}{100}
  \PYG{k}{else}\PYG{p}{:}
    \PYG{k}{if} \PYG{n}{server}\PYG{o}{.}\PYG{n}{size} \PYG{o}{\PYGZgt{}} \PYG{l+m+mi}{23}\PYG{p}{:}
      \PYG{n}{cap} \PYG{o}{=} \PYG{l+m+mf}{0.5}
    \PYG{n}{cap} \PYG{o}{+=} \PYG{p}{(}\PYG{l+m+mf}{2.7} \PYG{o}{\PYGZhy{}} \PYG{l+m+mf}{4.95} \PYG{o}{*} \PYG{n}{server}\PYG{o}{.}\PYG{n}{size} \PYG{o}{/} \PYG{l+m+mi}{125}\PYG{p}{)} \PYG{o}{*} \PYG{p}{(}
        \PYG{n}{server}\PYG{o}{.}\PYG{n}{capacity} \PYG{o}{/} \PYG{l+m+mi}{125}\PYG{p}{)}
    \PYG{k}{if} \PYG{n}{server}\PYG{o}{.}\PYG{n}{capacity} \PYG{o}{/} \PYG{n}{server}\PYG{o}{.}\PYG{n}{size} \PYG{o}{\PYGZgt{}} \PYG{l+m+mf}{7.5}\PYG{p}{:}
      \PYG{n}{cap} \PYG{o}{*=} \PYG{l+m+mf}{5.0}
    \PYG{k}{if} \PYG{n}{server}\PYG{o}{.}\PYG{n}{capacity} \PYG{o}{/} \PYG{n}{server}\PYG{o}{.}\PYG{n}{size} \PYG{o}{\PYGZgt{}} \PYG{l+m+mf}{7.95}\PYG{p}{:}
      \PYG{n}{cap} \PYG{o}{*=} \PYG{l+m+mf}{11.0}
    \PYG{k}{return} \PYG{n}{cap}
\PYG{k}{else}\PYG{p}{:}
  \PYG{n}{n\PYGZus{}pools\PYGZus{}full} \PYG{o}{=} \PYG{n+nb}{sum}\PYG{p}{(}\PYG{l+m+mi}{1} \PYG{k}{if} \PYG{n}{pool\PYGZus{}cap} \PYG{o}{\PYGZgt{}} \PYG{l+m+mi}{7357} \PYG{k}{else} \PYG{p}{(}
      \PYG{l+m+mi}{0} \PYG{k}{for} \PYG{n}{pool\PYGZus{}cap} \PYG{o+ow}{in} \PYG{n}{pools\PYGZus{}per\PYGZus{}row}\PYG{p}{[}\PYG{n}{row}\PYG{p}{]}\PYG{o}{.}\PYG{n}{values}\PYG{p}{()))}
  \PYG{k}{assert} \PYG{n}{pool} \PYG{o+ow}{is} \PYG{o+ow}{not} \PYG{k+kc}{None}
  \PYG{n}{max\PYGZus{}cap} \PYG{o}{=} \PYG{n+nb}{max}\PYG{p}{(}\PYG{n}{pools\PYGZus{}per\PYGZus{}row}\PYG{p}{[}\PYG{n}{row}\PYG{p}{][}\PYG{n}{pool}\PYG{p}{]} \PYG{o}{/} \PYG{l+m+mi}{1150}\PYG{p}{,} \PYG{l+m+mf}{0.475}\PYG{p}{)}
  \PYG{k}{assert} \PYG{n}{max\PYGZus{}cap} \PYG{o}{\PYGZgt{}} \PYG{l+m+mi}{0}
  \PYG{n}{total\PYGZus{}cap} \PYG{o}{=} \PYG{l+m+mf}{1.16} \PYG{o}{\PYGZhy{}} \PYG{p}{(}\PYG{l+m+mf}{1.16} \PYG{o}{\PYGZhy{}} \PYG{l+m+mf}{1.1}\PYG{p}{)} \PYG{o}{*} \PYG{p}{(}\PYG{l+m+mf}{0.9} \PYG{o}{\PYGZhy{}} \PYG{n}{n\PYGZus{}pools\PYGZus{}full} \PYG{o}{/} \PYG{l+m+mi}{5}\PYG{p}{)}
  \PYG{n}{min\PYGZus{}cap} \PYG{o}{=} \PYG{l+m+mf}{13000.0}
  \PYG{k}{for} \PYG{n}{c\PYGZus{}row}\PYG{p}{,} \PYG{n}{pool\PYGZus{}cap} \PYG{o+ow}{in} \PYG{n+nb}{sorted}\PYG{p}{(}\PYG{n}{pools\PYGZus{}per\PYGZus{}row}\PYG{o}{.}\PYG{n}{items}\PYG{p}{()):}
    \PYG{n}{total\PYGZus{}cap} \PYG{o}{+=} \PYG{n+nb}{max}\PYG{p}{(}\PYG{n}{pool\PYGZus{}cap}\PYG{p}{[}\PYG{n}{pool}\PYG{p}{],} \PYG{l+m+mf}{0.1}\PYG{p}{)}
    \PYG{k}{if} \PYG{n}{prev\PYGZus{}row} \PYG{o+ow}{is} \PYG{o+ow}{not} \PYG{k+kc}{None} \PYG{o+ow}{and} \PYG{p}{(}
        \PYG{n}{c\PYGZus{}row} \PYG{o}{!=} \PYG{n}{row}\PYG{p}{)} \PYG{o+ow}{and} \PYG{p}{(}\PYG{n}{c\PYGZus{}row} \PYG{o}{!=} \PYG{n}{prev\PYGZus{}row}\PYG{p}{):}
      \PYG{k}{assert} \PYG{n}{c\PYGZus{}row} \PYG{o+ow}{in} \PYG{n}{pools\PYGZus{}per\PYGZus{}row}
      \PYG{k}{assert} \PYG{n}{pool} \PYG{o+ow}{in} \PYG{n}{pools\PYGZus{}per\PYGZus{}row}\PYG{p}{[}\PYG{n}{c\PYGZus{}row}\PYG{p}{]}
      \PYG{n}{max\PYGZus{}cap} \PYG{o}{=} \PYG{n+nb}{max}\PYG{p}{(}\PYG{n}{max\PYGZus{}cap}\PYG{p}{,} \PYG{n}{pools\PYGZus{}per\PYGZus{}row}\PYG{p}{[}\PYG{n}{c\PYGZus{}row}\PYG{p}{][}\PYG{n}{pool}\PYG{p}{])}
      \PYG{n}{min\PYGZus{}cap} \PYG{o}{=} \PYG{n+nb}{min}\PYG{p}{(}\PYG{n}{min\PYGZus{}cap}\PYG{p}{,} \PYG{n}{pools\PYGZus{}per\PYGZus{}row}\PYG{p}{[}\PYG{n}{c\PYGZus{}row}\PYG{p}{][}\PYG{n}{pool}\PYG{p}{])}
    \PYG{n}{prev\PYGZus{}row} \PYG{o}{=} \PYG{n}{c\PYGZus{}row}
    \PYG{k}{if} \PYG{n}{min\PYGZus{}cap} \PYG{o}{\PYGZgt{}} \PYG{n}{server}\PYG{o}{.}\PYG{n}{capacity}\PYG{p}{:}
      \PYG{n}{min\PYGZus{}cap} \PYG{o}{=} \PYG{n}{server}\PYG{o}{.}\PYG{n}{capacity} \PYG{o}{*} \PYG{l+m+mf}{0.7}

  \PYG{n}{total\PYGZus{}cap} \PYG{o}{+=} \PYG{n+nb}{max}\PYG{p}{(}\PYG{n}{pools\PYGZus{}per\PYGZus{}row}\PYG{p}{[}\PYG{n}{row}\PYG{p}{][}\PYG{n}{pool}\PYG{p}{]} \PYG{o}{*} \PYG{l+m+mf}{0.95}\PYG{p}{,} \PYG{l+m+mf}{0.03}\PYG{p}{)}
  \PYG{n}{total\PYGZus{}cap} \PYG{o}{\PYGZhy{}=} \PYG{n+nb}{max}\PYG{p}{(}\PYG{n}{pools\PYGZus{}per\PYGZus{}row}\PYG{p}{[}\PYG{n}{row}\PYG{p}{][}\PYG{n}{pool}\PYG{p}{]} \PYG{o}{/} \PYG{l+m+mi}{650}\PYG{p}{,} \PYG{l+m+mf}{0.005}\PYG{p}{)}
  \PYG{n}{cap} \PYG{o}{+=} \PYG{n+nb}{max}\PYG{p}{(}\PYG{n+nb}{max}\PYG{p}{(}
      \PYG{n}{pools\PYGZus{}per\PYGZus{}row}\PYG{p}{[}\PYG{n}{row}\PYG{p}{][}\PYG{n}{pool}\PYG{p}{],} \PYG{l+m+mf}{0.1}\PYG{p}{)} \PYG{o}{/} \PYG{l+m+mf}{1.32}\PYG{p}{,} \PYG{l+m+mf}{710.0} \PYG{o}{/} \PYG{p}{(}
          \PYG{n}{server}\PYG{o}{.}\PYG{n}{size} \PYG{o}{+} \PYG{l+m+mf}{1.0}\PYG{p}{))}
  \PYG{k}{assert} \PYG{n}{cap} \PYG{o}{\PYGZgt{}} \PYG{l+m+mi}{0}
  \PYG{k}{if} \PYG{n}{server}\PYG{o}{.}\PYG{n}{size} \PYG{o}{\PYGZgt{}} \PYG{n}{max\PYGZus{}cap}\PYG{p}{:}
    \PYG{k}{return} \PYG{o}{\PYGZhy{}}\PYG{l+m+mi}{100}
  \PYG{k}{elif} \PYG{n}{total\PYGZus{}cap} \PYG{o}{\PYGZlt{}} \PYG{n}{server}\PYG{o}{.}\PYG{n}{capacity}\PYG{p}{:}
    \PYG{k}{return} \PYG{o}{\PYGZhy{}}\PYG{l+m+mi}{10000}
  \PYG{k}{else}\PYG{p}{:}
    \PYG{k}{return} \PYG{p}{(}\PYG{n}{server}\PYG{o}{.}\PYG{n}{capacity} \PYG{o}{\PYGZhy{}} \PYG{p}{(}\PYG{n}{total\PYGZus{}cap} \PYG{o}{\PYGZhy{}} \PYG{n}{max\PYGZus{}cap}\PYG{p}{))} \PYG{o}{+} \PYG{n}{cap}
\end{Verbatim}
\end{tiny}
This function also implements an elaborate algebraic score expression with evolved constants. However, it is interesting to note how the negative numbers are being used in this case. For example, with the check \texttt{if server.size > 125: return -100}, the system is implementing a hard threshold which automatically discourages from scheduling all servers occupying more than $125$ slots. Similarly, with the final \texttt{if/elif/else} chain, it is discouraged to allocate large servers to any particular pool (\texttt{if}), and even more so to allocate servers to a pool if they have substantially larger capacity than a computed cap value (\texttt{elif}).

\subsection{Qualification 2018: \emph{Self-driving rides}}

Next, we analyse the evolution of the scoring function for the 2018 Hash Code qualification round, where the objective is to find an optimal schedule of rides for a fleet of self-driving cars.

Specifically, we are given a fleet of a certain number of self-driving cars, which need to complete a certain amount of ordered rides. Each ride order consists of a start and end point, earliest start time and the latest finish time. Each ride completed on time accrues a certain number of points, with a certain bonus obtained for rides that start in the earliest possible moment. The task is to decide a schedule in which each of the cars will complete a certain list of rides sequentially, always driving along the shortest path to their next destination.

We focus on the testcase \texttt{d\_metropolis.in}, where the model made the most significant jump in score compared to the backbone. As before, in the Appendix we provide the key parts of the backbone we used for this input (Figure \ref{fig:backbone_self}).

The backbone greedy solution in this case maintains a priority queue of all cars in the fleet, keyed by the time at which they finish their latest ride (initially zero for all cars). At each step, the top car in the priority queue is taken, and a ride is allocated to it. Our base scoring function simply picks the first feasible ride:

\begin{footnotesize}
\begin{Verbatim}[commandchars=\\\{\}]
\PYG{k}{for} \PYG{n}{i}\PYG{p}{,} \PYG{n}{r} \PYG{o+ow}{in} \PYG{n+nb}{enumerate}\PYG{p}{(}\PYG{n}{rides}\PYG{p}{):}
  \PYG{n}{pickup\PYGZus{}time} \PYG{o}{=} \PYG{n}{time} \PYG{o}{+} \PYG{n}{r}\PYG{o}{.}\PYG{n}{distance\PYGZus{}to\PYGZus{}start}\PYG{p}{(}
        \PYG{n}{coords}\PYG{p}{)}
  \PYG{k}{if} \PYG{p}{(}\PYG{n}{pickup\PYGZus{}time} \PYG{o}{\PYGZgt{}=} \PYG{n}{r}\PYG{o}{.}\PYG{n}{earliest\PYGZus{}start}\PYG{p}{)} \PYG{o+ow}{and} \PYG{p}{(}
      \PYG{n}{pickup\PYGZus{}time} \PYG{o}{+} \PYG{p}{(}
          \PYG{n}{r}\PYG{o}{.}\PYG{n}{length}\PYG{p}{())} \PYG{o}{\PYGZlt{}} \PYG{n}{r}\PYG{o}{.}\PYG{n}{latest\PYGZus{}finish}
  \PYG{p}{):}
    \PYG{k}{return} \PYG{n}{i}
\PYG{c+c1}{\PYGZsh{} We failed to find any feasible ride.}
\PYG{k}{return} \PYG{o}{\PYGZhy{}}\PYG{l+m+mi}{1}
\end{Verbatim}
\end{footnotesize}
and it achieves $3,528,556$ points on this testcase. After evolving for two hours, we recover the following intuitive solution:

\begin{tiny}
\begin{Verbatim}[commandchars=\\\{\}]
\PYG{n}{best\PYGZus{}time}\PYG{p}{,} \PYG{n}{best\PYGZus{}idx} \PYG{o}{=} \PYG{o}{\PYGZhy{}}\PYG{l+m+mi}{1}\PYG{p}{,} \PYG{o}{\PYGZhy{}}\PYG{l+m+mi}{1}
\PYG{k}{for} \PYG{n}{i}\PYG{p}{,} \PYG{n}{r} \PYG{o+ow}{in} \PYG{n+nb}{enumerate}\PYG{p}{(}\PYG{n}{rides}\PYG{p}{):}
  \PYG{n}{pickup\PYGZus{}time} \PYG{o}{=} \PYG{n+nb}{max}\PYG{p}{(}
      \PYG{n}{r}\PYG{o}{.}\PYG{n}{earliest\PYGZus{}start}\PYG{p}{,} \PYG{n}{time} \PYG{o}{+} \PYG{n}{r}\PYG{o}{.}\PYG{n}{distance\PYGZus{}to\PYGZus{}start}\PYG{p}{(}\PYG{n}{coords}\PYG{p}{))}
  \PYG{k}{if} \PYG{n}{pickup\PYGZus{}time} \PYG{o}{+} \PYG{n}{r}\PYG{o}{.}\PYG{n}{length}\PYG{p}{()} \PYG{o}{\PYGZgt{}=} \PYG{n}{r}\PYG{o}{.}\PYG{n}{latest\PYGZus{}finish}\PYG{p}{:}
    \PYG{k}{continue}
  \PYG{k}{if} \PYG{n}{best\PYGZus{}time} \PYG{o}{\PYGZlt{}} \PYG{l+m+mi}{0} \PYG{o+ow}{or} \PYG{n}{pickup\PYGZus{}time} \PYG{o}{\PYGZlt{}} \PYG{n}{best\PYGZus{}time}\PYG{p}{:}
    \PYG{n}{best\PYGZus{}time}\PYG{p}{,} \PYG{n}{best\PYGZus{}idx} \PYG{o}{=} \PYG{n}{pickup\PYGZus{}time}\PYG{p}{,} \PYG{n}{i}
\PYG{k}{return} \PYG{n}{best\PYGZus{}idx}
\end{Verbatim}
\end{tiny}
which corrects for the fact that the pick-up time doesn't have to be \emph{after} the earliest start to be feasible (the car simply has to wait), and the loop does not break early when a ride is found---instead, the best solution is retrieved based on its proximity to the car's present point.

This solution amplifies the performance on this input significantly, to $11,739,630$ points.

From this point, a significant amount of time and additional evolution steps are needed to discover a useful non-trivial solution. 

\begin{tiny}
\begin{Verbatim}[commandchars=\\\{\}]
\PYG{n}{best\PYGZus{}score} \PYG{o}{=} \PYG{n+nb}{float}\PYG{p}{(}\PYG{l+s+s2}{\PYGZdq{}\PYGZhy{}inf\PYGZdq{}}\PYG{p}{)}
\PYG{n}{best\PYGZus{}ride} \PYG{o}{=} \PYG{o}{\PYGZhy{}}\PYG{l+m+mi}{1}
\PYG{n}{free\PYGZus{}time} \PYG{o}{=} \PYG{l+m+mi}{0}

\PYG{c+c1}{\PYGZsh{} Rides in descending order by distance to the starting point.}
\PYG{n}{rides\PYGZus{}by\PYGZus{}length} \PYG{o}{=} \PYG{p}{[(}\PYG{n}{i}\PYG{p}{,} \PYG{n}{r}\PYG{o}{.}\PYG{n}{length}\PYG{p}{(),} \PYG{n}{distance}\PYG{p}{(}\PYG{n}{coords}\PYG{p}{,} \PYG{n}{r}\PYG{o}{.}\PYG{n}{start}\PYG{p}{))}
                   \PYG{k}{for} \PYG{n}{i}\PYG{p}{,} \PYG{n}{r} \PYG{o+ow}{in} \PYG{n+nb}{enumerate}\PYG{p}{(}\PYG{n}{rides}\PYG{p}{)}
                   \PYG{k}{if} \PYG{n}{r}\PYG{o}{.}\PYG{n}{latest\PYGZus{}finish} \PYG{o}{\PYGZgt{}=} \PYG{n}{time} \PYG{o}{//} \PYG{l+m+mf}{2.}\PYG{p}{]}

\PYG{n}{rides\PYGZus{}by\PYGZus{}length}\PYG{o}{.}\PYG{n}{sort}\PYG{p}{(}\PYG{n}{reverse}\PYG{o}{=}\PYG{k+kc}{True}\PYG{p}{)}

\PYG{k}{for} \PYG{p}{(}\PYG{n}{i}\PYG{p}{,} \PYG{n}{ride\PYGZus{}length}\PYG{p}{,} \PYG{n}{distance\PYGZus{}to\PYGZus{}start}\PYG{p}{)} \PYG{o+ow}{in} \PYG{n}{rides\PYGZus{}by\PYGZus{}length}\PYG{p}{:}
  \PYG{n}{r} \PYG{o}{=} \PYG{n}{rides}\PYG{p}{[}\PYG{n}{i}\PYG{p}{]}

  \PYG{n}{pickup\PYGZus{}time} \PYG{o}{=} \PYG{n}{time} \PYG{o}{+} \PYG{n}{distance\PYGZus{}to\PYGZus{}start}
  \PYG{k}{if} \PYG{n}{pickup\PYGZus{}time} \PYG{o}{\PYGZlt{}} \PYG{n}{r}\PYG{o}{.}\PYG{n}{earliest\PYGZus{}start}\PYG{p}{:}
    \PYG{n}{pickup\PYGZus{}time} \PYG{o}{=} \PYG{n}{r}\PYG{o}{.}\PYG{n}{earliest\PYGZus{}start}
  \PYG{n}{free\PYGZus{}time} \PYG{o}{=} \PYG{n}{pickup\PYGZus{}time} \PYG{o}{+} \PYG{n}{ride\PYGZus{}length}

  \PYG{n}{bonus\PYGZus{}points} \PYG{o}{=} \PYG{l+m+mi}{20000} \PYG{k}{if} \PYG{n}{time} \PYG{o}{\PYGZlt{}} \PYG{l+m+mi}{3600} \PYG{o+ow}{or} \PYG{n}{time} \PYG{o}{\PYGZgt{}} \PYG{l+m+mi}{20900} \PYG{k}{else} \PYG{l+m+mi}{0}
  \PYG{n}{bonus\PYGZus{}points} \PYG{o}{=} \PYG{n}{bonus\PYGZus{}points} \PYG{k}{if} \PYG{p}{(}
      \PYG{n}{r}\PYG{o}{.}\PYG{n}{latest\PYGZus{}finish} \PYG{o}{\PYGZlt{}=} \PYG{n}{time} \PYG{o}{+} \PYG{l+m+mf}{1.5} \PYG{o}{*} \PYG{n}{ride\PYGZus{}length}\PYG{p}{)} \PYG{k}{else} \PYG{l+m+mi}{0}
  \PYG{k}{if} \PYG{n}{time} \PYG{o}{\PYGZlt{}=} \PYG{l+m+mi}{3600}\PYG{p}{:}
    \PYG{n}{bonus\PYGZus{}points} \PYG{o}{=} \PYG{n}{bonus\PYGZus{}points} \PYG{o}{+} \PYG{l+m+mi}{75} \PYG{o}{*} \PYG{n}{free\PYGZus{}time}
  \PYG{k}{else}\PYG{p}{:}
    \PYG{n}{bonus\PYGZus{}points} \PYG{o}{=} \PYG{n}{bonus\PYGZus{}points} \PYG{o}{\PYGZhy{}} \PYG{l+m+mf}{7.5} \PYG{o}{*} \PYG{n}{free\PYGZus{}time}

  \PYG{k}{if} \PYG{n}{free\PYGZus{}time} \PYG{o}{\PYGZlt{}=} \PYG{n}{r}\PYG{o}{.}\PYG{n}{latest\PYGZus{}finish} \PYG{o+ow}{and} \PYG{p}{(}
      \PYG{n}{r}\PYG{o}{.}\PYG{n}{earliest\PYGZus{}start} \PYG{o}{\PYGZlt{}=} \PYG{n}{pickup\PYGZus{}time}\PYG{p}{):}
    \PYG{n}{score} \PYG{o}{=} \PYG{n}{ride\PYGZus{}length} \PYG{o}{+} \PYG{n}{bonus\PYGZus{}points} \PYG{o}{\PYGZhy{}}\PYGZbs{}
        \PYG{l+m+mi}{15} \PYG{o}{*} \PYG{p}{(}\PYG{n+nb}{abs}\PYG{p}{(}
            \PYG{n}{r}\PYG{o}{.}\PYG{n}{start}\PYG{p}{[}\PYG{l+m+mi}{0}\PYG{p}{]} \PYG{o}{\PYGZhy{}} \PYG{n}{coords}\PYG{p}{[}\PYG{l+m+mi}{0}\PYG{p}{])} \PYG{o}{+} \PYG{n+nb}{abs}\PYG{p}{(}
                \PYG{n}{r}\PYG{o}{.}\PYG{n}{start}\PYG{p}{[}\PYG{l+m+mi}{1}\PYG{p}{]} \PYG{o}{\PYGZhy{}} \PYG{n}{coords}\PYG{p}{[}\PYG{l+m+mi}{1}\PYG{p}{]))} \PYG{o}{\PYGZhy{}}\PYGZbs{}
        \PYG{l+m+mi}{200} \PYG{o}{*} \PYG{n+nb}{max}\PYG{p}{([}\PYG{l+m+mi}{0}\PYG{p}{,} \PYG{n}{pickup\PYGZus{}time} \PYG{o}{\PYGZhy{}} \PYG{n}{time}\PYG{p}{])} \PYG{o}{\PYGZhy{}}\PYGZbs{}
        \PYG{l+m+mf}{1.} \PYG{o}{*} \PYG{n+nb}{sum}\PYG{p}{([}
            \PYG{n+nb}{abs}\PYG{p}{(}\PYG{n}{pickup\PYGZus{}time} \PYG{o}{\PYGZhy{}} \PYG{n}{r}\PYG{o}{.}\PYG{n}{earliest\PYGZus{}start}\PYG{p}{),}
            \PYG{n+nb}{abs}\PYG{p}{(}\PYG{n}{free\PYGZus{}time} \PYG{o}{\PYGZhy{}} \PYG{n}{r}\PYG{o}{.}\PYG{n}{latest\PYGZus{}finish}\PYG{p}{),}
            \PYG{n+nb}{sum}\PYG{p}{([}\PYG{n+nb}{abs}\PYG{p}{(}
                \PYG{n}{r}\PYG{o}{.}\PYG{n}{start}\PYG{p}{[}\PYG{n}{j}\PYG{p}{]} \PYG{o}{\PYGZhy{}} \PYG{n}{r}\PYG{o}{.}\PYG{n}{end}\PYG{p}{[}\PYG{n}{j}\PYG{p}{])} \PYG{k}{for} \PYG{n}{j} \PYG{o+ow}{in} \PYG{n+nb}{range}\PYG{p}{(}\PYG{l+m+mi}{2}\PYG{p}{)])}
        \PYG{p}{])}

    \PYG{c+c1}{\PYGZsh{} distance penalty from ending location}
    \PYG{n}{score} \PYG{o}{=} \PYG{n}{score} \PYG{o}{+} \PYG{l+m+mi}{15} \PYG{o}{*} \PYG{p}{(}
        \PYG{l+m+mi}{1100} \PYG{o}{\PYGZhy{}} \PYG{n}{r}\PYG{o}{.}\PYG{n}{distance\PYGZus{}to\PYGZus{}start}\PYG{p}{(}
            \PYG{n}{coords}\PYG{p}{))} \PYG{o}{+} \PYG{l+m+mi}{90} \PYG{o}{*} \PYG{n}{r}\PYG{o}{.}\PYG{n}{distance\PYGZus{}to\PYGZus{}start}\PYG{p}{(}\PYG{n}{coords}\PYG{p}{)} \PYG{o}{/} \PYG{l+m+mf}{1200.}
    \PYG{c+c1}{\PYGZsh{} penalty for driving far in the late night and early morning}
    \PYG{k}{if} \PYG{n}{rides\PYGZus{}by\PYGZus{}length}\PYG{p}{[}\PYG{l+m+mi}{0}\PYG{p}{][}\PYG{l+m+mi}{0}\PYG{p}{]} \PYG{o}{==} \PYG{n}{i}\PYG{p}{:}
      \PYG{n}{score} \PYG{o}{=} \PYG{n}{score} \PYG{o}{\PYGZhy{}} \PYG{l+m+mi}{25} \PYG{o}{*} \PYG{n}{free\PYGZus{}time}
    \PYG{k}{if} \PYG{n}{time} \PYG{o}{\PYGZgt{}=} \PYG{l+m+mi}{39480} \PYG{o+ow}{and} \PYG{n}{free\PYGZus{}time} \PYG{o}{\PYGZgt{}} \PYG{n}{time} \PYG{o}{//} \PYG{l+m+mf}{2.}\PYG{p}{:}
      \PYG{n}{score} \PYG{o}{=} \PYG{n}{score} \PYG{o}{+} \PYG{l+m+mi}{10} \PYG{o}{*} \PYG{p}{(}
          \PYG{n}{free\PYGZus{}time} \PYG{o}{\PYGZhy{}} \PYG{l+m+mi}{1100} \PYG{o}{+} \PYG{n}{r}\PYG{o}{.}\PYG{n}{distance\PYGZus{}to\PYGZus{}start}\PYG{p}{(}\PYG{n}{coords}\PYG{p}{))}
    \PYG{k}{if} \PYG{n}{r}\PYG{o}{.}\PYG{n}{latest\PYGZus{}finish} \PYG{o}{\PYGZlt{}=} \PYG{n}{time} \PYG{o}{+} \PYG{l+m+mi}{2} \PYG{o}{*} \PYG{n}{r}\PYG{o}{.}\PYG{n}{length}\PYG{p}{():}
      \PYG{n}{score} \PYG{o}{=} \PYG{n}{score} \PYG{o}{+} \PYG{l+m+mi}{8000}
    \PYG{k}{if} \PYG{n}{r}\PYG{o}{.}\PYG{n}{length}\PYG{p}{()} \PYG{o}{\PYGZlt{}} \PYG{l+m+mi}{1500}\PYG{p}{:}
      \PYG{n}{score} \PYG{o}{==} \PYG{n}{score} \PYG{o}{+} \PYG{l+m+mi}{2000}
    \PYG{k}{if} \PYG{n}{r}\PYG{o}{.}\PYG{n}{length}\PYG{p}{()} \PYG{o}{\PYGZgt{}} \PYG{l+m+mi}{6000}\PYG{p}{:}
      \PYG{n}{score} \PYG{o}{=} \PYG{n}{score} \PYG{o}{\PYGZhy{}} \PYG{l+m+mi}{3000}
    \PYG{k}{if} \PYG{n}{r}\PYG{o}{.}\PYG{n}{length}\PYG{p}{()} \PYG{o}{\PYGZgt{}} \PYG{l+m+mi}{7000}\PYG{p}{:}
      \PYG{n}{score} \PYG{o}{=} \PYG{n}{score} \PYG{o}{\PYGZhy{}} \PYG{l+m+mi}{3000}
    \PYG{k}{if} \PYG{n}{r}\PYG{o}{.}\PYG{n}{length}\PYG{p}{()} \PYG{o}{\PYGZgt{}} \PYG{l+m+mi}{5000}\PYG{p}{:}
      \PYG{n}{score} \PYG{o}{=} \PYG{n}{score} \PYG{o}{\PYGZhy{}} \PYG{l+m+mi}{5000}

    \PYG{k}{if} \PYG{n}{score} \PYG{o}{\PYGZgt{}} \PYG{n}{best\PYGZus{}score}\PYG{p}{:}
      \PYG{n}{best\PYGZus{}ride} \PYG{o}{=} \PYG{n}{i}
      \PYG{n}{best\PYGZus{}score} \PYG{o}{=} \PYG{n}{score}

\PYG{k}{return} \PYG{n}{best\PYGZus{}ride}
\end{Verbatim}
\end{tiny}
This solution achieves a score of $12,296,845$ points on the \texttt{metropolis} input, and is generally sufficient for outperforming the rank-1 team when aggregating across all inputs. 

While it clearly implements a more elaborate scoring function than its predecessors, it also makes some peculiar or redundant decisions. For example, while its comment indicates it will sort rides in descending order by distance, the effect of the sorting function is to sort in reverse using the first key -- the index. This only has the effect of processing the feasible rides in the reverse order from the one that they were given.

Then, the evolved function computes a bonus which can be allocated for rides without a lot of margin for error that are queried early or late. Besides this bonus, several other heuristics, including the distance of the car to the starting point, and whether the endpoint of one ride can be chained to the start point of another (\texttt{-200 * max([0, pickup\_time - time])}), all factor into the computed score. There are also several penalties applied to the score designed to encourage choosing shorter rides (e.g. \texttt{if r.length() > 6000: score = score - 3000}), with an interesting no-op command being accidentally executed in \texttt{score == score + 2000}.

\subsection{Qualification 2021: \emph{Traffic signaling}}

Finally, we study how the scoring function evolves on the 2021 Hash Code qualification round, where the objective is to find an optimal traffic light schedule at various city intersections.

Specifically, we are given a graph of one-way streets connecting intersections, as well as a collection of cars making their commute---for each car, we are given the list of streets they need to traverse. At each intersection there is a traffic light which can only be green for one of its incoming streets at a time, at which point it lets through one waiting car per timestep. The task is to decide on a traffic light schedule---in which order do the traffic lights turn green, and for how long is each intersection green---to optimise the number of cars completing their trips in time. Further, bonus points are allocated for each car completing their trip, depending on how quickly they finish.

For this round, we focus on the testcase \texttt{f\_forever\_jammed.in}, where the model made the most gradual improvement in score compared to the backbone---likely due to the higher incidence of heavily congested intersections. Just as in the previous examples, we provide essential parts of our backbone function in the Appendix (Figure \ref{fig:backbone_traf}).

For this task, the backbone greedy solution relies on two separate predictions for each street: which \emph{position} will it have in the traffic light schedule for its incoming intersection, and which \emph{duration} will it be green for. These are not computed at the same time for every street, but rather, a three-phase solution is executed:
\begin{itemize}
    \item First, a simulation of the system is performed with all the cars, under the assumption the green light durations will all be $1$. Whenever a car arrives at an intersection that hasn't previously been assigned a position, the position prediction is invoked on it. If the predicted position is already occupied, it is incremented until a free slot is found.
    \item Second, with the predicted positions assumed fixed, the duration is predicted for every incoming street of every intersection.
    \item Finally, another full simulation is performed under the computed parameters. Some cars may \emph{fail} to finish their trips under this simulation---if there are any streets for which only failed cars enter, we remove these streets from the schedule (i.e., assume a constant red light for them) to avoid these cars further congesting the traffic.
\end{itemize}
The base scoring function simply attempts to place every incoming street in position zero, and assigns to every street a green light time of one:

\begin{footnotesize}
\begin{Verbatim}[commandchars=\\\{\}]
\PYG{k}{if} \PYG{n}{give\PYGZus{}pos}\PYG{p}{:}
  \PYG{k}{return} \PYG{l+m+mi}{0}
\PYG{k}{else}\PYG{p}{:}
  \PYG{k}{return} \PYG{l+m+mi}{1}
\end{Verbatim}
\end{footnotesize}
Note the variable \texttt{give\_pos} which controls whether we are predicting position or green light duration. This simple scoring, coupled with the elaborate backbone, is sufficient to score an impressive $1,019,868$ points on this input. 

We can improve this score significantly after evolving for two hours---after a chain of over $20$ mutation calls to the LLM, the following scoring is obtained:

\begin{tiny}
\begin{Verbatim}[commandchars=\\\{\}]
\PYG{k}{if} \PYG{n}{give\PYGZus{}pos}\PYG{p}{:}
  \PYG{k}{return} \PYG{n+nb}{int}\PYG{p}{(}\PYG{n}{used\PYGZus{}streets}\PYG{p}{[}\PYG{n}{street}\PYG{o}{.}\PYG{n}{name}\PYG{p}{]} \PYG{o}{/} \PYG{l+m+mi}{1000} \PYG{o}{+} \PYG{l+m+mf}{0.5}\PYG{p}{)}
\PYG{k}{else}\PYG{p}{:}
  \PYG{k}{return} \PYG{n+nb}{int}\PYG{p}{(}
      \PYG{p}{(}\PYG{n}{used\PYGZus{}streets}\PYG{p}{[}\PYG{n}{street}\PYG{o}{.}\PYG{n}{name}\PYG{p}{]} \PYG{o}{*} \PYG{l+m+mf}{0.001} \PYG{o}{*} \PYG{n}{curr\PYGZus{}size} \PYG{o}{+} \PYG{l+m+mf}{0.1}\PYG{p}{)}
      \PYG{o}{*} \PYG{n}{math}\PYG{o}{.}\PYG{n}{log}\PYG{p}{((}\PYG{n}{used\PYGZus{}streets}\PYG{p}{[}\PYG{n}{street}\PYG{o}{.}\PYG{n}{name}\PYG{p}{]} \PYG{o}{+} \PYG{l+m+mi}{1}\PYG{p}{))} \PYG{o}{+} \PYG{l+m+mi}{1}\PYG{p}{)}
\end{Verbatim}
\end{tiny}
This solution subtly biases the semantics of the \texttt{give\_pos == True} case: it allocates the position to $\left \lfloor{N_\mathrm{cars}/1000 + 1/2}\right \rfloor $ where $N_\mathrm{cars}$ is the number of cars using this street in total. This has some level of bias towards later positions for more heavily used streets, but a load of at least $500$ cars is needed before any shift is triggered.

A significantly wider shift may be observed for predicting green light durations: $\left \lfloor{(N_\mathrm{cars}N_\mathrm{str}/1000 + 1/10)\ln (N_\mathrm{cars} + 1) + 1}\right \rfloor$ where $N_\mathrm{str}$ is the total number of streets entering this intersection. This formula amplifies the priority of the streets based on how congested they can get ($N_\mathrm{cars}$) as well as how much waiting there could be while the light cycles through all other streets ($N_\mathrm{str}$). Another key improvement is that this light duration should grow \emph{logarithmically} with the level of congestion expected.

With these two formulae in place, a score of $1,463,336$ is achieved---and it is one that takes a significant time to meaningfully outperform. Given sufficient time, our model has managed to discover functions that fine-tune the learnt formula somewhat to release additional traffic:

\begin{tiny}
\begin{Verbatim}[commandchars=\\\{\}]
\PYG{n}{l\PYGZus{}used} \PYG{o}{=} \PYG{n}{used\PYGZus{}streets}\PYG{o}{.}\PYG{n}{get}\PYG{p}{(}\PYG{n}{street}\PYG{o}{.}\PYG{n}{name}\PYG{p}{,} \PYG{l+m+mi}{0}\PYG{p}{)}
\PYG{k}{if} \PYG{n}{give\PYGZus{}pos}\PYG{p}{:}
  \PYG{k}{return} \PYG{n+nb}{max}\PYG{p}{(}\PYG{l+m+mi}{0}\PYG{p}{,} \PYG{n+nb}{int}\PYG{p}{(}\PYG{n+nb}{min}\PYG{p}{(}\PYG{n}{curr\PYGZus{}size} \PYG{o}{\PYGZhy{}} \PYG{l+m+mi}{1}\PYG{p}{,} \PYG{n}{l\PYGZus{}used} \PYG{o}{//} \PYG{l+m+mi}{200} \PYG{o}{*} \PYG{l+m+mi}{2}\PYG{p}{)))}
\PYG{k}{else}\PYG{p}{:}
  \PYG{k}{return} \PYG{n+nb}{max}\PYG{p}{(}\PYG{l+m+mi}{1}\PYG{p}{,} \PYG{n+nb}{int}\PYG{p}{(}\PYG{n+nb}{min}\PYG{p}{(}\PYG{l+m+mi}{1000}\PYG{p}{,} \PYG{p}{(}
      \PYG{n}{l\PYGZus{}used} \PYG{o}{*} \PYG{l+m+mf}{0.001} \PYG{o}{*} \PYG{n}{curr\PYGZus{}size} \PYG{o}{+} \PYG{l+m+mf}{0.1}\PYG{p}{)} \PYG{o}{*} \PYG{n}{math}\PYG{o}{.}\PYG{n}{log}\PYG{p}{(}
          \PYG{p}{(}\PYG{n}{l\PYGZus{}used} \PYG{o}{+} \PYG{l+m+mi}{1}\PYG{p}{))} \PYG{o}{+} \PYG{l+m+mi}{1}\PYG{p}{)))}
\end{Verbatim}
\end{tiny}
In this case, the prediction for green light duration (\texttt{give\_pos == False}) is largely unchanged: the main difference is capping the length at a maximum value of $1,000$. A more substantial change occurs in predicting the scheduling order of the green lights: the predicted position becomes $\max(0, \min(\widetilde{N}_\mathrm{str} - 1, \left\lfloor{N_\mathrm{cars} / 200}\right\rfloor * 2)$, where $\widetilde{N}_\mathrm{str}$ is the total number of streets that entered this intersection \emph{so far} (note the difference between what the variable \texttt{curr\_size} will have when assigning positions vs. assigning durations!).

This intricate evolved formula gives precedence to streets that are encountered earlier for a particular intersection (smaller $\widetilde{N}_\mathrm{str}$) as well as streets with a lower volume of cars using it (smaller $N_\mathrm{cars}$). One reasoning behind this could be to allow early, less-frequently traversed streets to be passed quickly, while delaying the more busy streets for later in the schedule so that the cars are more likely to queue up and make full advantage of the longer green cycle. 

These tuned constants additionally amplify the score on this input to $1,465,888$. Taken together with all evolved functions on individual inputs, our method achieves a comfortable rank-1 result. 

\section{Held-out combinatorial contest case study: \emph{AtCoder Heuristic Contest 039}}

As an additional piece of evidence that our approach provides meaningful improvements regardless of whether prior data relevant to the contest exists, we attempt to apply our FunSearch-backed method to a variant of the recently-held AtCoder Heuristic Contest (AHC 039). 

This AHC took place on 10 November 2024, which was \emph{after} the global release of Gemini 1.5 Flash 002. Hence, any information or strategies pertaining to the contest are not present in the data used to pre-train or fine-tune the model.

\subsection{AtCoder Heuristic Contests}

In addition to the held-out nature of the contest w.r.t. Gemini 1.5 Flash 002's training, it is also a substantially different setup to Hash Code:
\begin{itemize}
    \item The input test-cases are no longer given but are kept \emph{private} -- though the procedure used to generate them is known in advance.
    \item The contestants are required to submit their code to the testing system, and the code needs to be able to compile and execute in its entirety---it is not possible to offload any computation to an external tool.
\end{itemize}
This setting is expected to be more challenging for FunSearch, as its internal fitness function will not correspond to the actual evaluation that will be used to determine its solutions' rank.

\subsection{AHC 039: \emph{Purse Seine Fishing}}

In this task, we are given a collection of points on the 2D integer grid (i.e., all points have to be of the form $(a, b)$, for $a, b\in\mathbb{N}$). In addition, each point is one of two types (``mackerels'' and ``sardines'' in the task statement). The objective is to discover a polygon with only horizontal and vertical edges, which covers as many points of type 1 and as little points of type 2. 

There are also constraints in terms of the maximal number of vertices and maximal total edge length that such a polygon can have, encouraging polygons that are more strongly contiguous.

For the testcases used to evaluate solutions, all points are generated from a mixture of Gaussians, with randomly generated parameters (e.g. cluster centers and standard deviations, mixture weights\dots) for points of type 1 and type 2.

The strategy our backbone will pursue is to start from one grid cell, then maintaining a connected polygon by iteratively adding more points to it. Since the raw grid is too dense for building directly in it, we partition the grid into coarser square cells of size $N \times N$, and only consider adding entire such cells at once to the polygon.

\begin{figure}
    \centering
    \includegraphics[width=\linewidth]{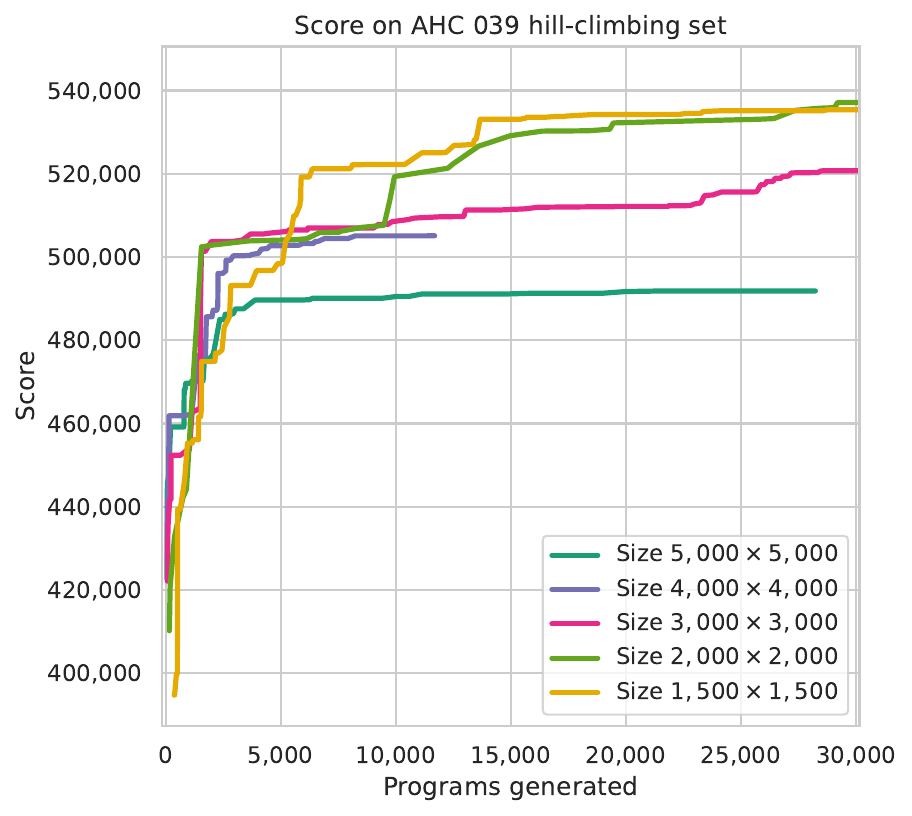}
    \caption{The improvement in score obtained by FunSearch on the hill-climbing dataset we used for AHC 039, over the first $30,000$ programs. Each line corresponds to a particular cell size.}
    \label{fig:ahc}
\end{figure}

We provide several aspects of our backbone solution for FunSearch in the Appendix (Figure \ref{fig:backbone_fish}). Note that this solution is \emph{not} the fastest possible implementation---in several locations, data points are overwritten in order to prevent FunSearch from deliberately modifying internal state. Our starting scoring function simply rates each cell of the coarse grid by their difference of points of type 1 and 2, without taking into account which cells are already in the polygon:

\begin{footnotesize}
\begin{Verbatim}[commandchars=\\\{\}]
\PYG{k}{def} \PYG{n+nf}{score\PYGZus{}greedy}\PYG{p}{(}
        \PYG{n}{grid}\PYG{p}{:} \PYG{n}{Grid}\PYG{p}{,}
        \PYG{n}{row}\PYG{p}{:} \PYG{n+nb}{int}\PYG{p}{,}
        \PYG{n}{col}\PYG{p}{:} \PYG{n+nb}{int}\PYG{p}{,}
        \PYG{n}{picked\PYGZus{}cells}\PYG{p}{:} \PYG{n+nb}{set}\PYG{p}{[}\PYG{n+nb}{tuple}\PYG{p}{[}\PYG{n+nb}{int}\PYG{p}{,} \PYG{n+nb}{int}\PYG{p}{]]):}
  \PYG{l+s+sd}{\PYGZdq{}\PYGZdq{}\PYGZdq{}Returns the score of picking a cell.\PYGZdq{}\PYGZdq{}\PYGZdq{}}
  \PYG{k}{return} \PYG{n}{grid}\PYG{o}{.}\PYG{n}{mackerels}\PYG{p}{[}\PYG{n}{row}\PYG{p}{][}
      \PYG{n}{col}\PYG{p}{]} \PYG{o}{\PYGZhy{}} \PYG{n}{grid}\PYG{o}{.}\PYG{n}{sardines}\PYG{p}{[}\PYG{n}{row}\PYG{p}{][}\PYG{n}{col}\PYG{p}{]}
\end{Verbatim}
\end{footnotesize}
We evolve this scoring function using FunSearch on a hill-climbing dataset of 150 examples which we pre-generated following the procedure in the task statement. As can be seen in Figure \ref{fig:ahc}, our models can make steady progress on improving this function, across many different choices of cell size $N \times N$.

The best-performing function we obtained (on cell size $2,000 \times 2,000$) was of the following form:

\begin{tiny}
\begin{Verbatim}[commandchars=\\\{\},codes={\catcode`\$=3\catcode`\^=7\catcode`\_=8\relax}]
\PYG{k}{if} \PYG{p}{(}\PYG{n}{row}\PYG{p}{,} \PYG{n}{col}\PYG{p}{)} \PYG{o+ow}{in} \PYG{n}{picked\PYGZus{}cells}\PYG{p}{:}
  \PYG{k}{return} \PYG{l+m+mi}{0}
\PYG{n}{m} \PYG{o}{=} \PYG{n}{grid}\PYG{o}{.}\PYG{n}{mackerels}\PYG{p}{[}\PYG{n}{row}\PYG{p}{]}\PYG{p}{[}\PYG{n}{col}\PYG{p}{]}
\PYG{n}{s} \PYG{o}{=} \PYG{n}{grid}\PYG{o}{.}\PYG{n}{sardines}\PYG{p}{[}\PYG{n}{row}\PYG{p}{]}\PYG{p}{[}\PYG{n}{col}\PYG{p}{]}
\PYG{n}{score} \PYG{o}{=} \PYG{n}{m} \PYG{o}{\PYGZhy{}} \PYG{l+m+mf}{1.3} \PYG{o}{*} \PYG{n}{s}
\PYG{n}{num\PYGZus{}picked} \PYG{o}{=} \PYG{n+nb}{len}\PYG{p}{(}\PYG{n}{picked\PYGZus{}cells}\PYG{p}{)}
\PYG{n}{score\PYGZus{}multiplier} \PYG{o}{=} \PYG{l+m+mf}{1.0} \PYG{o}{+} \PYG{l+m+mf}{0.01} \PYG{o}{*} \PYG{n+nb}{min}\PYG{p}{(}\PYG{n}{num\PYGZus{}picked}\PYG{p}{,} \PYG{l+m+mi}{100}\PYG{p}{)}
\PYG{k}{if} \PYG{n}{num\PYGZus{}picked} \PYG{o}{==} \PYG{l+m+mi}{0}\PYG{p}{:}
  \PYG{n}{score} \PYG{o}{*}\PYG{o}{=} \PYG{l+m+mf}{1.5}
\PYG{esc}{\xglobal\colorlet{FancyVerbHighlightColor}{dmgreen50}}\PYG{n}{dist\PYGZus{}center} \PYG{o}{=} \PYG{n+nb}{abs}\PYG{p}{(}\PYG{n}{grid}\PYG{o}{.}\PYG{n}{rows} \PYG{o}{/}\PYG{o}{/} \PYG{l+m+mi}{2} \PYG{o}{\PYGZhy{}} \PYG{n}{row}\PYG{p}{)} \PYG{o}{+} \PYG{n+nb}{abs}\PYG{p}{(}
    \PYG{n}{grid}\PYG{o}{.}\PYG{n}{cols} \PYG{o}{/}\PYG{o}{/} \PYG{l+m+mi}{2} \PYG{o}{\PYGZhy{}} \PYG{n}{col}\PYG{p}{)}
\PYG{n}{center\PYGZus{}bias} \PYG{o}{=} \PYG{n+nb}{max}\PYG{p}{(}\PYG{l+m+mi}{0}\PYG{p}{,} \PYG{l+m+mf}{1.0} \PYG{o}{/} \PYG{p}{(}\PYG{l+m+mf}{1.0} \PYG{o}{+} \PYG{n}{dist\PYGZus{}center}\PYG{o}{*}\PYG{o}{*}\PYG{l+m+mi}{2}\PYG{p}{)}\PYG{p}{)}
\PYG{n}{score} \PYG{o}{+}\PYG{o}{=} \PYG{n}{center\PYGZus{}bias} \PYG{o}{*} \PYG{n}{score} \PYG{o}{\PYGZhy{}} \PYG{n}{center\PYGZus{}bias} \PYG{o}{*} \PYG{l+m+mf}{0.4} \PYG{o}{*} \PYG{n}{s}
\PYG{n}{dist\PYGZus{}weight} \PYG{o}{=} \PYG{l+m+mf}{1.3}
\PYG{esc}{\xglobal\colorlet{FancyVerbHighlightColor}{dmcyan50}}\PYG{k}{for} \PYG{n}{r} \PYG{o+ow}{in} \PYG{n+nb}{range}\PYG{p}{(}\PYG{n+nb}{max}\PYG{p}{(}\PYG{l+m+mi}{0}\PYG{p}{,} \PYG{n}{row} \PYG{o}{\PYGZhy{}} \PYG{l+m+mi}{15}\PYG{p}{)}\PYG{p}{,} \PYG{n+nb}{min}\PYG{p}{(}\PYG{n}{grid}\PYG{o}{.}\PYG{n}{rows}\PYG{p}{,} \PYG{n}{row} \PYG{o}{+} \PYG{l+m+mi}{16}\PYG{p}{)}\PYG{p}{)}\PYG{p}{:}
  \PYG{k}{for} \PYG{n}{c} \PYG{o+ow}{in} \PYG{n+nb}{range}\PYG{p}{(}\PYG{n+nb}{max}\PYG{p}{(}\PYG{l+m+mi}{0}\PYG{p}{,} \PYG{n}{col} \PYG{o}{\PYGZhy{}} \PYG{l+m+mi}{15}\PYG{p}{)}\PYG{p}{,} \PYG{n+nb}{min}\PYG{p}{(}\PYG{n}{grid}\PYG{o}{.}\PYG{n}{cols}\PYG{p}{,} \PYG{n}{col} \PYG{o}{+} \PYG{l+m+mi}{16}\PYG{p}{)}\PYG{p}{)}\PYG{p}{:}
    \PYG{k}{if} \PYG{p}{(}\PYG{n}{r}\PYG{p}{,} \PYG{n}{c}\PYG{p}{)} \PYG{o+ow}{not} \PYG{o+ow}{in} \PYG{n}{picked\PYGZus{}cells} \PYG{o+ow}{and} \PYG{p}{(}\PYG{n}{r}\PYG{p}{,} \PYG{n}{c}\PYG{p}{)} \PYG{o}{!=} \PYG{p}{(}\PYG{n}{row}\PYG{p}{,} \PYG{n}{col}\PYG{p}{)}\PYG{p}{:}
      \PYG{n}{dist} \PYG{o}{=} \PYG{n+nb}{abs}\PYG{p}{(}\PYG{n}{row} \PYG{o}{\PYGZhy{}} \PYG{n}{r}\PYG{p}{)} \PYG{o}{+} \PYG{n+nb}{abs}\PYG{p}{(}\PYG{n}{col} \PYG{o}{\PYGZhy{}} \PYG{n}{c}\PYG{p}{)}
      \PYG{n}{adj\PYGZus{}macks} \PYG{o}{=} \PYG{l+m+mi}{0}
      \PYG{k}{for} \PYG{n}{dx}\PYG{p}{,} \PYG{n}{dy} \PYG{o+ow}{in} \PYG{p}{[}\PYG{p}{(}\PYG{o}{\PYGZhy{}}\PYG{l+m+mi}{1}\PYG{p}{,} \PYG{l+m+mi}{0}\PYG{p}{)}\PYG{p}{,} \PYG{p}{(}\PYG{l+m+mi}{0}\PYG{p}{,} \PYG{o}{\PYGZhy{}}\PYG{l+m+mi}{1}\PYG{p}{)}\PYG{p}{,} \PYG{p}{(}\PYG{o}{+}\PYG{l+m+mi}{1}\PYG{p}{,} \PYG{l+m+mi}{0}\PYG{p}{)}\PYG{p}{,} \PYG{p}{(}\PYG{l+m+mi}{0}\PYG{p}{,} \PYG{o}{+}\PYG{l+m+mi}{1}\PYG{p}{)}\PYG{p}{]}\PYG{p}{:}
        \PYG{n}{test\PYGZus{}r} \PYG{o}{=} \PYG{n}{r} \PYG{o}{+} \PYG{n}{dx}
        \PYG{n}{test\PYGZus{}c} \PYG{o}{=} \PYG{n}{c} \PYG{o}{+} \PYG{n}{dy}
        \PYG{k}{if} \PYG{p}{(}\PYG{l+m+mi}{0} \PYG{o}{\PYGZlt{}}\PYG{o}{=} \PYG{n}{test\PYGZus{}r} \PYG{o}{\PYGZlt{}} \PYG{n}{grid}\PYG{o}{.}\PYG{n}{rows}\PYG{p}{)} \PYG{o+ow}{and} \PYG{p}{(}
            \PYG{l+m+mi}{0} \PYG{o}{\PYGZlt{}}\PYG{o}{=} \PYG{n}{test\PYGZus{}c} \PYG{o}{\PYGZlt{}} \PYG{n}{grid}\PYG{o}{.}\PYG{n}{cols}\PYG{p}{)} \PYG{o+ow}{and} \PYG{p}{(}
                \PYG{n}{test\PYGZus{}r}\PYG{p}{,} \PYG{n}{test\PYGZus{}c}\PYG{p}{)} \PYG{o+ow}{in} \PYG{n}{picked\PYGZus{}cells}\PYG{p}{:}
          \PYG{n}{adj\PYGZus{}macks} \PYG{o}{+}\PYG{o}{=} \PYG{n}{grid}\PYG{o}{.}\PYG{n}{mackerels}\PYG{p}{[}\PYG{n}{test\PYGZus{}r}\PYG{p}{]}\PYG{p}{[}\PYG{n}{test\PYGZus{}c}\PYG{p}{]}
      \PYG{n}{weight} \PYG{o}{=} \PYG{n+nb}{max}\PYG{p}{(}\PYG{l+m+mf}{1e\PYGZhy{}6}\PYG{p}{,} \PYG{l+m+mi}{1} \PYG{o}{/} \PYG{p}{(}
          \PYG{l+m+mi}{1} \PYG{o}{+} \PYG{n}{dist}\PYG{o}{*}\PYG{o}{*}\PYG{l+m+mi}{2} \PYG{o}{+} \PYG{n}{m} \PYG{o}{+} \PYG{n}{s} \PYG{o}{+} \PYG{n}{adj\PYGZus{}macks} \PYG{o}{+} \PYG{l+m+mf}{0.001} \PYG{o}{*} \PYG{n}{num\PYGZus{}picked}\PYG{p}{)}\PYG{p}{)}
      \PYG{n}{score} \PYG{o}{+}\PYG{o}{=} \PYG{n+nb}{max}\PYG{p}{(}\PYG{l+m+mi}{0}\PYG{p}{,} \PYG{p}{(}
          \PYG{n}{grid}\PYG{o}{.}\PYG{n}{mackerels}\PYG{p}{[}\PYG{n}{r}\PYG{p}{]}\PYG{p}{[}\PYG{n}{c}\PYG{p}{]} \PYG{o}{\PYGZhy{}} \PYG{l+m+mf}{1.6} \PYG{o}{*} \PYG{n}{grid}\PYG{o}{.}\PYG{n}{sardines}\PYG{p}{[}
              \PYG{n}{r}\PYG{p}{]}\PYG{p}{[}\PYG{n}{c}\PYG{p}{]}\PYG{p}{)} \PYG{o}{*} \PYG{n}{dist\PYGZus{}weight} \PYG{o}{*} \PYG{n}{weight}\PYG{p}{)}
\PYG{esc}{\xglobal\colorlet{FancyVerbHighlightColor}{dmred50}}\PYG{n}{adjacent\PYGZus{}cells} \PYG{o}{=} \PYG{l+m+mi}{0}
\PYG{k}{for} \PYG{n}{dx}\PYG{p}{,} \PYG{n}{dy} \PYG{o+ow}{in} \PYG{p}{[}\PYG{p}{(}\PYG{o}{\PYGZhy{}}\PYG{l+m+mi}{1}\PYG{p}{,} \PYG{l+m+mi}{0}\PYG{p}{)}\PYG{p}{,} \PYG{p}{(}\PYG{l+m+mi}{0}\PYG{p}{,} \PYG{o}{\PYGZhy{}}\PYG{l+m+mi}{1}\PYG{p}{)}\PYG{p}{,} \PYG{p}{(}\PYG{o}{+}\PYG{l+m+mi}{1}\PYG{p}{,} \PYG{l+m+mi}{0}\PYG{p}{)}\PYG{p}{,} \PYG{p}{(}\PYG{l+m+mi}{0}\PYG{p}{,} \PYG{o}{+}\PYG{l+m+mi}{1}\PYG{p}{)}\PYG{p}{]}\PYG{p}{:}
  \PYG{k}{if} \PYG{p}{(}\PYG{l+m+mi}{0} \PYG{o}{\PYGZlt{}}\PYG{o}{=} \PYG{n}{row} \PYG{o}{+} \PYG{n}{dx} \PYG{o}{\PYGZlt{}} \PYG{n}{grid}\PYG{o}{.}\PYG{n}{rows}\PYG{p}{)} \PYG{o+ow}{and} \PYG{p}{(}
      \PYG{l+m+mi}{0} \PYG{o}{\PYGZlt{}}\PYG{o}{=} \PYG{n}{col} \PYG{o}{+} \PYG{n}{dy} \PYG{o}{\PYGZlt{}} \PYG{n}{grid}\PYG{o}{.}\PYG{n}{cols}\PYG{p}{)} \PYG{o+ow}{and} \PYG{p}{(}
          \PYG{n}{row} \PYG{o}{+} \PYG{n}{dx}\PYG{p}{,} \PYG{n}{col} \PYG{o}{+} \PYG{n}{dy}\PYG{p}{)} \PYG{o+ow}{in} \PYG{n}{picked\PYGZus{}cells}\PYG{p}{:}
    \PYG{n}{adjacent\PYGZus{}cells} \PYG{o}{+}\PYG{o}{=} \PYG{l+m+mi}{1}
\PYG{n}{score} \PYG{o}{+}\PYG{o}{=} \PYG{l+m+mi}{7} \PYG{o}{*} \PYG{n}{adjacent\PYGZus{}cells}
\PYG{k}{return} \PYG{n+nb}{max}\PYG{p}{(}\PYG{l+m+mi}{0}\PYG{p}{,} \PYG{n}{score} \PYG{o}{*} \PYG{n}{score\PYGZus{}multiplier}\PYG{p}{)}
\end{Verbatim}
\end{tiny}
Since in this case the hill-climbing set containts many individual testcases rather than just one, this evolved function features novel, interpretable mechanisms which were not in any way suggested at the starting point. In comparison, it exploits the specifics of testcase distributions less.

We describe three of these key mechanisms that can be noticed in the evolved code snippet:
\begin{itemize}
    \item Adjusting the score depending on how proximal the currnt cell is to the \emph{center} of the grid (\texttt{center\_bias}).
    \item Expressing the score using a \emph{weighted average} over a subgrid of $30\times 30$ cells around the current cell, encapsulated in the \texttt{for} loops:
    
    \begin{Verbatim}[commandchars=\\\{\},codes={\catcode`\$=3\catcode`\^=7\catcode`\_=8\relax}]
\PYG{k}{for} \PYG{n}{r} \PYG{o+ow}{in} \PYG{n+nb}{range}\PYG{p}{(}\dots\PYG{p}{)}\PYG{p}{:}
  \PYG{k}{for} \PYG{n}{c} \PYG{o+ow}{in} \PYG{n+nb}{range}\PYG{p}{(}\dots\PYG{p}{)}\PYG{p}{:}
    \end{Verbatim}
    This aligns well with the assumption that the data is generated from a mixture of Gaussians, and is more likely to identify cluster centers.
    \item Preferring cells that have a greater number of adjacent cells in the polygon (\texttt{adjacent\_cells}). This encourages polygons to be more strongly contiguous, minimising total edge length and number of vertices, and increasing likelihood that the polygon will be valid.
\end{itemize}
This function achieved a score of $541,173$ on our pre-generated hill-climbing set of data points. Even though the pre-generated dataset is diverse, it is still likely that the hill climber performed some spurious mutations, that optimise score on those specific instances without better fitting the distribution overall. 

Accordingly, to estimate what is the likely rank our evolved scoring function would have achieved in the actual contest, we generate a new dataset of $1,500$ data points ($10\times$ the size of what was used in AHC 039) and use it to bootstrap our rank estimates. Note that we also make changes to the backbone by removing any inefficiencies induced by protecting FunSearch from overwriting global state, and we also attempt to use our evolved scoring function with several grid size lengths ($\{1,500, 2,000, 3,000, 4,000\}$) and report the maximal-scoring polygon across them.

Across our larger test dataset, we found that our final solution achieves an average score of $3,521.9 \pm 424.4$ (where the error bars are standard deviation).

To estimate the $95\%$ confidence interval of the total score across a randomly chosen dataset of $150$ data points, we multiply the mean value by $150$ and the standard deviation by $2\sqrt{150}$. 

Performing this calculation, we conclude the $95\%$ confidence interval of the final score is $528,285.4 \pm 10,396.6$. This means that, with $95\%$ confidence, our solution would rank between the $9^\text{th}$ place and the $17^\text{th}$ place---a significantly better score than the backbone, and a very commendable performance.

\section{Conclusions}

In this technical report, we explored a collaborative approach between human competitors and evolutionary search in the program space, to significantly amplify the competitors' performance in combinatorial competitive programming. Our approach yields significant performance improvements on all previous Hash Code Online Qualification Rounds---even when we restrict the amount of time allocated to the execution of the AI method to under two hours. In many cases, it is sufficient for a rank-1 result. The method also proves potent on a recent held-out AtCoder Heuristic Contest, with a setting substantially different to the one of Hash Code.

\bibliographystyle{abbrvnat}
\nobibliography*
\bibliography{googledeepmind-test}

%

\section*{Acknowledgements}
The authors wish to thank Bernardino Romera-Paredes, Amin Barekatain and Alhussein Fawzi for their helpful discussions on evolutionary programming with LLMs, as well as Ng\^{a}n V\~{u}, George Holland, Simon Osindero, Shakir Mohamed and Pushmeet Kohli for their many helpful remarks and comments on the work. We highly appreciate the efforts of R\'{e}mi Leblond, Juanita Bawagan, Zoubin Ghahramani, Bakhodir Ashirmatov, Przemek Pietrzkiewicz and Petr Mitrichev for reviewing our paper prior to the release. Last but not least, we are indebted to the entire Hash Code team for making the data relevant to previous competitions broadly available.

\section*{Author Contributions}
P.V. concieved the research direction, with helpful steering from M.B. and A.N. in terms of framing it in a scope where the approach from \citet{funsearch} would be mostly applicable. The backbones were designed by P.V., A.V., L.M., B.I. and L.B., with M.B. and A.N. providing significant assistance in terms of executing them scalably. P.V. analysed the four case studies. All authors contributed to writing the paper.

\section*{Funding}
This research was funded by Google DeepMind.

\appendix

\section{Supplementary backbone examples}
\begin{itemize}
    \item Hash Code 2015 Qualification Round (\emph{Optimize a Data Center}): Figure \ref{fig:backbone_data}.
    \item Hash Code 2018 Qualification Round (\emph{Self-driving rides}): Figure \ref{fig:backbone_self}.
    \item Hash Code 2021 Qualification Round (\emph{Traffic scheduling}): Figure \ref{fig:backbone_traf}.
    \item AtCoder Heuristic Contest 039 (\emph{Purse Seine Fishing}): Figure \ref{fig:backbone_fish}.
\end{itemize}

\begin{figure*}
\centering
\begin{minipage}[c]{0.48\textwidth}
\begin{tiny}
\begin{Verbatim}[commandchars=\\\{\}]
\PYG{n+nd}{@dataclasses}\PYG{o}{.}\PYG{n}{dataclass}\PYG{p}{(}\PYG{n}{frozen}\PYG{o}{=}\PYG{k+kc}{True}\PYG{p}{)}
\PYG{k}{class} \PYG{n+nc}{Server}\PYG{p}{:}
  \PYG{n}{index}\PYG{p}{:} \PYG{n+nb}{int}
  \PYG{n}{size}\PYG{p}{:} \PYG{n+nb}{int}
  \PYG{n}{capacity}\PYG{p}{:} \PYG{n+nb}{int}


\PYG{k}{def} \PYG{n+nf}{parse\PYGZus{}data}\PYG{p}{(}\PYG{n}{file\PYGZus{}path}\PYG{p}{:} \PYG{n+nb}{str}\PYG{p}{):}
  \PYG{l+s+sd}{\PYGZdq{}\PYGZdq{}\PYGZdq{}Parse data file.}

\PYG{l+s+sd}{  Args:}
\PYG{l+s+sd}{    file\PYGZus{}path: path to the datafile.}

\PYG{l+s+sd}{  Returns:}
\PYG{l+s+sd}{    row\PYGZus{}blocks}
\PYG{l+s+sd}{    servers}
\PYG{l+s+sd}{    n\PYGZus{}pools}
\PYG{l+s+sd}{  \PYGZdq{}\PYGZdq{}\PYGZdq{}}
  \PYG{k}{with} \PYG{n+nb}{open}\PYG{p}{(}\PYG{n}{file\PYGZus{}path}\PYG{p}{,} \PYG{l+s+s1}{\PYGZsq{}rt\PYGZsq{}}\PYG{p}{)} \PYG{k}{as} \PYG{n}{f}\PYG{p}{:}
    \PYG{n}{main\PYGZus{}line} \PYG{o}{=} \PYG{n}{f}\PYG{o}{.}\PYG{n}{readline}\PYG{p}{()}
    \PYG{n}{n\PYGZus{}rows}\PYG{p}{,} \PYG{n}{n\PYGZus{}slots}\PYG{p}{,} \PYG{n}{n\PYGZus{}unavailable}\PYG{p}{,} \PYG{n}{n\PYGZus{}pools}\PYG{p}{,} \PYG{n}{n\PYGZus{}servers} \PYG{o}{=} \PYG{n+nb}{map}\PYG{p}{(}
        \PYG{n+nb}{int}\PYG{p}{,} \PYG{n}{main\PYGZus{}line}\PYG{o}{.}\PYG{n}{split}\PYG{p}{())}
    \PYG{n}{row\PYGZus{}blocks} \PYG{o}{=} \PYG{p}{[]}
    \PYG{n}{servers} \PYG{o}{=} \PYG{p}{[]}
    \PYG{k}{for} \PYG{n}{\PYGZus{}} \PYG{o+ow}{in} \PYG{n+nb}{range}\PYG{p}{(}\PYG{n}{n\PYGZus{}rows}\PYG{p}{):}
      \PYG{n}{row\PYGZus{}blocks}\PYG{o}{.}\PYG{n}{append}\PYG{p}{([])}
    \PYG{k}{for} \PYG{n}{\PYGZus{}} \PYG{o+ow}{in} \PYG{n+nb}{range}\PYG{p}{(}\PYG{n}{n\PYGZus{}unavailable}\PYG{p}{):}
      \PYG{n}{u\PYGZus{}x}\PYG{p}{,} \PYG{n}{u\PYGZus{}y} \PYG{o}{=} \PYG{n+nb}{map}\PYG{p}{(}\PYG{n+nb}{int}\PYG{p}{,} \PYG{n}{f}\PYG{o}{.}\PYG{n}{readline}\PYG{p}{()}\PYG{o}{.}\PYG{n}{split}\PYG{p}{())}
      \PYG{n}{row\PYGZus{}blocks}\PYG{p}{[}\PYG{n}{u\PYGZus{}x}\PYG{p}{]}\PYG{o}{.}\PYG{n}{append}\PYG{p}{(}\PYG{n}{u\PYGZus{}y}\PYG{p}{)}
    \PYG{k}{for} \PYG{n}{i\PYGZus{}r} \PYG{o+ow}{in} \PYG{n+nb}{range}\PYG{p}{(}\PYG{n}{n\PYGZus{}rows}\PYG{p}{):}
      \PYG{n}{row\PYGZus{}blocks}\PYG{p}{[}\PYG{n}{i\PYGZus{}r}\PYG{p}{]}\PYG{o}{.}\PYG{n}{append}\PYG{p}{(}\PYG{n}{n\PYGZus{}slots}\PYG{p}{)}
      \PYG{n}{row\PYGZus{}blocks}\PYG{p}{[}\PYG{n}{i\PYGZus{}r}\PYG{p}{]} \PYG{o}{=} \PYG{n+nb}{sorted}\PYG{p}{(}\PYG{n}{row\PYGZus{}blocks}\PYG{p}{[}\PYG{n}{i\PYGZus{}r}\PYG{p}{])}
    \PYG{k}{for} \PYG{n}{i\PYGZus{}s} \PYG{o+ow}{in} \PYG{n+nb}{range}\PYG{p}{(}\PYG{n}{n\PYGZus{}servers}\PYG{p}{):}
      \PYG{n}{s\PYGZus{}size}\PYG{p}{,} \PYG{n}{s\PYGZus{}capacity} \PYG{o}{=} \PYG{n+nb}{map}\PYG{p}{(}\PYG{n+nb}{int}\PYG{p}{,} \PYG{n}{f}\PYG{o}{.}\PYG{n}{readline}\PYG{p}{()}\PYG{o}{.}\PYG{n}{split}\PYG{p}{())}
      \PYG{n}{servers}\PYG{o}{.}\PYG{n}{append}\PYG{p}{(}\PYG{n}{Server}\PYG{p}{(}\PYG{n}{i\PYGZus{}s}\PYG{p}{,} \PYG{n}{s\PYGZus{}size}\PYG{p}{,} \PYG{n}{s\PYGZus{}capacity}\PYG{p}{))}
  \PYG{k}{return} \PYG{n}{row\PYGZus{}blocks}\PYG{p}{,} \PYG{n}{servers}\PYG{p}{,} \PYG{n}{n\PYGZus{}slots}\PYG{p}{,} \PYG{n}{n\PYGZus{}pools}
\end{Verbatim}
\end{tiny}
\end{minipage}
\begin{minipage}[c]{0.48\textwidth}
\begin{tiny}
\begin{Verbatim}[commandchars=\\\{\}]
\PYG{k}{def} \PYG{n+nf}{evaluate}\PYG{p}{(}\PYG{n}{file\PYGZus{}path}\PYG{p}{:} \PYG{n+nb}{str}\PYG{p}{)} \PYG{o}{\PYGZhy{}\PYGZgt{}} \PYG{n+nb}{int}\PYG{p}{:}
  \PYG{l+s+sd}{\PYGZdq{}\PYGZdq{}\PYGZdq{}Returns data center value when using the evolved optimisation heuristic.\PYGZdq{}\PYGZdq{}\PYGZdq{}}
  \PYG{n}{row\PYGZus{}blocks}\PYG{p}{,} \PYG{n}{servers}\PYG{p}{,} \PYG{n}{n\PYGZus{}slots}\PYG{p}{,} \PYG{n}{n\PYGZus{}pools} \PYG{o}{=} \PYG{n}{parse\PYGZus{}data}\PYG{p}{(}\PYG{n}{file\PYGZus{}path}\PYG{p}{)}
  \PYG{n}{curr\PYGZus{}ind} \PYG{o}{=} \PYG{p}{[}\PYG{l+m+mi}{0} \PYG{k}{for} \PYG{n}{\PYGZus{}} \PYG{o+ow}{in} \PYG{n}{row\PYGZus{}blocks}\PYG{p}{]}
  \PYG{n}{curr\PYGZus{}block} \PYG{o}{=} \PYG{p}{[}\PYG{l+m+mi}{0} \PYG{k}{for} \PYG{n}{\PYGZus{}} \PYG{o+ow}{in} \PYG{n}{row\PYGZus{}blocks}\PYG{p}{]}
  \PYG{n}{open\PYGZus{}rows} \PYG{o}{=} \PYG{n+nb}{set}\PYG{p}{(}\PYG{n+nb}{range}\PYG{p}{(}\PYG{n+nb}{len}\PYG{p}{(}\PYG{n}{row\PYGZus{}blocks}\PYG{p}{)))}
  \PYG{n}{open\PYGZus{}servers} \PYG{o}{=} \PYG{n+nb}{set}\PYG{p}{(}\PYG{n+nb}{range}\PYG{p}{(}\PYG{n+nb}{len}\PYG{p}{(}\PYG{n}{servers}\PYG{p}{)))}
  \PYG{n}{placement} \PYG{o}{=} \PYG{p}{\PYGZob{}\PYGZcb{}}
  \PYG{k}{while} \PYG{n}{open\PYGZus{}rows} \PYG{o+ow}{and} \PYG{n}{open\PYGZus{}servers}\PYG{p}{:}
    \PYG{k}{for} \PYG{n}{row} \PYG{o+ow}{in} \PYG{n+nb}{set}\PYG{p}{(}\PYG{n}{open\PYGZus{}rows}\PYG{p}{):}
      \PYG{k}{if} \PYG{o+ow}{not} \PYG{n}{open\PYGZus{}servers}\PYG{p}{:}
        \PYG{k}{break}
      \PYG{k}{while} \PYG{n}{curr\PYGZus{}ind}\PYG{p}{[}\PYG{n}{row}\PYG{p}{]} \PYG{o}{==} \PYG{n}{row\PYGZus{}blocks}\PYG{p}{[}\PYG{n}{row}\PYG{p}{][}\PYG{n}{curr\PYGZus{}block}\PYG{p}{[}\PYG{n}{row}\PYG{p}{]]:}
        \PYG{n}{curr\PYGZus{}ind}\PYG{p}{[}\PYG{n}{row}\PYG{p}{]} \PYG{o}{+=} \PYG{l+m+mi}{1}
        \PYG{n}{curr\PYGZus{}block}\PYG{p}{[}\PYG{n}{row}\PYG{p}{]} \PYG{o}{+=} \PYG{l+m+mi}{1}
        \PYG{k}{if} \PYG{n}{curr\PYGZus{}ind}\PYG{p}{[}\PYG{n}{row}\PYG{p}{]} \PYG{o}{\PYGZgt{}=} \PYG{n}{n\PYGZus{}slots}\PYG{p}{:}
          \PYG{n}{open\PYGZus{}rows}\PYG{o}{.}\PYG{n}{remove}\PYG{p}{(}\PYG{n}{row}\PYG{p}{)}
          \PYG{k}{break}
      \PYG{k}{if} \PYG{n}{row} \PYG{o+ow}{not} \PYG{o+ow}{in} \PYG{n}{open\PYGZus{}rows}\PYG{p}{:}
        \PYG{k}{continue}
      \PYG{n}{next\PYGZus{}pos} \PYG{o}{=} \PYG{n}{row\PYGZus{}blocks}\PYG{p}{[}\PYG{n}{row}\PYG{p}{][}\PYG{n}{curr\PYGZus{}block}\PYG{p}{[}\PYG{n}{row}\PYG{p}{]]}
      \PYG{n}{best\PYGZus{}server} \PYG{o}{=} \PYG{k+kc}{None}
      \PYG{n}{best\PYGZus{}score} \PYG{o}{=} \PYG{o}{\PYGZhy{}}\PYG{l+m+mi}{1}
      \PYG{k}{for} \PYG{n}{s\PYGZus{}id} \PYG{o+ow}{in} \PYG{n}{open\PYGZus{}servers}\PYG{p}{:}
        \PYG{k}{if} \PYG{n}{curr\PYGZus{}ind}\PYG{p}{[}\PYG{n}{row}\PYG{p}{]} \PYG{o}{+} \PYG{n}{servers}\PYG{p}{[}\PYG{n}{s\PYGZus{}id}\PYG{p}{]}\PYG{o}{.}\PYG{n}{size} \PYG{o}{\PYGZlt{}=} \PYG{n}{next\PYGZus{}pos}\PYG{p}{:}
          \PYG{n}{curr\PYGZus{}score} \PYG{o}{=} \PYG{n}{score\PYGZus{}greedy}\PYG{p}{(}\PYG{n}{servers}\PYG{p}{[}\PYG{n}{s\PYGZus{}id}\PYG{p}{],} \PYG{n}{row}\PYG{p}{,} \PYG{k+kc}{None}\PYG{p}{,} \PYG{k+kc}{None}\PYG{p}{,} \PYG{k+kc}{True}\PYG{p}{)}
          \PYG{k}{if} \PYG{n}{best\PYGZus{}server} \PYG{o+ow}{is} \PYG{k+kc}{None} \PYG{o+ow}{or} \PYG{n}{curr\PYGZus{}score} \PYG{o}{\PYGZgt{}} \PYG{n}{best\PYGZus{}score}\PYG{p}{:}
            \PYG{n}{best\PYGZus{}score} \PYG{o}{=} \PYG{n}{curr\PYGZus{}score}
            \PYG{n}{best\PYGZus{}server} \PYG{o}{=} \PYG{n}{s\PYGZus{}id}
      \PYG{k}{if} \PYG{n}{best\PYGZus{}server} \PYG{o+ow}{is} \PYG{k+kc}{None}\PYG{p}{:}
        \PYG{n}{curr\PYGZus{}ind}\PYG{p}{[}\PYG{n}{row}\PYG{p}{]} \PYG{o}{=} \PYG{n}{next\PYGZus{}pos}
        \PYG{k}{if} \PYG{n}{curr\PYGZus{}ind}\PYG{p}{[}\PYG{n}{row}\PYG{p}{]} \PYG{o}{\PYGZgt{}=} \PYG{n}{n\PYGZus{}slots}\PYG{p}{:}
          \PYG{n}{open\PYGZus{}rows}\PYG{o}{.}\PYG{n}{remove}\PYG{p}{(}\PYG{n}{row}\PYG{p}{)}
      \PYG{k}{else}\PYG{p}{:}
        \PYG{n}{placement}\PYG{p}{[}\PYG{n}{best\PYGZus{}server}\PYG{p}{]} \PYG{o}{=} \PYG{p}{(}\PYG{n}{row}\PYG{p}{,} \PYG{n}{curr\PYGZus{}ind}\PYG{p}{[}\PYG{n}{row}\PYG{p}{],} \PYG{o}{\PYGZhy{}}\PYG{l+m+mi}{1}\PYG{p}{)}
        \PYG{n}{curr\PYGZus{}ind}\PYG{p}{[}\PYG{n}{row}\PYG{p}{]} \PYG{o}{=} \PYG{n}{curr\PYGZus{}ind}\PYG{p}{[}\PYG{n}{row}\PYG{p}{]} \PYG{o}{+} \PYG{n}{servers}\PYG{p}{[}\PYG{n}{best\PYGZus{}server}\PYG{p}{]}\PYG{o}{.}\PYG{n}{size}
        \PYG{n}{open\PYGZus{}servers}\PYG{o}{.}\PYG{n}{remove}\PYG{p}{(}\PYG{n}{best\PYGZus{}server}\PYG{p}{)}
  \PYG{n}{pools\PYGZus{}per\PYGZus{}row} \PYG{o}{=} \PYG{p}{\PYGZob{}\PYGZcb{}}
  \PYG{k}{for} \PYG{n}{row} \PYG{o+ow}{in} \PYG{n+nb}{range}\PYG{p}{(}\PYG{n+nb}{len}\PYG{p}{(}\PYG{n}{row\PYGZus{}blocks}\PYG{p}{)):}
    \PYG{n}{pools\PYGZus{}per\PYGZus{}row}\PYG{p}{[}\PYG{n}{row}\PYG{p}{]} \PYG{o}{=} \PYG{p}{\PYGZob{}\PYGZcb{}}
    \PYG{k}{for} \PYG{n}{p} \PYG{o+ow}{in} \PYG{n+nb}{range}\PYG{p}{(}\PYG{n}{n\PYGZus{}pools}\PYG{p}{):}
      \PYG{n}{pools\PYGZus{}per\PYGZus{}row}\PYG{p}{[}\PYG{n}{row}\PYG{p}{][}\PYG{n}{p}\PYG{p}{]} \PYG{o}{=} \PYG{l+m+mi}{0}
  \PYG{k}{for} \PYG{n}{s\PYGZus{}id} \PYG{o+ow}{in} \PYG{n+nb}{dict}\PYG{p}{(}\PYG{n}{placement}\PYG{p}{):}
    \PYG{n}{row}\PYG{p}{,} \PYG{n}{col}\PYG{p}{,} \PYG{n}{\PYGZus{}} \PYG{o}{=} \PYG{n}{placement}\PYG{p}{[}\PYG{n}{s\PYGZus{}id}\PYG{p}{]}
    \PYG{n}{best\PYGZus{}pool} \PYG{o}{=} \PYG{k+kc}{None}
    \PYG{n}{best\PYGZus{}score} \PYG{o}{=} \PYG{o}{\PYGZhy{}}\PYG{l+m+mi}{1}
    \PYG{k}{for} \PYG{n}{p} \PYG{o+ow}{in} \PYG{n+nb}{range}\PYG{p}{(}\PYG{n}{n\PYGZus{}pools}\PYG{p}{):}
      \PYG{n}{curr\PYGZus{}score} \PYG{o}{=} \PYG{n}{score\PYGZus{}greedy}\PYG{p}{(}\PYG{n}{servers}\PYG{p}{[}\PYG{n}{s\PYGZus{}id}\PYG{p}{],} \PYG{n}{row}\PYG{p}{,} \PYG{n}{p}\PYG{p}{,} \PYG{n}{pools\PYGZus{}per\PYGZus{}row}\PYG{p}{,} \PYG{k+kc}{False}\PYG{p}{)}
      \PYG{k}{if} \PYG{n}{best\PYGZus{}pool} \PYG{o+ow}{is} \PYG{k+kc}{None} \PYG{o+ow}{or} \PYG{n}{curr\PYGZus{}score} \PYG{o}{\PYGZgt{}} \PYG{n}{best\PYGZus{}score}\PYG{p}{:}
        \PYG{n}{best\PYGZus{}score} \PYG{o}{=} \PYG{n}{curr\PYGZus{}score}
        \PYG{n}{best\PYGZus{}pool} \PYG{o}{=} \PYG{n}{p}
    \PYG{k}{assert} \PYG{n}{best\PYGZus{}pool} \PYG{o+ow}{is} \PYG{o+ow}{not} \PYG{k+kc}{None}
    \PYG{n}{placement}\PYG{p}{[}\PYG{n}{s\PYGZus{}id}\PYG{p}{]} \PYG{o}{=} \PYG{p}{(}\PYG{n}{row}\PYG{p}{,} \PYG{n}{col}\PYG{p}{,} \PYG{n}{best\PYGZus{}pool}\PYG{p}{)}
    \PYG{n}{pools\PYGZus{}per\PYGZus{}row}\PYG{p}{[}\PYG{n}{row}\PYG{p}{][}\PYG{n}{best\PYGZus{}pool}\PYG{p}{]} \PYG{o}{+=} \PYG{n}{servers}\PYG{p}{[}\PYG{n}{s\PYGZus{}id}\PYG{p}{]}\PYG{o}{.}\PYG{n}{capacity}
  \PYG{k}{return} \PYG{n}{get\PYGZus{}guaranteed\PYGZus{}capacity}\PYG{p}{(}\PYG{n}{placement}\PYG{p}{,} \PYG{n}{file\PYGZus{}path}\PYG{p}{)}
\end{Verbatim}
\end{tiny}
\end{minipage}
\caption{The input parsing function and the greedy algorithm backbone for the 2015 Hash Code online qualification (Optimizing a Data Center). Note that the backbone is calling \texttt{score\_greedy}---the scoring function to optimise---and \texttt{get\_guaranteed\_capacity}---the evaluation function.}
\label{fig:backbone_data}
\end{figure*}

\begin{figure*}
\centering
\begin{minipage}[c]{0.48\textwidth}
\begin{tiny}
\begin{Verbatim}[commandchars=\\\{\}]
\PYG{n+nd}{@dataclasses}\PYG{o}{.}\PYG{n}{dataclass}\PYG{p}{(}\PYG{n}{frozen}\PYG{o}{=}\PYG{k+kc}{True}\PYG{p}{)}
\PYG{k}{class} \PYG{n+nc}{Ride}\PYG{p}{:}
  \PYG{n}{start}\PYG{p}{:} \PYG{n+nb}{tuple}\PYG{p}{[}\PYG{n+nb}{int}\PYG{p}{,} \PYG{n+nb}{int}\PYG{p}{]}
  \PYG{n}{end}\PYG{p}{:} \PYG{n+nb}{tuple}\PYG{p}{[}\PYG{n+nb}{int}\PYG{p}{,} \PYG{n+nb}{int}\PYG{p}{]}
  \PYG{n}{earliest\PYGZus{}start}\PYG{p}{:} \PYG{n+nb}{int}
  \PYG{n}{latest\PYGZus{}finish}\PYG{p}{:} \PYG{n+nb}{int}

  \PYG{k}{def} \PYG{n+nf}{length}\PYG{p}{(}\PYG{n+nb+bp}{self}\PYG{p}{):}
    \PYG{k}{return} \PYG{n}{distance}\PYG{p}{(}\PYG{n+nb+bp}{self}\PYG{o}{.}\PYG{n}{start}\PYG{p}{,} \PYG{n+nb+bp}{self}\PYG{o}{.}\PYG{n}{end}\PYG{p}{)}

  \PYG{k}{def} \PYG{n+nf}{distance\PYGZus{}to\PYGZus{}start}\PYG{p}{(}\PYG{n+nb+bp}{self}\PYG{p}{,} \PYG{n}{coords}\PYG{p}{:} \PYG{n+nb}{tuple}\PYG{p}{[}\PYG{n+nb}{int}\PYG{p}{,} \PYG{n+nb}{int}\PYG{p}{]):}
    \PYG{k}{return} \PYG{n}{distance}\PYG{p}{(}\PYG{n+nb+bp}{self}\PYG{o}{.}\PYG{n}{start}\PYG{p}{,} \PYG{n}{coords}\PYG{p}{)}


\PYG{k}{def} \PYG{n+nf}{distance}\PYG{p}{(}\PYG{n}{start}\PYG{p}{:} \PYG{n+nb}{tuple}\PYG{p}{[}\PYG{n+nb}{int}\PYG{p}{,} \PYG{n+nb}{int}\PYG{p}{],} \PYG{n}{end}\PYG{p}{:} \PYG{n+nb}{tuple}\PYG{p}{[}\PYG{n+nb}{int}\PYG{p}{,} \PYG{n+nb}{int}\PYG{p}{])} \PYG{o}{\PYGZhy{}\PYGZgt{}} \PYG{n+nb}{int}\PYG{p}{:}
  \PYG{k}{return} \PYG{n+nb}{abs}\PYG{p}{(}\PYG{n}{end}\PYG{p}{[}\PYG{l+m+mi}{0}\PYG{p}{]} \PYG{o}{\PYGZhy{}} \PYG{n}{start}\PYG{p}{[}\PYG{l+m+mi}{0}\PYG{p}{])} \PYG{o}{+} \PYG{n+nb}{abs}\PYG{p}{(}\PYG{n}{end}\PYG{p}{[}\PYG{l+m+mi}{1}\PYG{p}{]} \PYG{o}{\PYGZhy{}} \PYG{n}{start}\PYG{p}{[}\PYG{l+m+mi}{1}\PYG{p}{])}


\PYG{k}{def} \PYG{n+nf}{parse\PYGZus{}file}\PYG{p}{(}\PYG{n}{path}\PYG{p}{:} \PYG{n+nb}{str}\PYG{p}{):}
  \PYG{l+s+sd}{\PYGZdq{}\PYGZdq{}\PYGZdq{}Parses file.}

\PYG{l+s+sd}{  Args:}
\PYG{l+s+sd}{    path: Path to file.}

\PYG{l+s+sd}{  Returns:}
\PYG{l+s+sd}{    r, c, f, b, t, rides.}
\PYG{l+s+sd}{  \PYGZdq{}\PYGZdq{}\PYGZdq{}}
  \PYG{k}{with} \PYG{n+nb}{open}\PYG{p}{(}\PYG{n}{path}\PYG{p}{,} \PYG{l+s+s1}{\PYGZsq{}rt\PYGZsq{}}\PYG{p}{)} \PYG{k}{as} \PYG{n}{fp}\PYG{p}{:}
    \PYG{n}{r}\PYG{p}{,} \PYG{n}{c}\PYG{p}{,} \PYG{n}{f}\PYG{p}{,} \PYG{n}{n}\PYG{p}{,} \PYG{n}{b}\PYG{p}{,} \PYG{n}{t} \PYG{o}{=} \PYG{p}{[}\PYG{n+nb}{int}\PYG{p}{(}\PYG{n}{x}\PYG{p}{)} \PYG{k}{for} \PYG{n}{x} \PYG{o+ow}{in} \PYG{n}{fp}\PYG{o}{.}\PYG{n}{readline}\PYG{p}{()}\PYG{o}{.}\PYG{n}{split}\PYG{p}{()]}
    \PYG{n}{rides} \PYG{o}{=} \PYG{p}{[]}
    \PYG{k}{for} \PYG{n}{\PYGZus{}} \PYG{o+ow}{in} \PYG{n+nb}{range}\PYG{p}{(}\PYG{n}{n}\PYG{p}{):}
      \PYG{n}{rides}\PYG{o}{.}\PYG{n}{append}\PYG{p}{(}\PYG{n+nb}{tuple}\PYG{p}{(}\PYG{n+nb}{int}\PYG{p}{(}\PYG{n}{x}\PYG{p}{)} \PYG{k}{for} \PYG{n}{x} \PYG{o+ow}{in} \PYG{n}{fp}\PYG{o}{.}\PYG{n}{readline}\PYG{p}{()}\PYG{o}{.}\PYG{n}{split}\PYG{p}{()))}

  \PYG{k}{return} \PYG{n}{r}\PYG{p}{,} \PYG{n}{c}\PYG{p}{,} \PYG{n}{f}\PYG{p}{,} \PYG{n}{b}\PYG{p}{,} \PYG{n}{t}\PYG{p}{,} \PYG{n}{rides}
\end{Verbatim}
\end{tiny}
\end{minipage}
\begin{minipage}[c]{0.48\textwidth}
\begin{tiny}
\begin{Verbatim}[commandchars=\\\{\}]
\PYG{k}{def} \PYG{n+nf}{evaluate}\PYG{p}{(}\PYG{n}{path}\PYG{p}{:} \PYG{n+nb}{str}\PYG{p}{)} \PYG{o}{\PYGZhy{}\PYGZgt{}} \PYG{n+nb}{float}\PYG{p}{:}
  \PYG{l+s+sd}{\PYGZdq{}\PYGZdq{}\PYGZdq{}Assign cars to rides.\PYGZdq{}\PYGZdq{}\PYGZdq{}}
  \PYG{n}{unused\PYGZus{}r}\PYG{p}{,} \PYG{n}{unused\PYGZus{}c}\PYG{p}{,} \PYG{n}{f}\PYG{p}{,} \PYG{n}{b}\PYG{p}{,} \PYG{n}{tot\PYGZus{}time}\PYG{p}{,} \PYG{n}{rides} \PYG{o}{=} \PYG{n}{parse\PYGZus{}file}\PYG{p}{(}\PYG{n}{path}\PYG{p}{)}
  \PYG{n}{rides} \PYG{o}{=} \PYG{p}{[}
      \PYG{n}{Ride}\PYG{p}{(}\PYG{n}{start}\PYG{o}{=}\PYG{n}{x}\PYG{p}{[:}\PYG{l+m+mi}{2}\PYG{p}{],} \PYG{n}{end}\PYG{o}{=}\PYG{n}{x}\PYG{p}{[}\PYG{l+m+mi}{2}\PYG{p}{:}\PYG{l+m+mi}{4}\PYG{p}{],} \PYG{n}{earliest\PYGZus{}start}\PYG{o}{=}\PYG{n}{x}\PYG{p}{[}\PYG{l+m+mi}{4}\PYG{p}{],} \PYG{n}{latest\PYGZus{}finish}\PYG{o}{=}\PYG{n}{x}\PYG{p}{[}\PYG{l+m+mi}{5}\PYG{p}{])}
      \PYG{k}{for} \PYG{n}{x} \PYG{o+ow}{in} \PYG{n}{rides}
  \PYG{p}{]}
  \PYG{n}{cars} \PYG{o}{=} \PYG{p}{[(}\PYG{l+m+mi}{0}\PYG{p}{,} \PYG{p}{(}\PYG{l+m+mi}{0}\PYG{p}{,} \PYG{l+m+mi}{0}\PYG{p}{))]} \PYG{o}{*} \PYG{n}{f}
  \PYG{n}{score} \PYG{o}{=} \PYG{l+m+mi}{0}
  \PYG{k}{while} \PYG{k+kc}{True}\PYG{p}{:}
    \PYG{n}{time}\PYG{p}{,} \PYG{n}{coords} \PYG{o}{=} \PYG{n}{heapq}\PYG{o}{.}\PYG{n}{heappop}\PYG{p}{(}\PYG{n}{cars}\PYG{p}{)}
    \PYG{k}{if} \PYG{n}{time} \PYG{o}{\PYGZgt{}=} \PYG{n}{tot\PYGZus{}time}\PYG{p}{:}
      \PYG{k}{break}
    \PYG{n}{idx} \PYG{o}{=} \PYG{n}{pick\PYGZus{}ride}\PYG{p}{(}\PYG{n}{coords}\PYG{p}{,} \PYG{n}{time}\PYG{p}{,} \PYG{n+nb}{tuple}\PYG{p}{(}\PYG{n}{rides}\PYG{p}{))}
    \PYG{k}{assert} \PYG{n+nb}{isinstance}\PYG{p}{(}\PYG{n}{idx}\PYG{p}{,} \PYG{n+nb}{int}\PYG{p}{)}
    \PYG{k}{if} \PYG{n}{idx} \PYG{o}{\PYGZlt{}} \PYG{l+m+mi}{0} \PYG{o+ow}{or} \PYG{n}{idx} \PYG{o}{\PYGZgt{}=} \PYG{n+nb}{len}\PYG{p}{(}\PYG{n}{rides}\PYG{p}{):}
      \PYG{n}{heapq}\PYG{o}{.}\PYG{n}{heappush}\PYG{p}{(}\PYG{n}{cars}\PYG{p}{,} \PYG{p}{(}\PYG{n}{tot\PYGZus{}time}\PYG{p}{,} \PYG{n}{coords}\PYG{p}{))}
    \PYG{k}{else}\PYG{p}{:}
      \PYG{n}{r} \PYG{o}{=} \PYG{n}{rides}\PYG{o}{.}\PYG{n}{pop}\PYG{p}{(}\PYG{n}{idx}\PYG{p}{)}
      \PYG{n}{pickup\PYGZus{}time} \PYG{o}{=} \PYG{n}{time} \PYG{o}{+} \PYG{n}{r}\PYG{o}{.}\PYG{n}{distance\PYGZus{}to\PYGZus{}start}\PYG{p}{(}\PYG{n}{coords}\PYG{p}{)}
      \PYG{k}{if} \PYG{n}{pickup\PYGZus{}time} \PYG{o}{\PYGZlt{}} \PYG{n}{r}\PYG{o}{.}\PYG{n}{earliest\PYGZus{}start}\PYG{p}{:}
        \PYG{n}{pickup\PYGZus{}time} \PYG{o}{=} \PYG{n}{r}\PYG{o}{.}\PYG{n}{earliest\PYGZus{}start}
      \PYG{n}{free\PYGZus{}time} \PYG{o}{=} \PYG{n}{pickup\PYGZus{}time} \PYG{o}{+} \PYG{n}{r}\PYG{o}{.}\PYG{n}{length}\PYG{p}{()}
      \PYG{n}{new\PYGZus{}car} \PYG{o}{=} \PYG{p}{(}\PYG{n}{free\PYGZus{}time}\PYG{p}{,} \PYG{n}{r}\PYG{o}{.}\PYG{n}{end}\PYG{p}{)}
      \PYG{k}{if} \PYG{n}{free\PYGZus{}time} \PYG{o}{\PYGZlt{}=} \PYG{n}{r}\PYG{o}{.}\PYG{n}{latest\PYGZus{}finish}\PYG{p}{:}
        \PYG{n}{score} \PYG{o}{+=} \PYG{n}{r}\PYG{o}{.}\PYG{n}{length}\PYG{p}{()}
        \PYG{k}{if} \PYG{n}{pickup\PYGZus{}time} \PYG{o}{==} \PYG{n}{r}\PYG{o}{.}\PYG{n}{earliest\PYGZus{}start}\PYG{p}{:}
          \PYG{n}{score} \PYG{o}{+=} \PYG{n}{b}
      \PYG{n}{heapq}\PYG{o}{.}\PYG{n}{heappush}\PYG{p}{(}\PYG{n}{cars}\PYG{p}{,} \PYG{n}{new\PYGZus{}car}\PYG{p}{)}
  \PYG{k}{return} \PYG{n}{score}
\end{Verbatim}
\end{tiny}
\end{minipage}
\caption{The input parsing function and the greedy algorithm backbone for the 2018 Hash Code online qualification (Self-driving rides). Note that the backbone is calling \texttt{pick\_ride}---the function which chooses the next ride for a given car. We do not require a bespoke evaluation function in this case because it can be calculated on-the-fly during the execution.}
\label{fig:backbone_self}
\end{figure*}

\begin{figure*}
\centering
\begin{minipage}[c]{0.48\textwidth}
\begin{tiny}
\begin{Verbatim}[commandchars=\\\{\}]
\PYG{n+nd}{@dataclasses}\PYG{o}{.}\PYG{n}{dataclass}\PYG{p}{(}\PYG{n}{frozen}\PYG{o}{=}\PYG{k+kc}{True}\PYG{p}{)}
\PYG{k}{class} \PYG{n+nc}{Street}\PYG{p}{:}
  \PYG{n}{name}\PYG{p}{:} \PYG{n+nb}{str}
  \PYG{n}{length}\PYG{p}{:} \PYG{n+nb}{int}
  \PYG{n}{start\PYGZus{}id}\PYG{p}{:} \PYG{n+nb}{int}
  \PYG{n}{end\PYGZus{}id}\PYG{p}{:} \PYG{n+nb}{int}


\PYG{n+nd}{@dataclasses}\PYG{o}{.}\PYG{n}{dataclass}\PYG{p}{(}\PYG{n}{frozen}\PYG{o}{=}\PYG{k+kc}{True}\PYG{p}{)}
\PYG{k}{class} \PYG{n+nc}{Car}\PYG{p}{:}
  \PYG{n}{index}\PYG{p}{:} \PYG{n+nb}{int}
  \PYG{n}{route}\PYG{p}{:} \PYG{n+nb}{list}\PYG{p}{[}\PYG{n}{Street}\PYG{p}{]}


\PYG{n+nd}{@dataclasses}\PYG{o}{.}\PYG{n}{dataclass}\PYG{p}{(}\PYG{n}{frozen}\PYG{o}{=}\PYG{k+kc}{True}\PYG{p}{)}
\PYG{k}{class} \PYG{n+nc}{Intersection}\PYG{p}{:}
  \PYG{n}{index}\PYG{p}{:} \PYG{n+nb}{int}
  \PYG{n}{roads\PYGZus{}in}\PYG{p}{:} \PYG{n+nb}{list}\PYG{p}{[}\PYG{n}{Street}\PYG{p}{]}
  \PYG{n}{roads\PYGZus{}out}\PYG{p}{:} \PYG{n+nb}{list}\PYG{p}{[}\PYG{n}{Street}\PYG{p}{]}


\PYG{k}{def} \PYG{n+nf}{evaluate}\PYG{p}{(}\PYG{n}{file\PYGZus{}path}\PYG{p}{:} \PYG{n+nb}{str}\PYG{p}{)} \PYG{o}{\PYGZhy{}\PYGZgt{}} \PYG{n+nb}{int}\PYG{p}{:}
  \PYG{l+s+sd}{\PYGZdq{}\PYGZdq{}\PYGZdq{}Returns traffic scheduling value when using the evolved heuristic.\PYGZdq{}\PYGZdq{}\PYGZdq{}}
  \PYG{n}{intersections}\PYG{p}{,} \PYG{n}{streets}\PYG{p}{,} \PYG{n}{cars}\PYG{p}{,} \PYG{n}{deadline}\PYG{p}{,} \PYG{n}{bonus} \PYG{o}{=} \PYG{n}{parse\PYGZus{}data}\PYG{p}{(}\PYG{n}{file\PYGZus{}path}\PYG{p}{)}
  \PYG{n}{used\PYGZus{}streets} \PYG{o}{=} \PYG{p}{\PYGZob{}\PYGZcb{}}
  \PYG{c+c1}{\PYGZsh{} Make sure we do not schedule unused streets}
  \PYG{k}{for} \PYG{n}{car} \PYG{o+ow}{in} \PYG{n}{cars}\PYG{p}{:}
    \PYG{n}{travel\PYGZus{}time} \PYG{o}{=} \PYG{l+m+mi}{0}
    \PYG{k}{for} \PYG{n}{street} \PYG{o+ow}{in} \PYG{n}{car}\PYG{o}{.}\PYG{n}{route}\PYG{p}{[}\PYG{l+m+mi}{1}\PYG{p}{:]:}
      \PYG{n}{travel\PYGZus{}time} \PYG{o}{+=} \PYG{n}{street}\PYG{o}{.}\PYG{n}{length}
    \PYG{k}{if} \PYG{n}{travel\PYGZus{}time} \PYG{o}{\PYGZgt{}} \PYG{n}{deadline}\PYG{p}{:}
      \PYG{k}{continue}
    \PYG{k}{for} \PYG{n}{street} \PYG{o+ow}{in} \PYG{n}{car}\PYG{o}{.}\PYG{n}{route}\PYG{p}{[:}\PYG{o}{\PYGZhy{}}\PYG{l+m+mi}{1}\PYG{p}{]:}
      \PYG{k}{if} \PYG{n}{street}\PYG{o}{.}\PYG{n}{name} \PYG{o+ow}{not} \PYG{o+ow}{in} \PYG{n}{used\PYGZus{}streets}\PYG{p}{:}
        \PYG{n}{used\PYGZus{}streets}\PYG{p}{[}\PYG{n}{street}\PYG{o}{.}\PYG{n}{name}\PYG{p}{]} \PYG{o}{=} \PYG{l+m+mi}{0}
      \PYG{n}{used\PYGZus{}streets}\PYG{p}{[}\PYG{n}{street}\PYG{o}{.}\PYG{n}{name}\PYG{p}{]} \PYG{o}{+=} \PYG{l+m+mi}{1}
  \PYG{n}{curr\PYGZus{}index} \PYG{o}{=} \PYG{p}{[]}
  \PYG{n}{curr\PYGZus{}time\PYGZus{}at\PYGZus{}index} \PYG{o}{=} \PYG{p}{[]}
  \PYG{n}{deques} \PYG{o}{=} \PYG{p}{\PYGZob{}\PYGZcb{}}
  \PYG{n}{open\PYGZus{}streets} \PYG{o}{=} \PYG{n+nb}{set}\PYG{p}{(}\PYG{n}{used\PYGZus{}streets}\PYG{p}{)}
  \PYG{n}{cand\PYGZus{}streets} \PYG{o}{=} \PYG{n+nb}{set}\PYG{p}{()}
  \PYG{n}{schedule} \PYG{o}{=} \PYG{p}{[]}
  \PYG{k}{for} \PYG{n}{\PYGZus{}} \PYG{o+ow}{in} \PYG{n+nb}{range}\PYG{p}{(}\PYG{n+nb}{len}\PYG{p}{(}\PYG{n}{intersections}\PYG{p}{)):}
    \PYG{n}{schedule}\PYG{o}{.}\PYG{n}{append}\PYG{p}{([])}
  \PYG{k}{for} \PYG{n}{name} \PYG{o+ow}{in} \PYG{n}{used\PYGZus{}streets}\PYG{p}{:}
    \PYG{n}{schedule}\PYG{p}{[}\PYG{n}{streets}\PYG{p}{[}\PYG{n}{name}\PYG{p}{]}\PYG{o}{.}\PYG{n}{end\PYGZus{}id}\PYG{p}{]}\PYG{o}{.}\PYG{n}{append}\PYG{p}{(}\PYG{k+kc}{None}\PYG{p}{)}
    \PYG{n}{deques}\PYG{p}{[}\PYG{n}{name}\PYG{p}{]} \PYG{o}{=} \PYG{n}{collections}\PYG{o}{.}\PYG{n}{deque}\PYG{p}{()}
  \PYG{c+c1}{\PYGZsh{} Initialise intersections and empty queues}
  \PYG{k}{assert} \PYG{n+nb}{len}\PYG{p}{(}\PYG{n}{intersections}\PYG{p}{)} \PYG{o}{==} \PYG{n+nb}{len}\PYG{p}{(}\PYG{n}{schedule}\PYG{p}{)}
  \PYG{k}{for} \PYG{n}{i\PYGZus{}ind} \PYG{o+ow}{in} \PYG{n+nb}{range}\PYG{p}{(}\PYG{n+nb}{len}\PYG{p}{(}\PYG{n}{intersections}\PYG{p}{)):}
    \PYG{k}{if} \PYG{n}{schedule}\PYG{p}{[}\PYG{n}{i\PYGZus{}ind}\PYG{p}{]:}
      \PYG{n}{curr\PYGZus{}index}\PYG{o}{.}\PYG{n}{append}\PYG{p}{(}\PYG{l+m+mi}{0}\PYG{p}{)}
      \PYG{n}{curr\PYGZus{}time\PYGZus{}at\PYGZus{}index}\PYG{o}{.}\PYG{n}{append}\PYG{p}{(}\PYG{l+m+mi}{1}\PYG{p}{)}
    \PYG{k}{else}\PYG{p}{:}
      \PYG{n}{curr\PYGZus{}index}\PYG{o}{.}\PYG{n}{append}\PYG{p}{(}\PYG{k+kc}{None}\PYG{p}{)}
      \PYG{n}{curr\PYGZus{}time\PYGZus{}at\PYGZus{}index}\PYG{o}{.}\PYG{n}{append}\PYG{p}{(}\PYG{n}{deadline} \PYG{o}{+} \PYG{l+m+mi}{1}\PYG{p}{)}
  \PYG{c+c1}{\PYGZsh{} Initialise queues}
  \PYG{k}{for} \PYG{n}{car} \PYG{o+ow}{in} \PYG{n}{cars}\PYG{p}{:}
    \PYG{k}{if} \PYG{n}{car}\PYG{o}{.}\PYG{n}{route}\PYG{p}{[}\PYG{l+m+mi}{0}\PYG{p}{]}\PYG{o}{.}\PYG{n}{name} \PYG{o+ow}{not} \PYG{o+ow}{in} \PYG{n}{used\PYGZus{}streets}\PYG{p}{:}
      \PYG{n}{deques}\PYG{p}{[}\PYG{n}{car}\PYG{o}{.}\PYG{n}{route}\PYG{p}{[}\PYG{l+m+mi}{0}\PYG{p}{]}\PYG{o}{.}\PYG{n}{name}\PYG{p}{]} \PYG{o}{=} \PYG{n}{collections}\PYG{o}{.}\PYG{n}{deque}\PYG{p}{()}
    \PYG{n}{deques}\PYG{p}{[}\PYG{n}{car}\PYG{o}{.}\PYG{n}{route}\PYG{p}{[}\PYG{l+m+mi}{0}\PYG{p}{]}\PYG{o}{.}\PYG{n}{name}\PYG{p}{]}\PYG{o}{.}\PYG{n}{append}\PYG{p}{(}\PYG{n}{car}\PYG{o}{.}\PYG{n}{index}\PYG{p}{)}
    \PYG{n}{cand\PYGZus{}streets}\PYG{o}{.}\PYG{n}{add}\PYG{p}{(}\PYG{n}{car}\PYG{o}{.}\PYG{n}{route}\PYG{p}{[}\PYG{l+m+mi}{0}\PYG{p}{]}\PYG{o}{.}\PYG{n}{name}\PYG{p}{)}
  \PYG{c+c1}{\PYGZsh{} Initialise cars}
  \PYG{n}{curr\PYGZus{}road\PYGZus{}id} \PYG{o}{=} \PYG{p}{[}\PYG{l+m+mi}{0}\PYG{p}{]} \PYG{o}{*} \PYG{n+nb}{len}\PYG{p}{(}\PYG{n}{cars}\PYG{p}{)}
  \PYG{n}{curr\PYGZus{}travel} \PYG{o}{=} \PYG{p}{[}\PYG{l+m+mi}{0}\PYG{p}{]} \PYG{o}{*} \PYG{n+nb}{len}\PYG{p}{(}\PYG{n}{cars}\PYG{p}{)}
  \PYG{n}{moving\PYGZus{}cars} \PYG{o}{=} \PYG{n+nb}{set}\PYG{p}{()}
  \PYG{n}{score} \PYG{o}{=} \PYG{l+m+mi}{0}
  \PYG{c+c1}{\PYGZsh{} Perform a simulation, scheduling streets as they are hit}
  \PYG{k}{for} \PYG{n}{time} \PYG{o+ow}{in} \PYG{n+nb}{range}\PYG{p}{(}\PYG{n}{deadline}\PYG{p}{):}
    \PYG{c+c1}{\PYGZsh{} process any streets that are now open}
    \PYG{k}{for} \PYG{n}{s\PYGZus{}name} \PYG{o+ow}{in} \PYG{n+nb}{set}\PYG{p}{(}\PYG{n}{open\PYGZus{}streets}\PYG{p}{)}\PYG{o}{.}\PYG{n}{intersection}\PYG{p}{(}\PYG{n}{cand\PYGZus{}streets}\PYG{p}{):}
      \PYG{n}{i\PYGZus{}ind} \PYG{o}{=} \PYG{n}{streets}\PYG{p}{[}\PYG{n}{s\PYGZus{}name}\PYG{p}{]}\PYG{o}{.}\PYG{n}{end\PYGZus{}id}
      \PYG{n}{s\PYGZus{}pos} \PYG{o}{=} \PYG{n}{pos\PYGZus{}greedy}\PYG{p}{(}
          \PYG{n}{streets}\PYG{p}{[}\PYG{n}{s\PYGZus{}name}\PYG{p}{],}
          \PYG{n}{cars}\PYG{p}{,}
          \PYG{n}{intersections}\PYG{p}{,}
          \PYG{n}{used\PYGZus{}streets}\PYG{p}{,}
          \PYG{n}{bonus}\PYG{p}{,}
          \PYG{n}{time}\PYG{p}{,}
          \PYG{n+nb}{len}\PYG{p}{(}\PYG{n}{schedule}\PYG{p}{[}\PYG{n}{i\PYGZus{}ind}\PYG{p}{]),}
          \PYG{k+kc}{True}\PYG{p}{)}
      \PYG{n}{s\PYGZus{}pos} \PYG{o}{=} \PYG{n}{s\PYGZus{}pos} \PYG{o}{\PYGZpc{}} \PYG{n+nb}{len}\PYG{p}{(}\PYG{n}{schedule}\PYG{p}{[}\PYG{n}{i\PYGZus{}ind}\PYG{p}{])}
      \PYG{k}{while} \PYG{n}{schedule}\PYG{p}{[}\PYG{n}{i\PYGZus{}ind}\PYG{p}{][}\PYG{n}{s\PYGZus{}pos}\PYG{p}{]} \PYG{o+ow}{is} \PYG{o+ow}{not} \PYG{k+kc}{None}\PYG{p}{:}
        \PYG{n}{s\PYGZus{}pos} \PYG{o}{=} \PYG{p}{(}\PYG{n}{s\PYGZus{}pos} \PYG{o}{+} \PYG{l+m+mi}{1}\PYG{p}{)} \PYG{o}{\PYGZpc{}} \PYG{n+nb}{len}\PYG{p}{(}\PYG{n}{schedule}\PYG{p}{[}\PYG{n}{i\PYGZus{}ind}\PYG{p}{])}
      \PYG{n}{schedule}\PYG{p}{[}\PYG{n}{i\PYGZus{}ind}\PYG{p}{][}\PYG{n}{s\PYGZus{}pos}\PYG{p}{]} \PYG{o}{=} \PYG{p}{(}\PYG{n}{streets}\PYG{p}{[}\PYG{n}{s\PYGZus{}name}\PYG{p}{],} \PYG{l+m+mi}{1}\PYG{p}{)}
      \PYG{n}{open\PYGZus{}streets}\PYG{o}{.}\PYG{n}{remove}\PYG{p}{(}\PYG{n}{s\PYGZus{}name}\PYG{p}{)}
\end{Verbatim}
\end{tiny}
\end{minipage}
\begin{minipage}[c]{0.48\textwidth}
\begin{tiny}
\begin{Verbatim}[commandchars=\\\{\}]
    \PYG{c+c1}{\PYGZsh{} unblock any available traffic}
    \PYG{k}{for} \PYG{n}{i\PYGZus{}ind} \PYG{o+ow}{in} \PYG{n+nb}{range}\PYG{p}{(}\PYG{n+nb}{len}\PYG{p}{(}\PYG{n}{intersections}\PYG{p}{)):}
      \PYG{k}{if} \PYG{n}{curr\PYGZus{}index}\PYG{p}{[}\PYG{n}{i\PYGZus{}ind}\PYG{p}{]} \PYG{o+ow}{is} \PYG{k+kc}{None}\PYG{p}{:}
        \PYG{k}{continue}
      \PYG{k}{if} \PYG{n}{schedule}\PYG{p}{[}\PYG{n}{i\PYGZus{}ind}\PYG{p}{][}\PYG{n}{curr\PYGZus{}index}\PYG{p}{[}\PYG{n}{i\PYGZus{}ind}\PYG{p}{]]} \PYG{o+ow}{is} \PYG{k+kc}{None}\PYG{p}{:}
        \PYG{c+c1}{\PYGZsh{} unlikely, but could happen under more eccentric schedulers...}
        \PYG{k}{continue}
      \PYG{n}{curr\PYGZus{}open\PYGZus{}street} \PYG{o}{=} \PYG{n}{schedule}\PYG{p}{[}\PYG{n}{i\PYGZus{}ind}\PYG{p}{][}\PYG{n}{curr\PYGZus{}index}\PYG{p}{[}\PYG{n}{i\PYGZus{}ind}\PYG{p}{]][}\PYG{l+m+mi}{0}\PYG{p}{]}\PYG{o}{.}\PYG{n}{name}
      \PYG{k}{if} \PYG{n}{deques}\PYG{p}{[}\PYG{n}{curr\PYGZus{}open\PYGZus{}street}\PYG{p}{]:}
        \PYG{n}{top\PYGZus{}car\PYGZus{}id} \PYG{o}{=} \PYG{n}{deques}\PYG{p}{[}\PYG{n}{curr\PYGZus{}open\PYGZus{}street}\PYG{p}{]}\PYG{o}{.}\PYG{n}{popleft}\PYG{p}{()}
        \PYG{n}{moving\PYGZus{}cars}\PYG{o}{.}\PYG{n}{add}\PYG{p}{(}\PYG{n}{top\PYGZus{}car\PYGZus{}id}\PYG{p}{)}
        \PYG{n}{curr\PYGZus{}road\PYGZus{}id}\PYG{p}{[}\PYG{n}{top\PYGZus{}car\PYGZus{}id}\PYG{p}{]} \PYG{o}{+=} \PYG{l+m+mi}{1}
        \PYG{n}{curr\PYGZus{}travel}\PYG{p}{[}\PYG{n}{top\PYGZus{}car\PYGZus{}id}\PYG{p}{]} \PYG{o}{=} \PYG{n}{cars}\PYG{p}{[}\PYG{n}{top\PYGZus{}car\PYGZus{}id}\PYG{p}{]}\PYG{o}{.}\PYG{n}{route}\PYG{p}{[}
            \PYG{n}{curr\PYGZus{}road\PYGZus{}id}\PYG{p}{[}\PYG{n}{top\PYGZus{}car\PYGZus{}id}\PYG{p}{]]}\PYG{o}{.}\PYG{n}{length}
    \PYG{c+c1}{\PYGZsh{} progress any moving traffic}
    \PYG{k}{for} \PYG{n}{car\PYGZus{}id} \PYG{o+ow}{in} \PYG{n+nb}{set}\PYG{p}{(}\PYG{n}{moving\PYGZus{}cars}\PYG{p}{):}
      \PYG{k}{assert} \PYG{n}{curr\PYGZus{}travel}\PYG{p}{[}\PYG{n}{car\PYGZus{}id}\PYG{p}{]} \PYG{o}{\PYGZgt{}} \PYG{l+m+mi}{0}
      \PYG{n}{curr\PYGZus{}travel}\PYG{p}{[}\PYG{n}{car\PYGZus{}id}\PYG{p}{]} \PYG{o}{\PYGZhy{}=} \PYG{l+m+mi}{1}
      \PYG{k}{if} \PYG{n}{curr\PYGZus{}travel}\PYG{p}{[}\PYG{n}{car\PYGZus{}id}\PYG{p}{]} \PYG{o}{==} \PYG{l+m+mi}{0}\PYG{p}{:}
        \PYG{n}{moving\PYGZus{}cars}\PYG{o}{.}\PYG{n}{remove}\PYG{p}{(}\PYG{n}{car\PYGZus{}id}\PYG{p}{)}
        \PYG{k}{if} \PYG{n}{curr\PYGZus{}road\PYGZus{}id}\PYG{p}{[}\PYG{n}{car\PYGZus{}id}\PYG{p}{]} \PYG{o}{==} \PYG{n+nb}{len}\PYG{p}{(}\PYG{n}{cars}\PYG{p}{[}\PYG{n}{car\PYGZus{}id}\PYG{p}{]}\PYG{o}{.}\PYG{n}{route}\PYG{p}{)} \PYG{o}{\PYGZhy{}} \PYG{l+m+mi}{1}\PYG{p}{:}
          \PYG{n}{score} \PYG{o}{+=} \PYG{n}{bonus} \PYG{o}{+} \PYG{p}{(}\PYG{n}{deadline} \PYG{o}{\PYGZhy{}} \PYG{n}{time} \PYG{o}{\PYGZhy{}} \PYG{l+m+mi}{1}\PYG{p}{)}
        \PYG{k}{else}\PYG{p}{:}
          \PYG{n}{deques}\PYG{p}{[}\PYG{n}{cars}\PYG{p}{[}\PYG{n}{car\PYGZus{}id}\PYG{p}{]}\PYG{o}{.}\PYG{n}{route}\PYG{p}{[}\PYG{n}{curr\PYGZus{}road\PYGZus{}id}\PYG{p}{[}\PYG{n}{car\PYGZus{}id}\PYG{p}{]]}\PYG{o}{.}\PYG{n}{name}\PYG{p}{]}\PYG{o}{.}\PYG{n}{append}\PYG{p}{(}\PYG{n}{car\PYGZus{}id}\PYG{p}{)}
          \PYG{n}{cand\PYGZus{}streets}\PYG{o}{.}\PYG{n}{add}\PYG{p}{(}\PYG{n}{cars}\PYG{p}{[}\PYG{n}{car\PYGZus{}id}\PYG{p}{]}\PYG{o}{.}\PYG{n}{route}\PYG{p}{[}\PYG{n}{curr\PYGZus{}road\PYGZus{}id}\PYG{p}{[}\PYG{n}{car\PYGZus{}id}\PYG{p}{]]}\PYG{o}{.}\PYG{n}{name}\PYG{p}{)}
    \PYG{c+c1}{\PYGZsh{} process traffic lights}
    \PYG{k}{for} \PYG{n}{i\PYGZus{}ind} \PYG{o+ow}{in} \PYG{n+nb}{range}\PYG{p}{(}\PYG{n+nb}{len}\PYG{p}{(}\PYG{n}{intersections}\PYG{p}{)):}
      \PYG{n}{curr\PYGZus{}time\PYGZus{}at\PYGZus{}index}\PYG{p}{[}\PYG{n}{i\PYGZus{}ind}\PYG{p}{]} \PYG{o}{\PYGZhy{}=} \PYG{l+m+mi}{1}
      \PYG{k}{if} \PYG{n}{curr\PYGZus{}time\PYGZus{}at\PYGZus{}index}\PYG{p}{[}\PYG{n}{i\PYGZus{}ind}\PYG{p}{]} \PYG{o}{==} \PYG{l+m+mi}{0}\PYG{p}{:}
        \PYG{k}{assert} \PYG{n}{curr\PYGZus{}index}\PYG{p}{[}\PYG{n}{i\PYGZus{}ind}\PYG{p}{]} \PYG{o+ow}{is} \PYG{o+ow}{not} \PYG{k+kc}{None}
        \PYG{n}{curr\PYGZus{}index}\PYG{p}{[}\PYG{n}{i\PYGZus{}ind}\PYG{p}{]} \PYG{o}{=} \PYG{p}{(}\PYG{n}{curr\PYGZus{}index}\PYG{p}{[}\PYG{n}{i\PYGZus{}ind}\PYG{p}{]} \PYG{o}{+} \PYG{l+m+mi}{1}\PYG{p}{)} \PYG{o}{\PYGZpc{}} \PYG{n+nb}{len}\PYG{p}{(}\PYG{n}{schedule}\PYG{p}{[}\PYG{n}{i\PYGZus{}ind}\PYG{p}{])}
        \PYG{n}{curr\PYGZus{}time\PYGZus{}at\PYGZus{}index}\PYG{p}{[}\PYG{n}{i\PYGZus{}ind}\PYG{p}{]} \PYG{o}{=} \PYG{l+m+mi}{1}
  \PYG{c+c1}{\PYGZsh{} Schedule any unused streets.}
  \PYG{k}{for} \PYG{n}{s\PYGZus{}name} \PYG{o+ow}{in} \PYG{n}{used\PYGZus{}streets}\PYG{p}{:}
    \PYG{k}{if} \PYG{n}{s\PYGZus{}name} \PYG{o+ow}{in} \PYG{n}{open\PYGZus{}streets}\PYG{p}{:}
      \PYG{n}{i\PYGZus{}ind} \PYG{o}{=} \PYG{n}{streets}\PYG{p}{[}\PYG{n}{s\PYGZus{}name}\PYG{p}{]}\PYG{o}{.}\PYG{n}{end\PYGZus{}id}
      \PYG{n}{s\PYGZus{}pos} \PYG{o}{=} \PYG{n}{pos\PYGZus{}greedy}\PYG{p}{(}
          \PYG{n}{streets}\PYG{p}{[}\PYG{n}{s\PYGZus{}name}\PYG{p}{],}
          \PYG{n}{cars}\PYG{p}{,}
          \PYG{n}{intersections}\PYG{p}{,}
          \PYG{n}{used\PYGZus{}streets}\PYG{p}{,}
          \PYG{n}{bonus}\PYG{p}{,}
          \PYG{n}{deadline}\PYG{p}{,}
          \PYG{n+nb}{len}\PYG{p}{(}\PYG{n}{schedule}\PYG{p}{[}\PYG{n}{i\PYGZus{}ind}\PYG{p}{]),}
          \PYG{k+kc}{True}\PYG{p}{)}
      \PYG{n}{s\PYGZus{}pos} \PYG{o}{=} \PYG{n}{s\PYGZus{}pos} \PYG{o}{\PYGZpc{}} \PYG{n+nb}{len}\PYG{p}{(}\PYG{n}{schedule}\PYG{p}{[}\PYG{n}{i\PYGZus{}ind}\PYG{p}{])}
      \PYG{k}{while} \PYG{n}{schedule}\PYG{p}{[}\PYG{n}{i\PYGZus{}ind}\PYG{p}{][}\PYG{n}{s\PYGZus{}pos}\PYG{p}{]} \PYG{o+ow}{is} \PYG{o+ow}{not} \PYG{k+kc}{None}\PYG{p}{:}
        \PYG{n}{s\PYGZus{}pos} \PYG{o}{=} \PYG{p}{(}\PYG{n}{s\PYGZus{}pos} \PYG{o}{+} \PYG{l+m+mi}{1}\PYG{p}{)} \PYG{o}{\PYGZpc{}} \PYG{n+nb}{len}\PYG{p}{(}\PYG{n}{schedule}\PYG{p}{[}\PYG{n}{i\PYGZus{}ind}\PYG{p}{])}
      \PYG{n}{schedule}\PYG{p}{[}\PYG{n}{i\PYGZus{}ind}\PYG{p}{][}\PYG{n}{s\PYGZus{}pos}\PYG{p}{]} \PYG{o}{=} \PYG{p}{(}\PYG{n}{streets}\PYG{p}{[}\PYG{n}{s\PYGZus{}name}\PYG{p}{],} \PYG{l+m+mi}{1}\PYG{p}{)}
  \PYG{c+c1}{\PYGZsh{} Assign the durations for each green light.}
  \PYG{k}{for} \PYG{n}{i\PYGZus{}ind} \PYG{o+ow}{in} \PYG{n+nb}{range}\PYG{p}{(}\PYG{n+nb}{len}\PYG{p}{(}\PYG{n}{schedule}\PYG{p}{)):}
    \PYG{k}{for} \PYG{n}{s\PYGZus{}pos} \PYG{o+ow}{in} \PYG{n+nb}{range}\PYG{p}{(}\PYG{n+nb}{len}\PYG{p}{(}\PYG{n}{schedule}\PYG{p}{[}\PYG{n}{i\PYGZus{}ind}\PYG{p}{])):}
      \PYG{n}{l\PYGZus{}len} \PYG{o}{=} \PYG{n}{pos\PYGZus{}greedy}\PYG{p}{(}
          \PYG{n}{schedule}\PYG{p}{[}\PYG{n}{i\PYGZus{}ind}\PYG{p}{][}\PYG{n}{s\PYGZus{}pos}\PYG{p}{][}\PYG{l+m+mi}{0}\PYG{p}{],}
          \PYG{n}{cars}\PYG{p}{,}
          \PYG{n}{intersections}\PYG{p}{,}
          \PYG{n}{used\PYGZus{}streets}\PYG{p}{,}
          \PYG{n}{bonus}\PYG{p}{,}
          \PYG{n}{deadline}\PYG{p}{,}
          \PYG{n+nb}{len}\PYG{p}{(}\PYG{n}{schedule}\PYG{p}{[}\PYG{n}{i\PYGZus{}ind}\PYG{p}{]),}
          \PYG{k+kc}{False}\PYG{p}{)}
      \PYG{n}{schedule}\PYG{p}{[}\PYG{n}{i\PYGZus{}ind}\PYG{p}{][}\PYG{n}{s\PYGZus{}pos}\PYG{p}{]} \PYG{o}{=} \PYG{p}{(}\PYG{n}{schedule}\PYG{p}{[}\PYG{n}{i\PYGZus{}ind}\PYG{p}{][}\PYG{n}{s\PYGZus{}pos}\PYG{p}{][}\PYG{l+m+mi}{0}\PYG{p}{],} \PYG{n}{l\PYGZus{}len}\PYG{p}{)}
  \PYG{c+c1}{\PYGZsh{} Make sure we do not count streets with only unfinished cars}
  \PYG{n}{\PYGZus{}}\PYG{p}{,} \PYG{n}{fin\PYGZus{}cars} \PYG{o}{=} \PYG{n}{count\PYGZus{}blocked}\PYG{p}{(}
            \PYG{n}{streets}\PYG{p}{,}
            \PYG{n}{intersections}\PYG{p}{,}
            \PYG{n}{cars}\PYG{p}{,}
            \PYG{n}{schedule}\PYG{p}{,}
            \PYG{n}{deadline}\PYG{p}{,}
            \PYG{n}{bonus}\PYG{p}{,}
            \PYG{n}{used\PYGZus{}streets}\PYG{p}{)}
  \PYG{n}{n\PYGZus{}used\PYGZus{}streets} \PYG{o}{=} \PYG{p}{\PYGZob{}\PYGZcb{}}
  \PYG{k}{for} \PYG{n}{car} \PYG{o+ow}{in} \PYG{n}{cars}\PYG{p}{:}
    \PYG{k}{if} \PYG{n}{car}\PYG{o}{.}\PYG{n}{index} \PYG{o+ow}{in} \PYG{n}{fin\PYGZus{}cars}\PYG{p}{:}
      \PYG{k}{for} \PYG{n}{street} \PYG{o+ow}{in} \PYG{n}{car}\PYG{o}{.}\PYG{n}{route}\PYG{p}{[:}\PYG{o}{\PYGZhy{}}\PYG{l+m+mi}{1}\PYG{p}{]:}
        \PYG{k}{if} \PYG{n}{street}\PYG{o}{.}\PYG{n}{name} \PYG{o+ow}{not} \PYG{o+ow}{in} \PYG{n}{n\PYGZus{}used\PYGZus{}streets}\PYG{p}{:}
          \PYG{n}{n\PYGZus{}used\PYGZus{}streets}\PYG{p}{[}\PYG{n}{street}\PYG{o}{.}\PYG{n}{name}\PYG{p}{]} \PYG{o}{=} \PYG{l+m+mi}{0}
        \PYG{n}{n\PYGZus{}used\PYGZus{}streets}\PYG{p}{[}\PYG{n}{street}\PYG{o}{.}\PYG{n}{name}\PYG{p}{]} \PYG{o}{+=} \PYG{l+m+mi}{1}
  \PYG{n}{n\PYGZus{}schedule} \PYG{o}{=} \PYG{p}{[]}
  \PYG{k}{for} \PYG{n}{i\PYGZus{}ind} \PYG{o+ow}{in} \PYG{n+nb}{range}\PYG{p}{(}\PYG{n+nb}{len}\PYG{p}{(}\PYG{n}{schedule}\PYG{p}{)):}
    \PYG{n}{n\PYGZus{}schedule}\PYG{o}{.}\PYG{n}{append}\PYG{p}{([])}
    \PYG{k}{for} \PYG{n}{s\PYGZus{}ind} \PYG{o+ow}{in} \PYG{n+nb}{range}\PYG{p}{(}\PYG{n+nb}{len}\PYG{p}{(}\PYG{n}{schedule}\PYG{p}{[}\PYG{n}{i\PYGZus{}ind}\PYG{p}{])):}
      \PYG{n}{street} \PYG{o}{=} \PYG{n}{schedule}\PYG{p}{[}\PYG{n}{i\PYGZus{}ind}\PYG{p}{][}\PYG{n}{s\PYGZus{}ind}\PYG{p}{][}\PYG{l+m+mi}{0}\PYG{p}{]}
      \PYG{k}{if} \PYG{n}{street}\PYG{o}{.}\PYG{n}{name} \PYG{o+ow}{in} \PYG{n}{n\PYGZus{}used\PYGZus{}streets}\PYG{p}{:}
        \PYG{n}{n\PYGZus{}schedule}\PYG{p}{[}\PYG{n}{i\PYGZus{}ind}\PYG{p}{]}\PYG{o}{.}\PYG{n}{append}\PYG{p}{(}\PYG{n}{schedule}\PYG{p}{[}\PYG{n}{i\PYGZus{}ind}\PYG{p}{][}\PYG{n}{s\PYGZus{}ind}\PYG{p}{])}

  \PYG{k}{return} \PYG{n}{get\PYGZus{}value}\PYG{p}{(}
      \PYG{n}{streets}\PYG{p}{,} \PYG{n}{intersections}\PYG{p}{,} \PYG{n}{cars}\PYG{p}{,} \PYG{n}{n\PYGZus{}schedule}\PYG{p}{,} \PYG{n}{deadline}\PYG{p}{,} \PYG{n}{bonus}\PYG{p}{,} \PYG{n}{file\PYGZus{}path}\PYG{p}{)}
\end{Verbatim}
\end{tiny}
\end{minipage}
\caption{A snippet of the greedy algorithm backbone for the 2021 Hash Code online qualification (Traffic scheduling). Note that the backbone is calling \texttt{pos\_greedy}---the function which chooses either the position of a given street in the traffic light's schedule (if the final boolean is \texttt{True}) or its green light duration (if it is \texttt{False}). It also calls \texttt{count\_blocked}, a function which returns all cars that successfully completed their route, and \texttt{get\_value}, the evaluation function.}
\label{fig:backbone_traf}
\end{figure*}

\begin{figure*}
\centering
\begin{minipage}[c]{0.48\textwidth}
\begin{tiny}
\begin{Verbatim}[commandchars=\\\{\}]
\PYG{k+kn}{import} \PYG{n+nn}{dataclasses}
\PYG{k+kn}{import} \PYG{n+nn}{heapq}
\PYG{k+kn}{import} \PYG{n+nn}{random}
\PYG{k+kn}{import} \PYG{n+nn}{numpy} \PYG{k}{as} \PYG{n+nn}{np}


\PYG{n+nd}{@dataclasses}\PYG{o}{.}\PYG{n}{dataclass}\PYG{p}{(}\PYG{n}{frozen}\PYG{o}{=}\PYG{k+kc}{True}\PYG{p}{)}
\PYG{k}{class} \PYG{n+nc}{Grid}\PYG{p}{:}
  \PYG{n}{rows}\PYG{p}{:} \PYG{n+nb}{int}
  \PYG{n}{cols}\PYG{p}{:} \PYG{n+nb}{int}
  \PYG{n}{mackerels}\PYG{p}{:} \PYG{n+nb}{list}\PYG{p}{[}\PYG{n+nb}{list}\PYG{p}{[}\PYG{n+nb}{int}\PYG{p}{]]}
  \PYG{n}{sardines}\PYG{p}{:} \PYG{n+nb}{list}\PYG{p}{[}\PYG{n+nb}{list}\PYG{p}{[}\PYG{n+nb}{int}\PYG{p}{]]}


\PYG{k}{def} \PYG{n+nf}{parse\PYGZus{}data}\PYG{p}{(}\PYG{n}{file\PYGZus{}path}\PYG{p}{:} \PYG{n+nb}{str}\PYG{p}{):}
  \PYG{l+s+sd}{\PYGZdq{}\PYGZdq{}\PYGZdq{}Parse data file.}

\PYG{l+s+sd}{  Args:}
\PYG{l+s+sd}{    file\PYGZus{}path: path to the datafile.}

\PYG{l+s+sd}{  Returns:}
\PYG{l+s+sd}{    mackerels}
\PYG{l+s+sd}{    sardines}
\PYG{l+s+sd}{  \PYGZdq{}\PYGZdq{}\PYGZdq{}}
  \PYG{k}{with} \PYG{n+nb}{open}\PYG{p}{(}\PYG{n}{file\PYGZus{}path}\PYG{p}{,} \PYG{l+s+s1}{\PYGZsq{}rt\PYGZsq{}}\PYG{p}{)} \PYG{k}{as} \PYG{n}{f}\PYG{p}{:}
    \PYG{n}{n\PYGZus{}fishes} \PYG{o}{=} \PYG{n+nb}{int}\PYG{p}{(}\PYG{n}{f}\PYG{o}{.}\PYG{n}{readline}\PYG{p}{())}
    \PYG{n}{mackerels} \PYG{o}{=} \PYG{p}{\PYGZob{}\PYGZcb{}}
    \PYG{n}{sardines} \PYG{o}{=} \PYG{p}{\PYGZob{}\PYGZcb{}}
    \PYG{k}{for} \PYG{n}{\PYGZus{}} \PYG{o+ow}{in} \PYG{n+nb}{range}\PYG{p}{(}\PYG{n}{n\PYGZus{}fishes}\PYG{p}{):}
      \PYG{n}{coord\PYGZus{}line} \PYG{o}{=} \PYG{n}{f}\PYG{o}{.}\PYG{n}{readline}\PYG{p}{()}
      \PYG{n}{x}\PYG{p}{,} \PYG{n}{y} \PYG{o}{=} \PYG{n+nb}{map}\PYG{p}{(}\PYG{n+nb}{int}\PYG{p}{,} \PYG{n}{coord\PYGZus{}line}\PYG{o}{.}\PYG{n}{split}\PYG{p}{())}
      \PYG{n}{mackerels}\PYG{p}{[(}\PYG{n}{x}\PYG{p}{,} \PYG{n}{y}\PYG{p}{)]} \PYG{o}{=} \PYG{l+m+mi}{1}
    \PYG{k}{for} \PYG{n}{\PYGZus{}} \PYG{o+ow}{in} \PYG{n+nb}{range}\PYG{p}{(}\PYG{n}{n\PYGZus{}fishes}\PYG{p}{):}
      \PYG{n}{coord\PYGZus{}line} \PYG{o}{=} \PYG{n}{f}\PYG{o}{.}\PYG{n}{readline}\PYG{p}{()}
      \PYG{n}{x}\PYG{p}{,} \PYG{n}{y} \PYG{o}{=} \PYG{n+nb}{map}\PYG{p}{(}\PYG{n+nb}{int}\PYG{p}{,} \PYG{n}{coord\PYGZus{}line}\PYG{o}{.}\PYG{n}{split}\PYG{p}{())}
      \PYG{n}{sardines}\PYG{p}{[(}\PYG{n}{x}\PYG{p}{,} \PYG{n}{y}\PYG{p}{)]} \PYG{o}{=} \PYG{l+m+mi}{1}

\PYG{k}{return} \PYG{n}{mackerels}\PYG{p}{,} \PYG{n}{sardines}


\PYG{k}{def} \PYG{n+nf}{catch\PYGZus{}value}\PYG{p}{(}\PYG{n}{file\PYGZus{}path}\PYG{p}{:} \PYG{n+nb}{str}\PYG{p}{,} \PYG{n}{cell\PYGZus{}size}\PYG{p}{:} \PYG{n+nb}{int}\PYG{p}{)} \PYG{o}{\PYGZhy{}\PYGZgt{}} \PYG{n+nb}{int}\PYG{p}{:}
  \PYG{l+s+sd}{\PYGZdq{}\PYGZdq{}\PYGZdq{}Returns catch value for a given cell size.\PYGZdq{}\PYGZdq{}\PYGZdq{}}
  \PYG{n}{mackerels}\PYG{p}{,} \PYG{n}{sardines} \PYG{o}{=} \PYG{n}{parse\PYGZus{}data}\PYG{p}{(}\PYG{n}{file\PYGZus{}path}\PYG{p}{)}
  \PYG{n}{min\PYGZus{}x}\PYG{p}{,} \PYG{n}{max\PYGZus{}x} \PYG{o}{=} \PYG{l+m+mi}{100\PYGZus{}000}\PYG{p}{,} \PYG{l+m+mi}{0}
  \PYG{n}{min\PYGZus{}y}\PYG{p}{,} \PYG{n}{max\PYGZus{}y} \PYG{o}{=} \PYG{l+m+mi}{100\PYGZus{}000}\PYG{p}{,} \PYG{l+m+mi}{0}
  \PYG{k}{for} \PYG{p}{(}\PYG{n}{x}\PYG{p}{,} \PYG{n}{y}\PYG{p}{)} \PYG{o+ow}{in} \PYG{n}{mackerels}\PYG{p}{:}
    \PYG{n}{min\PYGZus{}x} \PYG{o}{=} \PYG{n+nb}{min}\PYG{p}{(}\PYG{n}{min\PYGZus{}x}\PYG{p}{,} \PYG{n}{x}\PYG{p}{)}
    \PYG{n}{max\PYGZus{}x} \PYG{o}{=} \PYG{n+nb}{max}\PYG{p}{(}\PYG{n}{max\PYGZus{}x}\PYG{p}{,} \PYG{n}{x}\PYG{p}{)}
    \PYG{n}{min\PYGZus{}y} \PYG{o}{=} \PYG{n+nb}{min}\PYG{p}{(}\PYG{n}{min\PYGZus{}y}\PYG{p}{,} \PYG{n}{y}\PYG{p}{)}
    \PYG{n}{max\PYGZus{}y} \PYG{o}{=} \PYG{n+nb}{max}\PYG{p}{(}\PYG{n}{max\PYGZus{}y}\PYG{p}{,} \PYG{n}{y}\PYG{p}{)}

  \PYG{n}{rows} \PYG{o}{=} \PYG{l+m+mi}{0}
  \PYG{n}{cols} \PYG{o}{=} \PYG{l+m+mi}{0}
  \PYG{n}{grid\PYGZus{}mackerels} \PYG{o}{=} \PYG{p}{[]}
  \PYG{n}{grid\PYGZus{}sardines} \PYG{o}{=} \PYG{p}{[]}

  \PYG{n}{curr\PYGZus{}x} \PYG{o}{=} \PYG{n}{min\PYGZus{}x}
  \PYG{k}{while} \PYG{n}{curr\PYGZus{}x} \PYG{o}{\PYGZlt{}=} \PYG{n}{max\PYGZus{}x}\PYG{p}{:}
    \PYG{n}{grid\PYGZus{}mackerels}\PYG{o}{.}\PYG{n}{append}\PYG{p}{([])}
    \PYG{n}{grid\PYGZus{}sardines}\PYG{o}{.}\PYG{n}{append}\PYG{p}{([])}
    \PYG{n}{rows} \PYG{o}{+=} \PYG{l+m+mi}{1}
    \PYG{n}{curr\PYGZus{}y} \PYG{o}{=} \PYG{n}{min\PYGZus{}y}
    \PYG{k}{while} \PYG{n}{curr\PYGZus{}y} \PYG{o}{\PYGZlt{}=} \PYG{n}{max\PYGZus{}y}\PYG{p}{:}
      \PYG{n}{grid\PYGZus{}mackerels}\PYG{p}{[}\PYG{o}{\PYGZhy{}}\PYG{l+m+mi}{1}\PYG{p}{]}\PYG{o}{.}\PYG{n}{append}\PYG{p}{(}\PYG{l+m+mi}{0}\PYG{p}{)}
      \PYG{n}{grid\PYGZus{}sardines}\PYG{p}{[}\PYG{o}{\PYGZhy{}}\PYG{l+m+mi}{1}\PYG{p}{]}\PYG{o}{.}\PYG{n}{append}\PYG{p}{(}\PYG{l+m+mi}{0}\PYG{p}{)}
      \PYG{n}{curr\PYGZus{}y} \PYG{o}{+=} \PYG{n}{cell\PYGZus{}size}
    \PYG{k}{if} \PYG{n}{rows} \PYG{o}{==} \PYG{l+m+mi}{1}\PYG{p}{:}
      \PYG{n}{cols} \PYG{o}{=} \PYG{n+nb}{len}\PYG{p}{(}\PYG{n}{grid\PYGZus{}mackerels}\PYG{p}{[}\PYG{o}{\PYGZhy{}}\PYG{l+m+mi}{1}\PYG{p}{])}
    \PYG{n}{curr\PYGZus{}x} \PYG{o}{+=} \PYG{n}{cell\PYGZus{}size}

  \PYG{k}{for} \PYG{p}{(}\PYG{n}{x}\PYG{p}{,} \PYG{n}{y}\PYG{p}{)} \PYG{o+ow}{in} \PYG{n}{mackerels}\PYG{p}{:}
    \PYG{n}{x} \PYG{o}{\PYGZhy{}=} \PYG{n}{min\PYGZus{}x}
    \PYG{n}{y} \PYG{o}{\PYGZhy{}=} \PYG{n}{min\PYGZus{}y}
    \PYG{k}{assert} \PYG{n}{x} \PYG{o}{\PYGZgt{}=} \PYG{l+m+mi}{0} \PYG{o+ow}{and} \PYG{n}{y} \PYG{o}{\PYGZgt{}=} \PYG{l+m+mi}{0}
    \PYG{n}{grid\PYGZus{}mackerels}\PYG{p}{[}\PYG{n}{x} \PYG{o}{//} \PYG{n}{cell\PYGZus{}size}\PYG{p}{][}\PYG{n}{y} \PYG{o}{//} \PYG{n}{cell\PYGZus{}size}\PYG{p}{]} \PYG{o}{+=} \PYG{l+m+mi}{1}
  \PYG{k}{for} \PYG{p}{(}\PYG{n}{x}\PYG{p}{,} \PYG{n}{y}\PYG{p}{)} \PYG{o+ow}{in} \PYG{n}{sardines}\PYG{p}{:}
    \PYG{k}{if} \PYG{n}{x} \PYG{o}{\PYGZlt{}} \PYG{n}{min\PYGZus{}x} \PYG{o+ow}{or} \PYG{n}{x} \PYG{o}{\PYGZgt{}} \PYG{n}{max\PYGZus{}x} \PYG{o+ow}{or} \PYG{n}{y} \PYG{o}{\PYGZlt{}} \PYG{n}{min\PYGZus{}y} \PYG{o+ow}{or} \PYG{n}{y} \PYG{o}{\PYGZgt{}} \PYG{n}{max\PYGZus{}y}\PYG{p}{:}
      \PYG{k}{continue}
    \PYG{n}{x} \PYG{o}{\PYGZhy{}=} \PYG{n}{min\PYGZus{}x}
    \PYG{n}{y} \PYG{o}{\PYGZhy{}=} \PYG{n}{min\PYGZus{}y}
    \PYG{k}{assert} \PYG{n}{x} \PYG{o}{\PYGZgt{}=} \PYG{l+m+mi}{0} \PYG{o+ow}{and} \PYG{n}{y} \PYG{o}{\PYGZgt{}=} \PYG{l+m+mi}{0}
    \PYG{n}{grid\PYGZus{}sardines}\PYG{p}{[}\PYG{n}{x} \PYG{o}{//} \PYG{n}{cell\PYGZus{}size}\PYG{p}{][}\PYG{n}{y} \PYG{o}{//} \PYG{n}{cell\PYGZus{}size}\PYG{p}{]} \PYG{o}{+=} \PYG{l+m+mi}{1}

  \PYG{n}{grid} \PYG{o}{=} \PYG{n}{Grid}\PYG{p}{(}\PYG{n}{rows}\PYG{p}{,} \PYG{n}{cols}\PYG{p}{,} \PYG{n}{grid\PYGZus{}mackerels}\PYG{p}{,} \PYG{n}{grid\PYGZus{}sardines}\PYG{p}{)}

  \PYG{n}{scores} \PYG{o}{=} \PYG{p}{[]}
  \PYG{n}{best\PYGZus{}cell} \PYG{o}{=} \PYG{k+kc}{None}
  \PYG{n}{best\PYGZus{}score} \PYG{o}{=} \PYG{l+m+mi}{0}

  \PYG{n}{grid\PYGZus{}2} \PYG{o}{=} \PYG{n}{Grid}\PYG{p}{(}\PYG{n}{rows}\PYG{p}{,} \PYG{n}{cols}\PYG{p}{,}
                \PYG{n+nb}{list}\PYG{p}{([}\PYG{n+nb}{list}\PYG{p}{(}\PYG{n}{row}\PYG{p}{)} \PYG{k}{for} \PYG{n}{row} \PYG{o+ow}{in} \PYG{n}{grid\PYGZus{}mackerels}\PYG{p}{]),}
                \PYG{n+nb}{list}\PYG{p}{([}\PYG{n+nb}{list}\PYG{p}{(}\PYG{n}{row}\PYG{p}{)} \PYG{k}{for} \PYG{n}{row} \PYG{o+ow}{in} \PYG{n}{grid\PYGZus{}sardines}\PYG{p}{]))}
\end{Verbatim}
\end{tiny}
\end{minipage}
\begin{minipage}[c]{0.48\textwidth}
\begin{tiny}
\begin{Verbatim}[commandchars=\\\{\}]
  \PYG{k}{for} \PYG{n}{i} \PYG{o+ow}{in} \PYG{n+nb}{range}\PYG{p}{(}\PYG{n}{rows}\PYG{p}{):}
    \PYG{n}{scores}\PYG{o}{.}\PYG{n}{append}\PYG{p}{([])}
    \PYG{k}{for} \PYG{n}{j} \PYG{o+ow}{in} \PYG{n+nb}{range}\PYG{p}{(}\PYG{n}{cols}\PYG{p}{):}
      \PYG{n}{curr\PYGZus{}score} \PYG{o}{=} \PYG{n}{score\PYGZus{}greedy}\PYG{p}{(}\PYG{n}{grid\PYGZus{}2}\PYG{p}{,} \PYG{n}{i}\PYG{p}{,} \PYG{n}{j}\PYG{p}{,} \PYG{n+nb}{set}\PYG{p}{())}
      \PYG{n}{scores}\PYG{p}{[}\PYG{n}{i}\PYG{p}{]}\PYG{o}{.}\PYG{n}{append}\PYG{p}{(}\PYG{n}{curr\PYGZus{}score}\PYG{p}{)}
      \PYG{k}{if} \PYG{n}{best\PYGZus{}cell} \PYG{o+ow}{is} \PYG{k+kc}{None} \PYG{o+ow}{or} \PYG{n}{curr\PYGZus{}score} \PYG{o}{\PYGZgt{}} \PYG{n}{best\PYGZus{}score}\PYG{p}{:}
        \PYG{n}{best\PYGZus{}cell} \PYG{o}{=} \PYG{p}{(}\PYG{n}{i}\PYG{p}{,} \PYG{n}{j}\PYG{p}{)}
        \PYG{n}{best\PYGZus{}score} \PYG{o}{=} \PYG{n}{curr\PYGZus{}score}

  \PYG{n}{picked\PYGZus{}cells} \PYG{o}{=} \PYG{n+nb}{set}\PYG{p}{()}
  \PYG{n}{pq} \PYG{o}{=} \PYG{p}{[]}
  \PYG{n}{heapq}\PYG{o}{.}\PYG{n}{heappush}\PYG{p}{(}\PYG{n}{pq}\PYG{p}{,} \PYG{p}{(}\PYG{o}{\PYGZhy{}}\PYG{n}{best\PYGZus{}score}\PYG{p}{,} \PYG{n}{best\PYGZus{}cell}\PYG{p}{))}

  \PYG{n}{curr\PYGZus{}mackerels} \PYG{o}{=} \PYG{l+m+mi}{0}
  \PYG{n}{curr\PYGZus{}sardines} \PYG{o}{=} \PYG{l+m+mi}{0}

  \PYG{k}{def} \PYG{n+nf}{push\PYGZus{}point}\PYG{p}{(}\PYG{n}{cell}\PYG{p}{,} \PYG{n}{picked\PYGZus{}cells}\PYG{p}{,} \PYG{n}{pq}\PYG{p}{):}
    \PYG{n}{delta\PYGZus{}mackerels} \PYG{o}{=} \PYG{l+m+mi}{0}
    \PYG{n}{delta\PYGZus{}sardines} \PYG{o}{=} \PYG{l+m+mi}{0}

    \PYG{k}{def} \PYG{n+nf}{track\PYGZus{}point}\PYG{p}{(}\PYG{n}{test\PYGZus{}cell}\PYG{p}{,} \PYG{n}{picked\PYGZus{}cells}\PYG{p}{):}
      \PYG{k}{assert} \PYG{n}{test\PYGZus{}cell} \PYG{o+ow}{not} \PYG{o+ow}{in} \PYG{n}{picked\PYGZus{}cells}
      \PYG{n}{queue} \PYG{o}{=} \PYG{p}{[}\PYG{n}{test\PYGZus{}cell}\PYG{p}{]}
      \PYG{n}{visited} \PYG{o}{=} \PYG{n+nb}{set}\PYG{p}{()}
      \PYG{k}{while} \PYG{n}{queue}\PYG{p}{:}
        \PYG{n}{curr\PYGZus{}cell} \PYG{o}{=} \PYG{n}{queue}\PYG{o}{.}\PYG{n}{pop}\PYG{p}{()}
        \PYG{k}{if} \PYG{n}{curr\PYGZus{}cell} \PYG{o+ow}{in} \PYG{n}{visited}\PYG{p}{:}
          \PYG{k}{continue}
        \PYG{n}{visited}\PYG{o}{.}\PYG{n}{add}\PYG{p}{(}\PYG{n}{curr\PYGZus{}cell}\PYG{p}{)}
        \PYG{k}{for} \PYG{n}{dx}\PYG{p}{,} \PYG{n}{dy} \PYG{o+ow}{in} \PYG{p}{[(}\PYG{o}{\PYGZhy{}}\PYG{l+m+mi}{1}\PYG{p}{,} \PYG{l+m+mi}{0}\PYG{p}{),} \PYG{p}{(}\PYG{l+m+mi}{0}\PYG{p}{,} \PYG{o}{\PYGZhy{}}\PYG{l+m+mi}{1}\PYG{p}{),} \PYG{p}{(}\PYG{o}{+}\PYG{l+m+mi}{1}\PYG{p}{,} \PYG{l+m+mi}{0}\PYG{p}{),} \PYG{p}{(}\PYG{l+m+mi}{0}\PYG{p}{,} \PYG{o}{+}\PYG{l+m+mi}{1}\PYG{p}{)]:}
          \PYG{n}{next\PYGZus{}x} \PYG{o}{=} \PYG{n}{curr\PYGZus{}cell}\PYG{p}{[}\PYG{l+m+mi}{0}\PYG{p}{]} \PYG{o}{+} \PYG{n}{dx}
          \PYG{n}{next\PYGZus{}y} \PYG{o}{=} \PYG{n}{curr\PYGZus{}cell}\PYG{p}{[}\PYG{l+m+mi}{1}\PYG{p}{]} \PYG{o}{+} \PYG{n}{dy}
          \PYG{k}{if} \PYG{p}{(}\PYG{n}{next\PYGZus{}x} \PYG{o}{\PYGZlt{}} \PYG{l+m+mi}{0}\PYG{p}{)} \PYG{o+ow}{or} \PYG{p}{(}\PYG{n}{next\PYGZus{}x} \PYG{o}{\PYGZgt{}=} \PYG{n+nb}{len}\PYG{p}{(}\PYG{n}{scores}\PYG{p}{))} \PYG{o+ow}{or} \PYG{p}{(}
              \PYG{n}{next\PYGZus{}y} \PYG{o}{\PYGZlt{}} \PYG{l+m+mi}{0}\PYG{p}{)} \PYG{o+ow}{or} \PYG{p}{(}\PYG{n}{next\PYGZus{}y} \PYG{o}{\PYGZgt{}=} \PYG{n+nb}{len}\PYG{p}{(}\PYG{n}{scores}\PYG{p}{[}\PYG{l+m+mi}{0}\PYG{p}{])):}
            \PYG{k}{return} \PYG{p}{[]}
          \PYG{k}{if} \PYG{p}{(}\PYG{n}{next\PYGZus{}x}\PYG{p}{,} \PYG{n}{next\PYGZus{}y}\PYG{p}{)} \PYG{o+ow}{in} \PYG{n}{picked\PYGZus{}cells}\PYG{p}{:}
            \PYG{k}{continue}
          \PYG{k}{if} \PYG{p}{(}\PYG{n}{next\PYGZus{}x}\PYG{p}{,} \PYG{n}{next\PYGZus{}y}\PYG{p}{)} \PYG{o+ow}{in} \PYG{n}{visited}\PYG{p}{:}
            \PYG{k}{continue}
          \PYG{n}{queue}\PYG{o}{.}\PYG{n}{append}\PYG{p}{((}\PYG{n}{next\PYGZus{}x}\PYG{p}{,} \PYG{n}{next\PYGZus{}y}\PYG{p}{))}
      \PYG{k}{return} \PYG{n+nb}{list}\PYG{p}{(}\PYG{n}{visited}\PYG{p}{)}

    \PYG{k}{assert} \PYG{n}{cell} \PYG{o+ow}{not} \PYG{o+ow}{in} \PYG{n}{picked\PYGZus{}cells}
    \PYG{n}{picked\PYGZus{}cells}\PYG{o}{.}\PYG{n}{add}\PYG{p}{(}\PYG{n}{cell}\PYG{p}{)}
    \PYG{n}{delta\PYGZus{}mackerels} \PYG{o}{+=} \PYG{n}{grid}\PYG{o}{.}\PYG{n}{mackerels}\PYG{p}{[}\PYG{n}{cell}\PYG{p}{[}\PYG{l+m+mi}{0}\PYG{p}{]][}\PYG{n}{cell}\PYG{p}{[}\PYG{l+m+mi}{1}\PYG{p}{]]}
    \PYG{n}{delta\PYGZus{}sardines} \PYG{o}{+=} \PYG{n}{grid}\PYG{o}{.}\PYG{n}{sardines}\PYG{p}{[}\PYG{n}{cell}\PYG{p}{[}\PYG{l+m+mi}{0}\PYG{p}{]][}\PYG{n}{cell}\PYG{p}{[}\PYG{l+m+mi}{1}\PYG{p}{]]}

    \PYG{n}{grid\PYGZus{}2} \PYG{o}{=} \PYG{n}{Grid}\PYG{p}{(}\PYG{n}{rows}\PYG{p}{,} \PYG{n}{cols}\PYG{p}{,}
                  \PYG{n+nb}{list}\PYG{p}{([}\PYG{n+nb}{list}\PYG{p}{(}\PYG{n}{row}\PYG{p}{)} \PYG{k}{for} \PYG{n}{row} \PYG{o+ow}{in} \PYG{n}{grid\PYGZus{}mackerels}\PYG{p}{]),}
                  \PYG{n+nb}{list}\PYG{p}{([}\PYG{n+nb}{list}\PYG{p}{(}\PYG{n}{row}\PYG{p}{)} \PYG{k}{for} \PYG{n}{row} \PYG{o+ow}{in} \PYG{n}{grid\PYGZus{}sardines}\PYG{p}{]))}

    \PYG{k}{for} \PYG{n}{dx}\PYG{p}{,} \PYG{n}{dy} \PYG{o+ow}{in} \PYG{p}{[(}\PYG{o}{\PYGZhy{}}\PYG{l+m+mi}{1}\PYG{p}{,} \PYG{l+m+mi}{0}\PYG{p}{),} \PYG{p}{(}\PYG{l+m+mi}{0}\PYG{p}{,} \PYG{o}{\PYGZhy{}}\PYG{l+m+mi}{1}\PYG{p}{),} \PYG{p}{(}\PYG{o}{+}\PYG{l+m+mi}{1}\PYG{p}{,} \PYG{l+m+mi}{0}\PYG{p}{),} \PYG{p}{(}\PYG{l+m+mi}{0}\PYG{p}{,} \PYG{o}{+}\PYG{l+m+mi}{1}\PYG{p}{)]:}
      \PYG{n}{next\PYGZus{}x} \PYG{o}{=} \PYG{n}{cell}\PYG{p}{[}\PYG{l+m+mi}{0}\PYG{p}{]} \PYG{o}{+} \PYG{n}{dx}
      \PYG{n}{next\PYGZus{}y} \PYG{o}{=} \PYG{n}{cell}\PYG{p}{[}\PYG{l+m+mi}{1}\PYG{p}{]} \PYG{o}{+} \PYG{n}{dy}
      \PYG{k}{if} \PYG{p}{(}\PYG{n}{next\PYGZus{}x} \PYG{o}{\PYGZlt{}} \PYG{l+m+mi}{0}\PYG{p}{)} \PYG{o+ow}{or} \PYG{p}{(}\PYG{n}{next\PYGZus{}x} \PYG{o}{\PYGZgt{}=} \PYG{n+nb}{len}\PYG{p}{(}\PYG{n}{scores}\PYG{p}{))} \PYG{o+ow}{or} \PYG{p}{(}
          \PYG{n}{next\PYGZus{}y} \PYG{o}{\PYGZlt{}} \PYG{l+m+mi}{0}\PYG{p}{)} \PYG{o+ow}{or} \PYG{p}{(}\PYG{n}{next\PYGZus{}y} \PYG{o}{\PYGZgt{}=} \PYG{n+nb}{len}\PYG{p}{(}\PYG{n}{scores}\PYG{p}{[}\PYG{l+m+mi}{0}\PYG{p}{])):}
        \PYG{k}{continue}
      \PYG{k}{if} \PYG{p}{(}\PYG{n}{next\PYGZus{}x}\PYG{p}{,} \PYG{n}{next\PYGZus{}y}\PYG{p}{)} \PYG{o+ow}{in} \PYG{n}{picked\PYGZus{}cells}\PYG{p}{:}
        \PYG{k}{continue}
      \PYG{k}{for} \PYG{n}{pt} \PYG{o+ow}{in} \PYG{n}{track\PYGZus{}point}\PYG{p}{((}\PYG{n}{next\PYGZus{}x}\PYG{p}{,} \PYG{n}{next\PYGZus{}y}\PYG{p}{),} \PYG{n}{picked\PYGZus{}cells}\PYG{p}{):}
        \PYG{n}{picked\PYGZus{}cells}\PYG{o}{.}\PYG{n}{add}\PYG{p}{(}\PYG{n}{pt}\PYG{p}{)}
        \PYG{n}{delta\PYGZus{}mackerels} \PYG{o}{+=} \PYG{n}{grid}\PYG{o}{.}\PYG{n}{mackerels}\PYG{p}{[}\PYG{n}{pt}\PYG{p}{[}\PYG{l+m+mi}{0}\PYG{p}{]][}\PYG{n}{pt}\PYG{p}{[}\PYG{l+m+mi}{1}\PYG{p}{]]}
        \PYG{n}{delta\PYGZus{}sardines} \PYG{o}{+=} \PYG{n}{grid}\PYG{o}{.}\PYG{n}{sardines}\PYG{p}{[}\PYG{n}{pt}\PYG{p}{[}\PYG{l+m+mi}{0}\PYG{p}{]][}\PYG{n}{pt}\PYG{p}{[}\PYG{l+m+mi}{1}\PYG{p}{]]}
      \PYG{k}{if} \PYG{p}{(}\PYG{n}{next\PYGZus{}x}\PYG{p}{,} \PYG{n}{next\PYGZus{}y}\PYG{p}{)} \PYG{o+ow}{not} \PYG{o+ow}{in} \PYG{n}{picked\PYGZus{}cells}\PYG{p}{:}
        \PYG{n}{scores}\PYG{p}{[}\PYG{n}{next\PYGZus{}x}\PYG{p}{][}\PYG{n}{next\PYGZus{}y}\PYG{p}{]} \PYG{o}{=} \PYG{n}{score\PYGZus{}greedy}\PYG{p}{(}
            \PYG{n}{grid\PYGZus{}2}\PYG{p}{,} \PYG{n}{next\PYGZus{}x}\PYG{p}{,} \PYG{n}{next\PYGZus{}y}\PYG{p}{,} \PYG{n+nb}{set}\PYG{p}{(}\PYG{n+nb}{list}\PYG{p}{(}\PYG{n}{picked\PYGZus{}cells}\PYG{p}{)))}
        \PYG{n}{heapq}\PYG{o}{.}\PYG{n}{heappush}\PYG{p}{(}\PYG{n}{pq}\PYG{p}{,} \PYG{p}{(}\PYG{o}{\PYGZhy{}}\PYG{n}{scores}\PYG{p}{[}\PYG{n}{next\PYGZus{}x}\PYG{p}{][}\PYG{n}{next\PYGZus{}y}\PYG{p}{],} \PYG{p}{(}\PYG{n}{next\PYGZus{}x}\PYG{p}{,} \PYG{n}{next\PYGZus{}y}\PYG{p}{)))}

    \PYG{k}{return} \PYG{n}{delta\PYGZus{}mackerels}\PYG{p}{,} \PYG{n}{delta\PYGZus{}sardines}

  \PYG{n}{max\PYGZus{}val} \PYG{o}{=} \PYG{o}{\PYGZhy{}}\PYG{l+m+mi}{1}
  \PYG{k}{while} \PYG{n}{pq}\PYG{p}{:}
    \PYG{n}{\PYGZus{}}\PYG{p}{,} \PYG{n}{best\PYGZus{}cell} \PYG{o}{=} \PYG{n}{heapq}\PYG{o}{.}\PYG{n}{heappop}\PYG{p}{(}\PYG{n}{pq}\PYG{p}{)}
    \PYG{k}{if} \PYG{n}{best\PYGZus{}cell} \PYG{o+ow}{in} \PYG{n}{picked\PYGZus{}cells}\PYG{p}{:}
      \PYG{k}{continue}

    \PYG{n}{delta\PYGZus{}mackerels}\PYG{p}{,} \PYG{n}{delta\PYGZus{}sardines} \PYG{o}{=} \PYG{n}{push\PYGZus{}point}\PYG{p}{(}\PYG{n}{best\PYGZus{}cell}\PYG{p}{,} \PYG{n}{picked\PYGZus{}cells}\PYG{p}{,} \PYG{n}{pq}\PYG{p}{)}
    \PYG{n}{curr\PYGZus{}mackerels} \PYG{o}{+=} \PYG{n}{delta\PYGZus{}mackerels}
    \PYG{n}{curr\PYGZus{}sardines} \PYG{o}{+=} \PYG{n}{delta\PYGZus{}sardines}

    \PYG{n}{val} \PYG{o}{=} \PYG{n+nb}{max}\PYG{p}{(}\PYG{l+m+mi}{0}\PYG{p}{,} \PYG{n}{curr\PYGZus{}mackerels} \PYG{o}{\PYGZhy{}} \PYG{n}{curr\PYGZus{}sardines} \PYG{o}{+} \PYG{l+m+mi}{1}\PYG{p}{)}
    \PYG{c+c1}{\PYGZsh{} Convert polygon cells to boundary line, return \PYGZhy{}1 if invalid or too long}
    \PYG{n}{traversal} \PYG{o}{=} \PYG{n}{decode\PYGZus{}to\PYGZus{}output}\PYG{p}{(}
        \PYG{n}{picked\PYGZus{}cells}\PYG{p}{,}
        \PYG{n}{cell\PYGZus{}size}\PYG{p}{,}
        \PYG{n}{min\PYGZus{}x}\PYG{p}{,} \PYG{n}{max\PYGZus{}x}\PYG{p}{,}
        \PYG{n}{min\PYGZus{}y}\PYG{p}{,} \PYG{n}{max\PYGZus{}y}\PYG{p}{)}
    \PYG{k}{if} \PYG{n}{traversal} \PYG{o}{!=} \PYG{o}{\PYGZhy{}}\PYG{l+m+mi}{1} \PYG{o+ow}{and} \PYG{n}{val} \PYG{o}{\PYGZgt{}=} \PYG{n}{max\PYGZus{}val}\PYG{p}{:}
      \PYG{n}{max\PYGZus{}val} \PYG{o}{=} \PYG{n}{val}
  \PYG{k}{return} \PYG{n}{max\PYGZus{}val}
\end{Verbatim}
\end{tiny}
\end{minipage}
\caption{A snippet of the input parsing function and the greedy algorithm backbone for the AtCoder Heuristic Contest 039 (Purse Seine Fishing) for a given \texttt{cell\_size}. Note that the backbone is calling \texttt{score\_greedy}---the function which provides the score of adding a particular cell to the polygon.}
\label{fig:backbone_fish}
\end{figure*}

\end{document}